\documentclass[journal]{IEEEtran}
\usepackage{amsmath,amsfonts}
\usepackage{amssymb}
\usepackage{mathtools}
\usepackage{amsthm}
\usepackage{bbm}
\usepackage{dsfont}
\usepackage{mathrsfs}
\usepackage{algorithmic}
\usepackage{algorithm}
\usepackage{array}
\usepackage[linkcolor=blue,colorlinks=true, citecolor = blue]{hyperref}
\usepackage{textcomp}
\usepackage{stfloats}
\usepackage{url}
\usepackage{verbatim}
\usepackage{graphicx}
\usepackage{cite}
\usepackage{xcolor}
\usepackage{booktabs} 
\usepackage{multirow}
\usepackage{multicol}
\usepackage{subfigure}
\usepackage{setspace}

\theoremstyle{plain}
\newtheorem{theorem}{Theorem}[section]

\newtheorem{lemma}[theorem]{Lemma}
\newtheorem{corollary}[theorem]{Corollary}
\newtheorem{definition}[theorem]{Definition}
\newtheorem{assumption}[theorem]{Assumption}
\theoremstyle{remark}

\begin{document}
\allowdisplaybreaks[4]
\title{On the Trade-off between Flatness and Optimization \\ in Distributed Learning}

\author{Ying Cao, Zhaoxian Wu, Kun Yuan, Ali H. Sayed, \IEEEmembership{Fellow,~IEEE}
\thanks{Authors Ying Cao and Ali H. Sayed are with the Institute of Electrical and Micro Engineering in EPFL,
Lausanne. Zhaoxian Wu was with the Sun Yat-sen University. Kun Yuan is with the Center for Machine Learning Research (CMLR) in Peking University.
Emails: \{ying.cao, ali.sayed\}@epfl.ch, wuzhx23@mail2.sysu.edu.cn, kunyuan@pku.edu.cn.}}



\maketitle

\begin{abstract}
This paper proposes a theoretical framework to evaluate and compare the performance of stochastic gradient algorithms for distributed learning in relation to their behavior around local minima in nonconvex environments. Previous works have noticed that convergence toward flat local minima tend to enhance the generalization ability of learning algorithms. This work discovers three interesting results. First, it shows that decentralized learning strategies are able to escape faster away from local minima and favor convergence toward flatter minima relative to the centralized solution.  Second, in decentralized methods, the consensus strategy has a worse excess-risk performance than diffusion, giving it a better chance of escaping from local minima and favoring flatter minima. Third, and importantly, the ultimate classification accuracy is not solely dependent on the flatness of the local minimum but also on how well a learning algorithm can approach that minimum. In other words, the classification accuracy is a function of both flatness and optimization performance. In this regard, since diffusion has a lower excess-risk than consensus, when both algorithms are trained starting from random initial points, diffusion enhances the classification  accuracy.  The paper examines the interplay between the two measures of flatness and optimization error closely. One important conclusion is that decentralized strategies deliver in general enhanced classification accuracy because they strike a more favorable balance between flatness and optimization performance compared to the centralized solution.
\end{abstract}


\section{Introduction and related works}\label{introduction}
As modern society continues to evolve into a more interconnected world, and large-scale applications become increasingly prevalent, the interest in distributed learning has grown significantly \cite{sayed2014adaptive,chang2020distributed, GoyalDGNWKTJH17}.
In the distributed scenario, a group of agents, each with its own objective function, collaborate to optimize a global objective. To achieve the goal, two popular gradient descent based methodologies are commonly employed in algorithm design \cite{nedic2009distributed,dimakis2010gossip, sayed2014adaptation,sayed_2023}. The first one, known as the \emph{centralized} method, requires all agents to send their data to a central server that manages all computations. In the second approach, which is called \emph{decentralized}, agents are connected by a graph topology, and they process data locally and exchange information with their neighbors. {It is widely known that decentralized implementations offer enhanced data privacy, resilience to failure and reduced computation burden compared to the centralized approach \cite{sayed2014adaptation, ZhuH0ST23, LianZZHZL17,abs-2111-04287, YingYCHPY21,SunLW23,xu2021dp}. \emph{Consensus} and \emph{diffusion} strategies are two prominent decentralized methods in the optimization and machine learning communities \cite{ChenS13, ZhuH0ST23, ZhuHZNST22, sayed2014adaptive, VlaskiS21, kayaalp2022dif, WangPS22, yuan2023removing}. }

\begin{figure}[!t]
\centering
\includegraphics[width=2.5in]{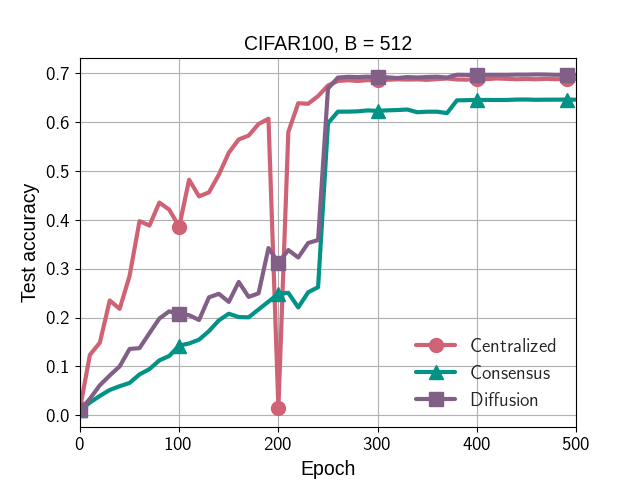}
\caption{The evolution of the test accuracy for centralized, consensus and diffusion training with local batch size 512 on CIFAR100 dataset. Consensus exhibits the worst test performance. The manuscript provides detailed theoretical analysis to explain how distributed learning methods perform in relation to flatness, test accuracy, and optimization performance. }
\label{fig_intro}
\end{figure}

{Extensive works exist in the literature on the fundamental properties of the three strategies including centralized, consensus, and diffusion. For example, }  \cite{sayed2014adaptation, sayed_2023, VlaskiS21, VlaskiS21a,LianZZHZL17,KoloskovaLBJS20, AlghunaimY22} provided convergence guarantees for convex and non-convex environments. The results generally suggest that decentralized operations could, at best, only match the optimization performance of the centralized approach \cite{sayed2014adaptation, ChenS15b, YuanAYS20}, or potentially degrade the generalization ability of models \cite{sun2021stability,ZhuHZNST22, DengSL023}. Nevertheless, the work \cite{00010KJS21} empirically demonstrated that the network disagreement, i.e., the distance among agents, in the middle of decentralized training can enhance generalization over centralized training. Afterwards, the work \cite{ZhuH0ST23} {verified} that the consensus strategy enhances the generalization performance of models over centralized solutions in the large-batch setting from the perspective of sharpness-aware minimization (SAM), which is primarily designed to reduce the sharpness of machine learning models \cite{ForetKMN21}. Specifically, the results in \cite{ZhuH0ST23} showed that the network disagreement among agents, which changes over time during training with the consensus method, implicitly introduces an extra regularization term related to SAM to the original risk function, which helps drive the iterates to flatter models  that are known to generalize better. However, as Figure \ref{fig_intro} shows, flatter models obtained by consensus do not always guarantee better test accuracy in the context of distributed learning. This is because the final test accuracy depends on both generalization and optimization performance. Although decentralized training tends to favor flatter minima that generalize better than those found by the centralized method, it could also implicitly degrade the optimization performance in some cases. This observation motivates us to examine the behavior of centralized and decentralized strategies more closely and to compare more directly both their generalization and optimization abilities. 

{As a result, this work discovers three interesting results. First, it shows that decentralized learning strategies are able to escape faster away from local minima and favor convergence toward flatter minima relative to the centralized solution.  Second, in decentralized methods, the consensus strategy has a worse excess-risk performance than diffusion, giving it a better chance of escaping from local minima and favoring flatter minima. Third, and importantly, the ultimate classification accuracy is not solely dependent on the flatness of the local minimum but also on how well a learning algorithm can approach that minimum. In other words, the classification accuracy is a function of both flatness and optimization performance. In this regard, diffusion has a better optimization performance than consensus. Therefore, when both algorithms are trained starting from random initial points, diffusion leads to better accuracy. The paper examines the interplay between the two measures of flatness and optimization error closely. One important conclusion is that decentralized strategies deliver in general enhanced classification accuracy because they strike a more favorable balance between flatness and optimization performance compared to the centralized solution. Moreover, consensus and diffusion exhibit different trade-off abilities between optimization and flatness across different training scenarios.} Due to space limitations, complete proofs for all lemmas and theorems are provided in the supplementary files.

\textbf{Notation.} In this paper, we use boldface letters to denote random variables. For a scalar-valued function $f(\mu)$ where $\mu$ is a scalar, we say that $f(\mu) = \pm O(\mu)$ if $\vert f(\mu)\vert \le c\vert\mu\vert$ for some constant $c>0$. We say $f(\mu) = o(\mu)$ if $f(\mu)/\mu \to 0$ as $\mu\to0$. For any matrix $X$, $X = O(\mu
)$ signifies that the magnitude of the individual entries of X are $O(\mu)$ or the norm of $X$ is $O(\mu)$. 
\subsection{Related works}
Understanding the optimization behavior of {mini-batch \textit{stochastic gradient algorithms} (SGA)} and their influence on the generalization of models has emerged as a prominent topic of interest in recent years. Large-batch training is increasingly important in distributed learning for its potential to enhance the training speed and scalability of modern neural networks. However, it has been observed in the literature that {small-batch SGA} tends to flatter minima than the large-batch version \cite{KeskarMNST17}, and that flat minima usually generalize better than sharp ones \cite{Lyu0A22,gatmiry2023inductive, WuS23, ZhuWYWM19, NacsonRSS22, JiangNMKB20}. Intuitively, the loss values around flat minima change slowly when the model parameters are adjusted, thereby reducing the disparity between the training and test data \cite{KeskarMNST17}. The works by \cite{MandtHB17, MoriLLU22, XieSS21, ZhouF0XHE20, ZhuWYWM19, PesmePF21} utilized the stochastic differential equation (SDE) approach as a fundamental tool to analyze the regularization effects of mini-batch SGA. This type of analysis inherently introduces an extra layer of approximation error due to substituting the discrete-time update by a {differential equation}. Moreover, to enable the analysis with SDE, it is necessary to introduce some extra assumptions on the gradient noise, such as being Gaussian distributed \cite{ZhuWYWM19,XieSS21}, parameter-independent \cite{MandtHB17} or heavy-tailed distributed \cite{SimsekliSG19}. To overcome these limitations, another line of work \cite{WuME18, WuWS22, WuS23}  focused on the dynamical stability of mini-batch SGA with discrete-time analysis. Specifically, they determined conditions on the Hessian matrices to ensure that the distance to a local minimum does not increase, thereby stabilizing the iterates in the vicinity of a local minimum.

\subsection{Contribution}
Contrary to the aforementioned works focusing on the single-agent case, we will study in this work the optimization bias of \emph{distributed} algorithms in the multi-agent setting, which allows us to compare the performance of various distributed strategies. Inspired by \cite{ZhuWYWM19}, we start from investigating the escaping efficiency of algorithms {in the vicinity of a local minimum}, and then relate it to the trade-off between flatness and optimization. Our contributions are listed as follows:

(1) We propose a general framework to examine the escaping efficiency of distributed algorithms from {a local minimum} while remaining in the discrete-time domain. Since the dependence of the Hessian matrix on the immediate iterate in the original recursion makes the optimization analysis intractable, it is necessary to resort to a short-term model where the original Hessian is replaced by one evaluated at {a local minimum}. We rigorously verify that the approximation error between the short-term and true models is negligible, ensuring that the short-term model represents the true model accurately enough.
Afterwards, we obtain closed-form expressions of the short-term excess risk which quantifies the escaping efficiency. Note that we follow a different discrete-time analysis from the dynamical stability approach used in \cite{WuME18, WuWS22, WuS23} that focused on studying the stability of algorithms. Our emphasis is on quantifying the extent to which algorithms can escape local minima.

(2) {We compare the escaping efficiency of the centralized, consensu, and diffusion methods. Our analysis reveals that decentralized approaches--namely, consensus and diffusion--gain additional efficacy from the network heterogeneity and graph structure, making them more efficient in escaping a local minimum than the centralized strategy. Also, we show that the consensus strategy has a better chance of escaping from local minima than diffusion. Furthermore, we establish a connection between escaping efficiency and the flatness metric, showing that higher escaping efficiency encourages algorithms to favor flatter minima. This suggests that decentralized methods are more likely to converge to flatter minima compared to their centralized counterpart. }

(3) If the additional power generated by the network heterogeneity and graph structure are not sufficiently strong to allow decentralized methods to successfully leave the current basin, then both decentralized and centralized methods will be stuck within the basin of the current local minimum. A similar rule also holds for the comparison between consensus and diffusion. This motivates us to pursue next a long-term analysis, which corresponds to the optimization performance, {i.e., how close algorithms approach the local minimum.} In this context, we verify that the extra power boosting the escaping efficiency could inversely deteriorate the optimization performance. This reveals an inherent trade-off between flatness and optimization.

(4) We finally illustrate the performance of centralized, consensus and diffusion training strategies on real data. {Basically, favoring flatter minima alone does not necessarily translate into a better classification accuracy. The accuracy is also dependent on the optimization performance of an algorithm. We show that decentralized methods in general achieve a more favorable balance between flatness and optimization compared to the centralized algorithm, thereby exhibiting better test accuracy. Furthermore, a closer comparison between consensus and diffusion reveals distinct behavior across different training scenarios. In the typical scenario of training from scratch, diffusion tends to lead to better accuracy because it provides a more favorable balance between flatness and optimization performance. However, when the decentralized methods are initialized near a local minimizer pretrained by the centralized algorithm, the ``worse" optimization performance by consensus is alleviated and its classification accuracy becomes better.}

\section{Problem Statement}\label{ps}
\subsection{Empirical risk minimization}
Consider a collection of $K$ agents (or nodes) linked by a graph topology. Each agent receives a streaming sequence of data realizations arising from independently distributed observations. The agents wish to collaborate to solve a distributed learning problem of the following form: 
\begin{equation}\label{global_j}
 \mathop{\mathrm{min}}\limits_{{w}\in \mathbbm{R}^M} \left\{J( w) \overset{\Delta}{=} \frac{1}{K}\sum\limits_{k=1}^K J_k( w)\right\}
\end{equation}
where $M$ is the dimension of the unknown vector $w$,  $J_k(w)$ is the risk function for the $k$-th agent, and $J(w)$ is the aggregate risk across the graph. Each individual risk is defined as the empirical average of a loss function over a collection of training data points, namely, 
\begin{equation}\label{jk_em}
    J_k(w) \overset{\Delta}{=}  \frac{1}{N_k}\sum\limits_{i=1}^{N_k} Q_k(w; {x}_{k,i})
\end{equation}
where $Q_k(\cdot)$ is the possibly nonconvex loss function and the $\{x_{k,i} = \{\gamma_{k,i}, h_{k,i}\}\}$ refers to $N_k$ training samples with feature vector $\gamma_{k,i}$ and true label $h_{k,i}$ arising as realizations from a random source $\boldsymbol{x}_k$ associated with agent $k$.

Since we aim to understand the optimization behavior of algorithms around local minima of nonconvex risk functions, we need to distinguish between the local minima of $J_k(w)$ and $J(w)$. Thus, we let $w^{\star}$ denote a local minimizer for the aggregate risk $J(w)$ {(i.e., meaning $w^{\star}$ admits a local minimum $J(w^{\star})$)}, and let $w_k^{\star}$ denote a local minimizer for the individual risk $J_k(w)$. If all agents in the network share the same risk function $J_k(w) = J(w)$, then $w^{\star} = w_k^{\star}$, and we refer to the network as being {\em homogeneous}. In this case, all local minimizers of $J(w)$ also admit local minima of $J_k(w)$. Otherwise, we refer to the network as being {\em heterogeneous}, where the $w_k^{\star}$ of all agents can be distinct among themselves and also in relation to $w^{\star}$. In this paper, we focus on this latter more general case.

\subsection{{Stochastic gradient algorithms}}
We consider three popular classes of algorithms that can be used to seek a solution for (\ref{global_j}). The first algorithm is the mini-batch {\em centralized} method, in which all data from across the graph are shared with a central processor. At every instant $n$, $B$ samples denoted by $\{\boldsymbol{x}_{k,n}^b\}$ are selected uniformly at random with replacement from the training data available for each agent $k$. Starting from a random initial condition $\boldsymbol{w}_{-1}$, the central processor updates the estimate by using:
\begin{equation}\label{a_centra}
    \boldsymbol{w}_n = \boldsymbol{w}_{n-1} - \mu\times\frac{1}{KB}\sum\limits_{k=1}^{K}\sum\limits_{b=1}^{B} \nabla_w Q_k(\boldsymbol{w}_{n-1};\boldsymbol{x}_{k,n}^b)
\end{equation}
Here, $B$ is the batch size and $\mu$ is a small positive step-size parameter {(also called learning rate)}. Observe that we are using boldface letters to refer to the data samples and iterates in (\ref{a_centra}); it is our convention in this paper to use the boldface notation to refer to random quantities. 

The other two algorithms are of the decentralized type, where the data remains local and agents interact locally with their neighbors to solve (\ref{global_j}) through a collaborative process. We consider two types of decentralized methods, which have been studied extensively in the literature, namely, the consensus and diffusion strategies \cite{nedic2009distributed,dimakis2010gossip, sayed2014adaptation,sayed_2023,ChenS15a,ChenS15b}.  

Before listing the algorithms, we describe the graph structure that drives their operation. The agents are assumed to be linked by a weighted graph topology. The weight on the link from agent $\ell$ to agent $k$ is denoted by $a_{\ell k}$; this value is used to scale information sent from $\ell$ to $k$. Each $a_{\ell k}$ is non-negative and lies within $[0,1]$; it will be strictly positive if there exists a link from $\ell$ to $k$ over which information can be shared. We collect the $\{a_{\ell k}\}$ into a $K\times K$ matrix $A$. In this paper, we consider a symmetric \emph{doubly-stochastic} matrix $A$: the entries on each column of $A$ are normalized to add up to $1$.
\begin{assumption}\label{as1}
    (\textbf{Strongly-connected graph}). The graph is assumed to be strongly connected. This means that there exists a path with nonzero weights $\{a_{\ell k}\}$ linking any pair of agents and, in addition, at least one node $k$ in the network has a self-loop with $a_{kk}>0$.
    
    $\hfill\square$ 
\end{assumption}
It follows from the Perron-Frobenius theorem \cite{sayed2014adaptation} that $A$ has a single eigenvalue at 1. Moreover, if we let $\pi = \{\pi_k\}_{k=1}^{K}$ denote the corresponding right eigenvector, then all its entries are positive and they can be normalized to add up to $1$:
\begin{align}\label{comb_matrix}
A\pi = \pi, \quad \mathbbm{1}^{\sf T}\pi=1, \quad \pi_k>0
\end{align}
For the doubly-stochastic matrix $A$, we have $\pi = \frac{1}{K}\mathbbm{1}$. This means that the entries $\{\pi_k\}$ of the Perron vector $\pi$ are identical.

The diffusion strategy consists of the following two steps:
\begin{subequations}
\begin{align}
\label{x_e}
    &\boldsymbol{\phi}_{k,n} =\boldsymbol{w}_{k,n-1} - \frac{\mu}{B}\sum\limits_{b=1}^{B}\nabla_w Q_k(\boldsymbol{w}_{k,n-1};\boldsymbol{x}_{k,n}^{b})\\
\label{combine_e}
&\boldsymbol{w}_{k,n}  = \sum\limits_{\ell \mathcal{\in N}_k} a_{\ell k} \boldsymbol{\phi}_{\ell,n}  
\end{align}
\end{subequations}
At every iteration $n$, every agent $k$ samples $B$ data points and uses $(\ref{x_e})$ to update its iterate $\boldsymbol{w}_{k,n-1}$ to the intermediate value $\boldsymbol{\phi}_{k,n}$. Subsequently, the same agent combines the intermediate iterates from across its neighbors using (\ref{combine_e}). The symbol $\mathcal{N}_k$ denotes the collection of neighbors of $k$.  It is useful to remark that the centralized implementation (\ref{a_centra}) can be viewed as a special case of (\ref{x_e})--(\ref{combine_e}) if we select the combination matrix as $A=\pi\mathbbm{1}^{\sf T}$. 

In comparison, the consensus strategy involves the following steps:
\begin{subequations}
\begin{align}
\label{x_e_consen}
&\boldsymbol{\phi}_{k,n} = \sum\limits_{\ell \mathcal{\in N}_k} a_{\ell k} \boldsymbol{w}_{\ell,n-1}\\
\label{combine_e_consen}
&\boldsymbol{w}_{k,n}  = \boldsymbol{\phi}_{k,n}  - \frac{\mu}{B}\sum\limits_{b=1}^{B}\nabla_w Q_k(\boldsymbol{w}_{k,n-1};\boldsymbol{x}_{k,n}^{b})
\end{align}
\end{subequations}
In this case, the existing iterates $\boldsymbol{w}_{\ell,n-1}$ are first combined to generate the intermediate value $\boldsymbol{\phi}_{k,n}$, after which (\ref{combine_e_consen}) is applied. Observe the asymmetry on the right-hand side in (\ref{combine_e_consen}). The starting iterate is $\boldsymbol{\phi}_{k,n}$, while the loss functions are evaluated at the different iterates $\boldsymbol{w}_{k,n-1}$. In contrast, the same iterate $\boldsymbol{w}_{k,n-1}$ appears in both terms on the RHS of (\ref{combine_e}). This symmetry has been shown to enlarge the stability range of diffusion implementations over its consensus counterparts in the convex case, namely, diffusion is mean-square stable for a wider range of step-sizes $\mu$ \cite{TuS12, sayed2014adaptive, sayed_2023}. 

 \subsection{Escaping efficiency from local minima}\label{eer}
We first characterize the basin (or valley) of a local minimum:
\begin{definition}
   (\textbf{Basin of attraction} \cite{MoriLLU22}.) For a given local minimizer $w^{\star}$ {(or the corresponding local minimum $J(w^{\star})$}, its basin of attraction $\Omega(w^\star)$ (or the valley of $w^{\star}$) is defined as the set of all points starting from which $\boldsymbol{w}_{k,n}$ or $\boldsymbol{w}_n \to w^{\star}$ as $n\to \infty$ if the step size $\mu$ is sufficiently small and there is no 
    noise in the gradient-descent algorithms.

$\hfill\square$
\end{definition}
 
\begin{figure}[ht]
\centering
\includegraphics[width=3.5in]{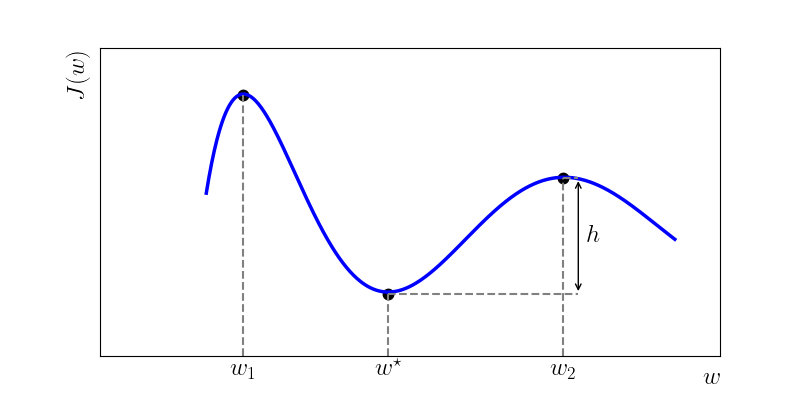}
\caption{An illustration of the valley around a local minimizer $w^\star$.}
\label{conceptual}
\end{figure}

 We illustrate the basin of attraction related to $w^{\star}$ in Figure \ref{conceptual}, where $\Omega(w^{\star}) = (w_1, w_2)$ is a bounded open set. Also, we use $\partial\Omega(w^{\star}) $ to denote the boundary of $\Omega(w^{\star})$, according to which $\partial \Omega(w^{\star}) = \{w_1, w_2\}$ in Figure \ref{conceptual}. If $\boldsymbol{w}_{k,n}$ or $\boldsymbol{w}_n \notin  \Omega(w^{\star})$, then we say the algorithm escapes from the basin of $w^{\star}$ exactly. 
 
To analyze the escaping behavior of learning algorithms from local minima, it is necessary to quantify their escaping ability. Inspired by the continuous-time definition of \emph{escaping efficiency} for the single-agent case from \cite{ZhuWYWM19}, and by the metrics used to assess the optimization performance of stochastic learning algorithms \cite{sayed2014adaptation, sayed_2023}, we use the following metric to measure the discrete-time escaping efficiency for general non-convex environments.
\begin{definition}
   (\textbf{Escaping efficiency}). Assume all agents start from points close to a local minimizer $w^{\star}$, the escaping efficiency over the network at iteration $n$ is defined by:
    \begin{align}\label{er_d}
    \mathrm{ER}_n \overset{\Delta}{=}\frac{1}{K}\sum\limits_{k=1}^{K} \mathds{E} J(\boldsymbol{w}_{k,n}) - J(w^\star)
\end{align}
{The larger the value of $\mathrm{ER}_n$ is, the farther the network model--comprising $K$ local models--will be from the local minimum $J(w^{\star})$ on average, indicating higher escaping efficiency}. If the algorithm ultimately converges to $w^\star$, then the larger $\mathrm{ER}_n$ is, the worse the optimization performance will be.
$\hfill\square$ 
\end{definition}

{We wish to highlight two aspects in relation to the definition of the escaping efficiency of algorithms:} 

{First, $\mathrm{ER}_n$ quantifies the ability of the network model to escape $w^{\star}$ on average. Basically, $\mathrm{ER}_n$ assesses the excess-risk value across the graph on average, serving as a metric for the deviation between the excess risk at the current model at iteration $n$ and the local minimum.  As Figure \ref{conceptual} shows, within the local basin $(w_1, w_2)$, a larger excess risk corresponds to a greater distance from the local minimum $J(w^\star)$ for any single model $w$. Since this paper focuses on network models composed of $K$ local agents, we analyze the mean deviation across the network and use (\ref{er_d}).
Therefore, for a fixed $n$, a larger value for $\mathrm{ER_n}$ implies an expected faster escape from a local minimum (since the iterates will be further away from it).}

{Second, a larger value for $\mathrm{ER}_n$ suggests a greater likelihood to leave $\Omega(w^{\star})$}. To justify whether an algorithm escapes from a local basin or not, earlier studies \cite{bovier2004metastability,ibayashi2023why, NguyenSGR19} have established various criteria based on the analysis methods they use. They nevertheless generally adhered to the principle of the minimum effort to reach the boundary of the local basin. In a similar vein, we introduce the \emph{risk barrier}, which is defined as the infimum of the risk gap between the boundary points and the local minimum:
 \begin{align}\label{risk_b}
     h \overset{\Delta}{=} \inf\limits_{w\in \partial \Omega(w^\star)} \; \left\{J(w) - J(w^{\star})\right\}
 \end{align}
For example, the risk barrier of $w^\star$ in Figure \ref{conceptual} is $J(w_2) - J(w^\star)$. {For a network model, a larger $\mathrm{ER}_n$ reflects a higher average deviation from the local minimum, making escape more probable. Specifically, when $\mathrm{ER}_n$ exceeds the corresponding risk barrier defined in (\ref{risk_b}), namely, $\mathrm{ER}_n \ge h$, the network exits $\Omega(w^{\star})$ on average.}

{Therefore, a larger $\mathrm{ER_n}$ indicates higher escaping efficiency. If, on the other hand, the network gets stuck around (or converges to) $w^{\star}$, then 
    $\mathrm{ER}_n$ in that case would serve as a measure of the resulting optimization performance\cite{sayed2014adaptation,sayed_2023}.}

\subsection{Flatness metrics}
Next, we formally characterize the notion of \emph{flat minima}. Consider a local minimizer $w^{\star}$ and a small drift $\Delta w$ around it. The change in the risk value can be approximated by
\begin{align}\label{fm}
    J(w^{\star} + \Delta w) - J(w^{\star}) \approx \frac{1}{2}\Vert\Delta w\Vert^2_{H^{\star}}
\end{align}
where $H^{\star}$ refers to the(positive definite) Hessian matrix of $J(w)$ at $w^{\star}$.
{If the above change in the risk value is small then we say $w^{\star}$ admits a \emph{flat minimum}. Otherwise, if the change in the risk value is large, then we say $w^{\star}$ admits a \emph{sharp} minimum.}  Motivated by (\ref{fm}), various metrics related to the Hessian matrix are applied in the literature to measure the flatness of local minima, e.g., the spectral norm $\Vert H^{\star}\Vert_2$\cite{WuS23,WuME18}, the Frobenius norm $\Vert H^{\star}\Vert_F$\cite{WuS23,WuWS22} and the trace $\mathrm{Tr}(H^{\star})$\cite{ahn2024escape,WuS23,wen2023sharpness}.
In this paper, we use the \emph{trace of the Hessian} as a flatness measure, but the other two metrics may also be applied due to the norm equivalence, namely:
\begin{align}\label{neq}
    \frac{1}{M}\mathrm{Tr}(H^{\star}) \le \Vert H^{\star} \Vert_2 \le \Vert H^{\star} \Vert_F \le \mathrm{Tr}(H^{\star}) \le M \Vert H^{\star}\Vert_2 
\end{align}
Note that for highly ill-conditioned minima, which is commonly observed in over-parameterized neural networks \cite{ZhuWYWM19,YaoGLKM18}, the first $d$ largest eigenvalues of $H^{\star}$ dominate the remaining $M-d$ eigenvalues where $d\ll M$. In this case, the norm equivalence in (\ref{neq}) can be more tightly guaranteed since now the $M$ in the upper and lower bounds can be approximately replaced by $d$. More insights related to the flatness measure $\mathrm{Tr}(H^{\star})$ can be found in \cite{ahn2024escape,wen2023sharpness}. 

\section{Escaping efficiency of multi-agent learning}\label{eemal}
Motivated by the definitions in the last section, we now examine the $\mathrm{ER_n}$ performance of decentralized and centralized methods.

\subsection{Modeling conditions}
We introduce the following commonly-used assumptions.  

{First, to analyze how an algorithm escapes from a given location, it is essential to assume that the algorithm is operating near that point. This point of view is widely adopted in the literature when examining the ability of algorithms to escape saddle points \cite{VlaskiS21, VlaskiS21a, vlaski2021second} or local minima \cite{ZhuWYWM19, XieSS21, ibayashi2023why, NguyenSGR19}. For example, references \cite{VlaskiS21, VlaskiS21a, vlaski2021second} assume that the stochastic gradient algorithm begins from a point sufficiently close to a saddle point, where the gradient norm is small. The work \cite{ZhuWYWM19} assumes the algorithm starts precisely at a local minimum, while \cite{XieSS21} assumes the model is near a local minimizer, thus allowing the Hessian at the current point to be well approximated by the Hessian at the local minimizer. Such assumptions enable a focused analysis of the algorithms' behavior near critical points. In our study, since we focus on analyzing the behavior of algorithms in the vicinity of a local minimizer, $w^{\star}$, we assume that all agents start from points close to it.
\begin{assumption}\label{ass_ori}
(\textbf{Starting points of all agents}). All models initiate their updates from points sufficiently close to $w^\star$, namely,
    \begin{align}\label{ass_ori_e}
        \mathds{E}\Vert \boldsymbol{w}_{k,-1} - w^\star\Vert^4 \le o\left(\frac{\mu^2}{B^2}\right)
    \end{align}
 $\hfill\square$ 
\end{assumption}
By applying Jensen's inequality to (\ref{ass_ori_e}), we obtain:
\begin{align}\label{ass_ori_e_2}
        \mathds{E}\Vert \boldsymbol{w}_{k,-1} - w^\star\Vert^2 \le  \left(\mathds{E}\Vert \boldsymbol{w}_{k,-1} - w^\star\Vert^4 \right)^{\frac{1}{2}}\le o\left(\frac{\mu}{B}\right)
    \end{align}}

Assumption \ref{ass_ori} is justified from traditional convergence results. For instance, assume we pick any step size $\mu' \le o(\mu)$, such as $\mu' = \mu^2$. Then, it is well-known that the centralized stochastic gradient algorithm approaches an 
 $O(\frac{\mu'}{B})$ neighborhood of a local minimum of $J(w)$ after sufficient iterations in nonconvex environments (see, e.g., \cite{sayed_2023,vlaski2021second, VlaskiS21a, vlaski2019diffusion}), namely,
\begin{align}\label{gra_d}
    \lim_{n \to \infty} \mathds{E}\Vert \nabla J(\boldsymbol{w}_{k,n}) - \nabla J(w^{\star})\Vert^2 \le O\left(\frac{\mu'}{B}\right) = o\left(\frac{\mu}{B}\right)
\end{align}
where $\nabla J(w^{\star}) = 0$ since $w^{\star}$ admits a local minimum of $J(w)$. Additionally, under the centralized setting, the influence of the term related to the difference between local models and their centroid is eliminated. Moreover, using the local convexity around $w^{\star}$, it follows from \cite{sayed2014adaptation} that:
\begin{align}\label{model_d}
    &\mathds{E}\Vert \boldsymbol{w}_{k,n} - w^\star \Vert^2 \le o\left(\frac{\mu}{B}\right), \; \mathds{E}\Vert \boldsymbol{w}_{k,n} - w^\star \Vert^4 \le o\left(\frac{\mu^2}{B^2}\right)
\end{align}
where the term related to the network heterogeneity vanishes in the centralized setting. As a result, the bounds in Eqs.~\eqref{gra_d} and \eqref{model_d} are determined solely by gradient noise. In summary, with sufficiently small step sizes, agents can attain an $o(\frac{\mu}{B})$-neighborhood of $w^{\star}$.

We further require the loss function $Q_k$ of all agents to be smooth \cite{VlaskiS21, VlaskiS21a}.
\begin{assumption} (\textbf{Smoothness condition}.)\label{ass_smooth}
    For each agent $k$, the gradient of $Q_k$ relative to $w$ is Lipschitz. Specifically, for any $w_1, w_2\in \mathbbm{R}^M$, it holds that:
    \begin{align}
       \Vert \nabla Q_k(w_2;x) - \nabla Q_k(w_1;x) \Vert  \le L\Vert w_2 - w_1\Vert
    \end{align}
     $\hfill\square$ 
\end{assumption}

Next, consider the Hessian matrix of agent $k$ at $w^{\star}$ denoted by:
\begin{align}\label{localhs}
  H_k^\star \overset{\Delta}{=} \nabla^2 J_k(w^\star) = \frac{1}{N_k}\sum\limits_{i=1}^{N_k}\nabla^2 Q_k (w^{\star}; x_{k,i})
\end{align}
and the global Hessian matrix of $J(w)$ at $w^{\star}$:
\begin{align}\label{glb_h}
        \bar{H} \overset{\Delta}{=} \frac{1}{K}\sum_{k=1}^K \nabla^2 J_k(w^\star) = \frac{1}{K}\sum_{k=1}^K H_k^\star
\end{align}
\begin{assumption}\label{sh}
(\textbf{Small Hessian disagreement}).  The local Hessian at $w^\star$ is sufficiently close to the global Hessian, namely,
   \begin{align}
       \Vert H_k^{\star} - \bar{H}\Vert \le \epsilon
   \end{align}
   with a small constant $\epsilon$. $\hfill\square$
\end{assumption}
Assumption \ref{sh} can be satisfied when the data heterogeneity among agents is sufficiently small. For example, if all agents observe independently and identically distributed data, and consider the stochastic Hessian defined by:
\begin{align}
    \tilde{H} = \mathds{E}_{\boldsymbol{x}}\nabla^2 Q(w;\boldsymbol{x})
\end{align}
Then by resorting to the matrix Bernstein inequality \cite{tropp2015introduction}, for any $\epsilon>0$, we have:
\begin{align}
    \mathbbm{P}\left(\Vert H_k - \tilde{H} \Vert \le \frac{\epsilon}{2}\right)\ge 1 - Me^{-\frac{3\epsilon^2N_k}{4L(3L+\epsilon)}}
\end{align}
Thus, when each agent collects sufficient amount of data, all local Hessian matrices are guaranteed to be close to the stochastic Hessian with high probability, from which we get:
\begin{align}\label{hkl}
\left\Vert H_k - \frac{1}{K}\sum\limits_\ell H_\ell \right\Vert &\le  \sum\limits_\ell \frac{1}{K} \Vert H_k - H_\ell \Vert 
\notag\\
&\le \frac{1}{K} \sum\limits_\ell (\Vert H_k - \tilde{H}\Vert + \Vert H_\ell - \tilde{H}\Vert) \notag\\
&\le \epsilon
\end{align}
Intuitively, even when all agents independently collect data from different distributions, if the data distribution among agents are sufficiently close to each other, then (\ref{hkl}) can still be satisfied. 

\subsection{Properties of gradient noise}
To enable the analysis, we describe properties associated with the gradient noise process for later use. For any $w\in \mathbbm{R}^M$, the stochastic gradient noise at agent $k$ at iteration $n$ is defined by the difference:
\begin{align}\label{sk1_s}
\boldsymbol{s}_{k,n}(w) \overset{\Delta}{=} \nabla Q_k(w;\boldsymbol{x}_{k,n}) - \nabla J_k(w) 
\end{align}
{which corresponds to the case of $B=1$ (i.e., batch size equal to 1)}. In the mini-batch case,  the gradient noise is instead given by:
\begin{align}\label{sk1}
\boldsymbol{s}_{k,n}^{B}(w) \overset{\Delta}{=} \frac{1}{B}\sum\limits_{b=1}^{B}\nabla Q_k(w;\boldsymbol{x}_{k,n}^{b}) - \nabla J_k(w) 
\end{align}
We denote the covariance matrix of the gradient noise by:
\begin{align}\label{drswn_s}
    R_{s,k,n}(w) \overset{\Delta}{=} \mathds{E}\left\{ \boldsymbol{s}_{k,n}(w)\boldsymbol{s}_{k,n}(w)^{\sf T}\right\}
\end{align}
which is symmetric and non-negative definite. Likewise, for the mini-batch case,
\begin{align}\label{drswn}
     R_{s,k,n}^B(w) \overset{\Delta}{=} \mathds{E}\left\{ \boldsymbol{s}^{B}_{k,n}(w)\boldsymbol{s}^{B}_{k,n}(w)^{\sf T}\right\}
\end{align}
\begin{lemma}\label{lm_gn}
(\textbf{Gradient noise terms}). Let $\mathcal{F}_{n-1}$ denote the filtration generated by the past history of the random process $\boldsymbol{w}_{k,j}$ for all $j\le n-1$ and $k = 1,\ldots,K$. For any $\boldsymbol{w} \in\mathcal{F}_{n-1}$, we define the error vector
\begin{align}
    \tilde{\boldsymbol{w}} \overset{\Delta}{ = } w^{\star} - \boldsymbol{w}
\end{align}
Then, under assumption \ref{ass_smooth}, it holds that the gradient noise defined in $(\ref{sk1})$ has zero mean:
\begin{align}
     \label{s0_convex} \mathds{E}\left[\boldsymbol{s}_{k,n}^{B }\vert{\mathcal{F}}_{n-1}\right]=0 
\end{align}
while its second and fourth-order moments are upper bounded by terms related to the batch size as follows:
\begin{align}
\label{f_s2_l}
\mathds{E}[\Vert \boldsymbol{s}_{k,n}^{B}(\boldsymbol{w})\Vert^2\vert\mathcal{F}_{n-1} ]&\le \alpha_2\Vert \tilde{\boldsymbol{w}}\Vert^2 + \beta_2^2\\
\label{f_s4_l}
\mathds{E}[\Vert \boldsymbol{s}_{k,n}^{B}(\boldsymbol{w})\Vert^4\vert\mathcal{F}_{n-1} ] & \le \alpha_4\Vert \tilde{\boldsymbol{w}}\Vert^4  +\beta_4^2
\end{align}
where the nonnegative scalars $\{\alpha_2, \beta_2^2, \alpha_4, \beta_4^2\}$ are on the order of
\begin{align}
    &\alpha_2 = O\left(\frac{1}{B}\right),\;\; \;\beta_2^2 = O\left(\frac{1}{B}\right)\\
    &\alpha_4 = O\left(\frac{1}{B^2}\right),\;\; \beta_4^2 = O\left(\frac{1}{B^2}\right)
\end{align}
Moreover, the covariance matrices (\ref{drswn_s}) and (\ref{drswn}) are scaled versions of each other:
\begin{align}\label{rsknbB}
     R_{s,k,n}^{B}(w) = \frac{1}{B}R_{s,k,n}(w)
\end{align}
\end{lemma}
\begin{proof}
See Appendix \ref{pfgs}.  
\end{proof}
For later use, we define the gradient covariance matrix of the global risk function at $w^\star$ for the case $B=1$ as:
 \begin{align}\label{d_barrs}
     \bar{R}_s &\overset{\Delta}{=} \mathds{E}\left\{\left(\frac{1}{K}\sum\limits_{k=1}^{K}\boldsymbol{s}_{k,n}(w^\star)\right)\left(\frac{1}{K}\sum\limits_{\ell=1}^{K} \boldsymbol{s}_{\ell,n}(w^\star)\right)^{\sf T}\right\}
 \end{align}

\subsection{Network performance analysis}
We now proceed with the network analysis, which will enable us to derive expressions for the value of $\mathrm{ER}_n$. We will then use these expressions to deduce properties about the behavior of the centralized and decentralized algorithms in relation to flat minima and escaping efficiency. To begin with, we follow the decomposition from \cite{sayed2014adaptation} and note that the combination matrix $A$ admits an eigen-decomposition of the form:
\begin{equation}\label{decomp_A}
    A = VPV^{\sf T}
\end{equation}
where the matrices $V$ and $P$ are:
\begin{equation}\label{vpv1}
     V = \left[\frac{1}{\sqrt{K}}\mathbbm{1} \quad V_\alpha \right],\quad
    P = \left[
        \begin{array}{cc}
        1&0\\
        0&P_\alpha
        \end{array}
    \right]
\end{equation}
Here, $P_{\alpha}\in \mathbbm{R}^{(K-1)\times(K-1)}$ is a diagonal matrix with elements from the second largest-magnitude eigenvalue $\lambda_2$ to the smallest-magnitude eigenvalue $\lambda_K$ of $A$ appearing on the diagonal, and $V_\alpha \in \mathbbm{R}^{K\times(K-1)}$. Consider the extended network version policy:
\begin{align}\label{e_A}
    \mathcal{A} \overset{\Delta}{= } A \otimes I_{M}
\end{align}
where $\otimes$ denotes the Kronecker product, and $I_M$ is the identity matrix of size $M$.
Then, $\mathcal{A}$ satisfies
\begin{align}\label{decom_A}
    \mathcal{A} = \mathcal{V}\mathcal{P}\mathcal{V}^{\sf T}
\end{align}
where 
\begin{equation}
    \mathcal{V} = V\otimes I_M,\quad \mathcal{P} = P\otimes I_M,\quad \mathcal{V}^{\sf T} = V^{\sf T}\otimes I_M
\end{equation}
We further collect quantities from across the network into the block variables:
\begin{align}\label{mcv}
    &\mathcal{H}_{n} \overset{\Delta}{= }\mathrm{diag}\left\{H_{1,n}(\boldsymbol{w}_{1,n}),H_{2,n}(\boldsymbol{w}_{2,n}),\ldots H_{K,n}(\boldsymbol{w}_{K,n})\right\}\notag\\
    &\boldsymbol{s}_n^B \overset{\Delta}{= } \mathrm{col}\left\{ \boldsymbol{s}_{k,n}^{B}(\boldsymbol{w}_{k,n-1}) \right\}\notag\\
    &d \overset{\Delta}{= }  \mathrm{col}\left\{\nabla J_k(w^{\star})\right\}\notag\\
    &{\widetilde{\boldsymbol{\scriptstyle\mathcal{W}}}}_{n} = \mathrm{col}\left\{\tilde{\boldsymbol{w}}_{k,n} \right\} \overset{\Delta}{= } \mathrm{col}\left\{ w^\star - \boldsymbol{w}_{k,n}\right\}
\end{align}
where $\mathrm{col}$ denotes a block column vector, and each Hessian matrix $H_{k,n}(\boldsymbol{w}_{k,n})$ is defined by
 \begin{align}
    H_{k,n}(\boldsymbol{w}_{k,n}) \overset{\Delta}{=} \left[\int_0^1 \nabla^2 J_k(w^\star - t\tilde{\boldsymbol{w}}_{k,n})dt\right]
\end{align}

Using  (\ref{e_A})--(\ref{mcv}), we can rewrite algorithms (\ref{a_centra}), (\ref{x_e})--(\ref{combine_e}) and (\ref{x_e_consen})--(\ref{combine_e_consen}) using a unified description as follows:
\begin{align}\label{uni_dis}
{\widetilde{\boldsymbol{\scriptstyle\mathcal{W}}}}_{n} = \mathcal{A}_2(\mathcal{A}_1 - \mu\mathcal{H}_{n-1}){\widetilde{\boldsymbol{\scriptstyle\mathcal{W}}}}_{n-1}  + \mu \mathcal{A}_2 d + \mu \mathcal{A}_2\boldsymbol{s}_n^B
\end{align}
where
\begin{align}
   \mathcal{A}_1 = A_1 \otimes I_M,\; \mathcal{A}_2 = A_2 \otimes I_M
\end{align}
and the choices for the matrices $\{A_1, A_2\}$ depend on the nature of the algorithm. For instance, for the consensus algorithm we set 
\begin{align}
   A_1 = A,\; A_2 = I_{K}
\end{align}
while for diffusion we set
\begin{align}
    A_1 = I_K,\; A_2 = A
\end{align}
Moreover, the centralized method can be viewed as a special case of diffusion for which
\begin{align}
    A_1 = I_K,\; A_2 = A = \pi\mathbbm{1}^{\sf T} = \frac{1}{K}\mathbbm{1}\mathbbm{1}^{\sf T}
\end{align}

Unfortunately, the dependence of $\mathcal{H}_{n-1}$ on ${\widetilde{\boldsymbol{\scriptstyle\mathcal{W}}}}_{n-1}$ makes the analysis with (\ref{uni_dis}) intractable. To overcome this challenge, and inspired by \cite{VlaskiS21,VlaskiS21a, sayed2014adaptation}, we introduce the alternative block diagonal matrix
\begin{align}
\mathcal{H} \overset{\Delta}{= } \mathrm{diag}\left\{H_{1}^\star,H_{2}^\star,\ldots H_{K}^\star\right\}
\end{align}
where the $H_k^{\star}$ are evaluated at $w^{\star}$. Using $\mathcal{H}$ in place of $\mathcal{H}_n$, we replace recursion (\ref{uni_dis}) by the following so-called short-term model:
\begin{align}\label{uni_dis_a}
{\widetilde{\boldsymbol{\scriptstyle\mathcal{W}}}}_{n}' = \mathcal{A}_2(\mathcal{A}_1 - \mu\mathcal{H}){\widetilde{\boldsymbol{\scriptstyle\mathcal{W}}}}_{n-1}'  + \mu \mathcal{A}_2 d + \mu \mathcal{A}_2\boldsymbol{s}_n^B
\end{align}
This approximation naturally raises the following question: How accurate can the short-term model in (\ref{uni_dis_a}) approximate the true recursion in (\ref{uni_dis})? {A similar issue has been addressed for nonconvex optimization problems in the vicinity of saddle points in \cite{VlaskiS21}. Motivated by this analysis, we answer the above question in the following two lemmas,} where we separately show the results for the decentralized and centralized methods over a finite time horizon.
\begin{lemma}\label{mse}
(\textbf{Deviation bounds of decentralized methods}). For a fixed small step size $\mu$ and local batch size $B$ such that 
\begin{align}\label{cbmu}
    \frac{1}{B} \le c\mu^{\eta}
\end{align}
where $c \ll \infty$ and $\eta \ge 0$, and under assumptions \ref{as1}, \ref{ass_ori}, and \ref{ass_smooth}, it can be verified for consensus and diffusion that the second and fourth-order moments of 
 ${\widetilde{\boldsymbol{\scriptstyle\mathcal{W}}}}_{n}$ are upper bounded in a finite number of iterations such that $n \le O(\frac{1}{\mu})$, namely,
 \begin{align}
  \label{th_3.6_w2}
\mathds{E}\Vert{\widetilde{\boldsymbol{\scriptstyle\mathcal{W}}}}_{n}\Vert^2 &\le O\left(\frac{\mu}{B}\right) + O(\mu^2) = O(\mu^\gamma)\\
 \label{th_3.6_w4}
\mathds{E}\Vert{\widetilde{\boldsymbol{\scriptstyle\mathcal{W}}}}_{n}\Vert^4 &\le O(\mu^{2\gamma})
\end{align}
where $\gamma = \min\{1+\eta,2\}$. Also, the second-order moment of  ${\widetilde{\boldsymbol{\scriptstyle\mathcal{W}}}}_{n}'$ is upper bounded by
\begin{align}
\label{th_3.6_wp2}
\mathds{E}\Vert{\widetilde{\boldsymbol{\scriptstyle\mathcal{W}}}}_{n}'\Vert^2 &\le O(\mu^\gamma)     
\end{align}
Moreover, the approximation error caused by the short-term model in (\ref{uni_dis_a}) is upper bounded by
\begin{align}
\Big\vert \mathds{E}\Vert\widetilde{\boldsymbol{\scriptstyle\mathcal{W}}}_n'\Vert^2 - \mathds{E}\Vert\widetilde{\boldsymbol{\scriptstyle\mathcal{W}}}_n \Vert^2 \Big\vert \le O(\mu^{1.5\gamma})
\end{align}
\end{lemma}
\begin{proof}
See Appendices \ref{mse2}, \ref{mse4} and \ref{aest}.
\end{proof}

\begin{lemma}\label{mse_centra}
(\textbf{Deviation bounds of the centralized method}). Under the same conditions of Lemma \ref{mse}, the second-order and fourth-order moments of 
 ${\widetilde{\boldsymbol{\scriptstyle\mathcal{W}}}}_{n}$, the second-order moment of  ${\widetilde{\boldsymbol{\scriptstyle\mathcal{W}}}}_{n}'$, and the approximation error of the short term model, related to the centralized method are guaranteed to be upper bounded in a finite number of iterations $n \le O(\frac{1}{\mu})$. Basically, it holds that 
\begin{align}
\mathds{E}\Vert{\widetilde{\boldsymbol{\scriptstyle\mathcal{W}}}}_{n}\Vert^2 &\le O(\mu^{1+\eta})\\
\mathds{E}\Vert{\widetilde{\boldsymbol{\scriptstyle\mathcal{W}}}}_{n}'\Vert^2 &\le O(\mu^{1+\eta})\\
\mathds{E}\Vert{\widetilde{\boldsymbol{\scriptstyle\mathcal{W}}}}_{n}\Vert^4 &\le O(\mu^{2(1+\eta))})
\end{align}
and
\begin{align}
\Big\vert \mathds{E}\Vert\widetilde{\boldsymbol{\scriptstyle\mathcal{W}}}_n'\Vert^2 - \mathds{E}\Vert\widetilde{\boldsymbol{\scriptstyle\mathcal{W}}}_n \Vert^2 \Big\vert \le O(\mu^{1.5(1+\eta))})
\end{align}
\end{lemma}
\begin{proof}
See Appendices \ref{mse2}, \ref{mse4} and \ref{aest}.
\end{proof}
{Comparing the bounds in Lemmas \ref{mse} and \ref{mse_centra}, we observe that the upper bounds on the second-order moments associated with the decentralized methods incorporate extra $O(\mu^2)$ terms compared to the centralized method. These extra terms arise from the consensus distance, as defined in \cite{00010KJS21}, which measures the distance between the local models and their centroid.  
This consensus distance is also related to both the network heterogeneity and the graph structure. Detailed proofs of these results can be found in Appendix \ref{mse2}. We will later discuss how the presence of the $O(\mu^2)$ terms impacts the escaping efficiency.}

Lemmas \ref{mse} and \ref{mse_centra} demonstrate that $\mathds{E}\Vert{\widetilde{\boldsymbol{\scriptstyle\mathcal{W}}}}_{n}\Vert^2$ and  $\mathds{E}\Vert{\widetilde{\boldsymbol{\scriptstyle\mathcal{W}}}}_{n}'\Vert^2$ dominate  $\vert \mathds{E}\Vert\widetilde{\boldsymbol{\scriptstyle\mathcal{W}}}_n'\Vert^2 - \mathds{E}\Vert\widetilde{\boldsymbol{\scriptstyle\mathcal{W}}}_n \Vert^2\vert $ for all methods when $\mu$ is sufficiently small. In other words, the approximation error between the mean square deviation of the short-term and true models can be omitted compared with the size of the true models. Furthermore, we can manipulate the bounds in  Lemmas \ref{mse} and \ref{mse_centra} to obtain expression for $\mathrm{ER}_n$ performance. Using (\ref{er_d}) and (\ref{fm}),  we can approximate the $\mathrm{ER}_n$ as follows:
\begin{align}\label{erave}
    \mathrm{ER}_n = \frac{1}{2K}\mathds{E}\Vert{\widetilde{\boldsymbol{\scriptstyle\mathcal{W}}}}_{n}\Vert^2_{I\otimes\bar{H}}+ o(\mathds{E}\Vert{\widetilde{\boldsymbol{\scriptstyle\mathcal{W}}}}_{n}\Vert^2) \notag\\
     = \frac{1}{2K}\mathds{E}\Vert{\widetilde{\boldsymbol{\scriptstyle\mathcal{W}}}}_{n}'\Vert^2_{I\otimes\bar{H}}+ o(\mathds{E}\Vert{\widetilde{\boldsymbol{\scriptstyle\mathcal{W}}}}_{n}\Vert^2) 
\end{align}
which means that $\mathrm{ER}_n$ can be evaluated by means of the short-term recursion. Thus, the original recursion in (\ref{uni_dis}) can be replaced by (\ref{uni_dis_a}) in the small step-size regime. More details about the equality (\ref{erave}) can be found in Appendix \ref{per}.

Building upon the results of Lemmas \ref{mse} and \ref{mse_centra}, we next analyze the escaping efficiency of  the three 
algorithms by using (\ref{uni_dis_a}), and establish the following theorem.
{
\begin{theorem}\label{th_er}
{(\textbf{Escaping efficiency of distributed algorithms})}.
Consider a network of agents running distributed algorithms covered by the short-term model (\ref{uni_dis_a}). Under assumptions \ref{as1}, \ref{ass_ori},  \ref{ass_smooth}, and \ref{sh}, and after $n$ iterations with 
\begin{align}
    n \le O(1/\mu)
\end{align}
it holds that  
\begin{align}
\label{ep_centra}
   \mathrm{ER}_{n,cen} = & {\frac{\mu}{B} e(n)} + o(\mu^{1+\eta})\\
\label{ep_consen}  
   \mathrm{ER}_{n,con} = &{\frac{\mu}{B} e(n)+ \mu^2 f_{con}(n)} \pm  o(\mu^2) \pm o(\mu^{1+\eta})\\
\label{ep_diff}
   \mathrm{ER}_{n,dif} = &  {\frac{\mu}{B}e(n)+ \mu^2 f_{dif}(n)} \pm  o(\mu^2) \pm  o(\mu^{1+\eta})
\end{align}
where
\begin{align}
\label{en}
&e(n) = \frac{1}{4} \mathrm{Tr}\left(\left(I - (I - \mu \bar{H})^{2(n+1)}\right)\bar{R}_s \right)\\
\label{fconn}
&f_{con}(n) = \frac{1}{2K}\Vert d^{\sf T}\mathcal{V}_\alpha( I - \mathcal{P}_{\alpha})^{-1}(I - \mathcal{P}_{\alpha}^{n+1})\Vert^2_{I\otimes\bar{H}}\\
\label{fdifn}
&f_{dif}(n) = \frac{1}{2K}\Vert d^{\sf T}\mathcal{V}_\alpha \mathcal{P}_\alpha( I - \mathcal{P}_{\alpha})^{-1}(I - \mathcal{P}_{\alpha}^{n+1})\Vert^2_{I\otimes\bar{H}}
\end{align}
and $\mathrm{{ER}}_{n,cen}$, $\mathrm{{ER}}_{n,con}$, and $ \mathrm{{ER}}_{n,dif}$ represent the excess risk of the centralized, consensus and diffusion methods at iteration $n$, respectively. 
\end{theorem}
\begin{proof}
See Appendix \ref{per}.
\end{proof}}

Theorem \ref{th_er} shows the escaping efficiency of the three algorithms around a local minimum over a finite time horizon. We emphasize that the $O(\frac{\mu}{B})$ or $O(\frac{\mu}{B}) + O(\mu^2)$ terms in (\ref{ep_centra})--(\ref{ep_diff}) are the dominant contributors. Therefore, focusing on these terms is sufficient to compare the escaping efficiency of the three distributed algorithms. Basically, for all of three methods, larger $\mu$ enables higher escaping efficiency. Also, as mentioned in Appendix \ref{per}, for the algorithms to successfully exit the local basin they would require $O(1/{\mu})$ iterations. Thus, if the algorithms cannot leave the local minimum in $O(1/\mu)$ iterations, then we say they cannot effectively escape the local basin and are therefore trapped in it. {Furthermore, as discussed in Appendix \ref{per}, decentralized methods implicitly include the additional $O(\mu^2)$ terms related to the consensus distance. The network heterogeneity and graph structure ensure that the consensus distance is non-zero so that improve the escaping efficiency.} That is, for homogeneous networks where all agents share same local minimizers $w^{\star}$, we have
\begin{align}
    \nabla J_k(w^\star) = \nabla J(w^\star) = 0
\end{align}
which makes $d = 0$. Meanwhile, in the centralized setting, we have $V_\alpha = 0$ and $P_\alpha= 0$. In these cases, the extra $O(\mu^2)$ terms are 0. On the other hand, in heterogeneous networks, the nonzero $O(\mu^2)$ terms lead to larger $\mathrm{ER}_{n}$ values for decentralized methods compared to the centralized strategy.

However, the difference between the centralized and decentralized methods is only significant in the large-batch regime. Specifically, if $B$ is not sufficiently large such that $\eta < 1 $, then we have $1+\eta <2$ under which the additional $O(\mu^2)$ terms in (\ref{ep_consen}) and (\ref{ep_diff}) will be dominated by the $o(\mu^{1+\eta})$ terms associated with the approximated error due to (\ref{uni_dis_a}) and the initialization condition in Assumption \ref{ass_ori}. As the batch size $B$ increases, however, the 
effect of the gradient noise will progressively decrease and the influence of the $O(\mu^2)$ terms will become more prominent.
In the extreme scenario when the full-batch gradient descent is applied, the centralized method becomes completely noise-free, which means that there is no noise to help the algorithm escape from local minima. 
In contrast, the noisy terms related to network heterogeneity and graph structure continue to exist in decentralized approaches. This makes decentralized methods more effective at escaping from local minima than the centralized method in the large-batch regime. However, larger values of $\mathrm{ER}_n$, while good for escaping efficiency, they nevertheless worsen the optimization performance. We will elaborate on this trade-off later in the steady state when $n\to \infty$.

We can further compare the escaping efficiency of diffusion and consensus, and show that the extra $\mathcal{P}_\alpha$ appearing in $ \mathrm{ER}_{n,dif}$ leads to smaller excess risk for diffusion than consensus. 
\begin{corollary}\label{coro2}
{\rm(\textbf{Smaller $\mathrm{ER}_n$ for diffusion}).}
   Under the same conditions of Theorem \ref{th_er}, it holds that
   \begin{align}
       \mathrm{ER}_{n,dif} \le \mathrm{ER}_{n,con}
   \end{align}
\end{corollary}
\begin{proof}
See Appendix \ref{pcoro2}.
\end{proof}
Corollary \ref{coro2} implies that consensus runs farther away from the local minimum than the diffusion strategy for the same number of iterations due to its worse excess-risk performance. This translates into faster escape for consensus compared to diffusion.

\section{Trade-off between flatness and optimization}
In this section, we elaborate on the important trade-off between flatness and optimization.

Suppose we run a decentralized or centralized algorithm to solve a possibly nonconvex optimization problem. One useful question is to investigate where the algorithm would prefer to go if it escapes from the current basin. In other words, assume the algorithm escapes from the basin around some local minima and starts evolving until it settles down in the basin of another local minimum, we would like to examine what preferential properties does this second minimum have relative to the earlier one. To answer this question, we will relate the escaping efficiency of algorithms measured by $\mathrm{ER}_n$ and the flatness of local minima measured by $\mathrm{Tr}(\bar{H})$.

From expressions (\ref{ep_centra})--(\ref{ep_diff}), and in the one-dimensional case where $\mathrm{Tr}(\bar{H})$ is a scalar, it is obvious that larger values of $\mathrm{Tr}(\bar{H})$, i.e., sharper minima, enable higher escaping efficiency for all three methods. Thus, it is more likely for the three algorithms to escape from sharp minima than flat ones in the one-dimensional case.  When it comes to the higher dimensions, it is generally intractable to clarify the relationship between $\mathrm{ER}_n$ and $\mathrm{Tr}(\bar{H})$ directly. Fortunately, motivated by the Markov inequality, we can appeal to an upper bound for $\mathrm{ER}_n$ denoted by $\mathcal{U}_n$. Specifically, according to the Markov inequality \cite{sayed_2023}, the probability of the excess risk value to remain below a basis threshold $h$ satisfies:
\begin{align}\label{markov}
    \mathds{P}\left(\frac{1}{K}\sum\limits_{k=1}^{K} J(\boldsymbol{w}_{k,n}) - J(w^\star) \le h\right) \ge 1 - \frac{\mathrm{ER}_n}{h} \ge 1 - \frac{\mathcal{U}_n}{h}
\end{align}
where, from (\ref{ep_centra})--(\ref{fdifn}), the upper bound for the dominate terms of the three methods can be derived:
\begin{align}
e(n) = & \frac{1}{4}\mathrm{Tr}\bigg(\Big(I - (I - \mu \bar{H})^{2(n+1)}\Big){\bar{R}_s} \bigg)  \notag\\
\le & {\frac{1}{4}\lambda_{\max}(\bar{R}_s)\mathrm{Tr}\Big(I - (I - \mu \bar{H})^{2(n+1)}\Big)}\notag\\
 \le & {\frac{1}{4}\lambda_{\max}(\bar{R}_s)\sum\limits_{i=1}^{M} \left(1 - (1 - \mu \lambda_i(\bar{H}))^{2(n+1)}\right)}\notag\\
 \overset{\Delta}{=}& \mathcal{U}\big(e(n)\big)
\end{align}
where $\lambda_i(\bar{H})$ represents the $i_{th}$ eigenvalue of $\bar{H}$, and $\lambda_{\max}(\bar{R}_s)$ denotes the largest eigenvalue of $\bar{R}_s$. Also,
\begin{align}
 f_{con}(n) &= \frac{1}{2K}\Vert d^{\sf T}\mathcal{V}_\alpha( I - \mathcal{P}_{\alpha})^{-1}(I - \mathcal{P}_{\alpha}^{n+1})\Vert^2_{I\otimes\bar{H}}\notag\\
 &\le\frac{1}{2}\lambda_{\max}(\bar{H})\Vert d^{\sf T}\mathcal{V}_\alpha( I - \mathcal{P}_{\alpha})^{-1}(I - \mathcal{P}_{\alpha}^{n+1})\Vert^2\notag\\
 &\le\frac{1}{2}\mathrm{Tr}(\bar{H})\Vert d^{\sf T}\mathcal{V}_\alpha( I - \mathcal{P}_{\alpha})^{-1}(I - \mathcal{P}_{\alpha}^{n+1})\Vert^2\notag\\
 &= \mathcal{U}\big(f_{con}(n)\big)
\end{align} 
and, similarly,
\begin{align}
 f_{dif}(n) &= \frac{1}{2K}\Vert d^{\sf T}\mathcal{V}_\alpha\mathcal{P}_{\alpha}( I - \mathcal{P}_{\alpha})^{-1}(I - \mathcal{P}_{\alpha}^{n+1})\Vert^2_{I\otimes\bar{H}}\notag\\
 &\le\frac{1}{2}\mathrm{Tr}(\bar{H})\Vert d^{\sf T}\mathcal{V}_\alpha\mathcal{P}_{\alpha}( I - \mathcal{P}_{\alpha})^{-1}(I - \mathcal{P}_{\alpha}^{n+1})\Vert^2 \notag\\
 & = \mathcal{U}\big(f_{dif}(n)\big)
\end{align}
Then, the upper bound variables for the three methods are given by
\begin{align}
\label{u_centra}
   \mathcal{U}_{n, cen} = &\frac{\mu}{B}\mathcal{U}\big(e(n)\big)\\
   \label{u_con}
    \mathcal{U}_{n, con} = &\frac{\mu}{B}\mathcal{U}\big(e(n)\big) + \mu^2\mathcal{U}\big(f_{con}(n)\big) \\
    \label{u_dif}
    \mathcal{U}_{n, dif} = &\frac{\mu}{B}\mathcal{U}\big(e(n)\big) + \mu^2\mathcal{U}\big(f_{dif}(n)\big)
\end{align}
Observe that these upper bounds on the escaping efficiency are positively correlated with the eigenvalues of $\bar{H}$, i.e., the sharpness of the local minimum. Substituting (\ref{u_centra})--(\ref{u_dif}) into (\ref{markov}) separately, we find that flat minima where $\mathrm{Tr}(\bar{H})$ is small provide high probability guarantees that the algorithms stay around the corresponding local basin. 

Combining the discussion of both one and higher-dimensional cases, we conclude that the three algorithms prefer to stay around flat minima, which means that they favor flatter basin if they successfully escape from a current minimum. Moreover, as discussed earlier for Theorem \ref{th_er}, higher escaping efficiency makes decentralized algorithms more likely to escape from a local minimum than the centralized method. This fact further indicates that decentralized methods favor flatter minima than their centralized counterpart. {Incidentally, the expression for $\mathrm{ER}_{n,cen}$ in Theorem \ref{th_er} is consistent with the well-known observations in the literature \cite{KeskarMNST17,WuS23}, which highlight that larger step sizes or smaller batch sizes with the mini-batch SGA in the single-agent setup steer the model toward flatter minima.} However, as already noted before, higher escaping efficiency may deteriorate the optimization performance. This motivates us to analyze the excess-risk performance in the long run when $n \to \infty$, which corresponds to the optimization performance in the steady state.

Strictly speaking, the short-term model defined in (\ref{uni_dis_a}) may not be valid for nonconvex risk functions in the long run. Fortunately, it has been rigorously verified that (\ref{uni_dis_a}) can approximate well the true model (\ref{uni_dis}) when $n\to \infty$ in the strongly convex case \cite{sayed2014adaptation}. Since here we want to examine the convergence behavior of algorithms around local minima given that the algorithms are stuck in the current basin, we can resort to a  strong convexity assumption around local minima, for which the following result can be guaranteed.
\begin{corollary} 
\label{th_er_op} {\rm{(\textbf{Steady-state excess risks}).}}
Consider a network of agents running distributed algorithms covered by (\ref{uni_dis_a}). After sufficient iterations such that $n\to \infty$, under assumptions \ref{as1}, \ref{ass_ori},  \ref{ass_smooth} and \ref{sh}, and assuming the algorithms are already trapped in the basin of a local minimizer $w^{\star}$ and $J(w)$ is locally strongly-convex around $w^\star$, it holds that 
\begin{align}
\label{ep_centra_op}
   \mathrm{ER}_{\infty,cen} = & \frac{\mu }{4B}\mathrm{Tr}\left(\bar{R}_s \right)\pm O\left(\mu^{1.5(1+\eta)}\right)\\
   \label{ep_consen_op}
   \mathrm{ER}_{\infty,con} = &\frac{\mu }{4B}\mathrm{Tr}\left(\bar{R}_s \right) + \frac{\mu^2}{2K}\Vert d^{\sf T}\mathcal{V}_\alpha( I - \mathcal{P}_{\alpha})^{-1}\Vert^2_{I\otimes\bar{H}} \notag\\
   &\pm O(\mu^{1.5\gamma})   \pm  o(\mu^2)\\
   \label{ep_diff_op}
   \mathrm{ER}_{\infty,dif} = & \frac{\mu }{4B}\mathrm{Tr}\left(\bar{R}_s\right) + \frac{\mu^2}{2K}\Vert d^{\sf T}\mathcal{V}_\alpha P_\alpha( I - \mathcal{P}_{\alpha})^{-1}\Vert^2_{I\otimes\bar{H}} \notag\\
   &\pm O(\mu^{1.5\gamma})   \pm  o(\mu^2)
\end{align}
and, moreover,
\begin{align}
   \mathrm{ER}_{\infty,cen} \le \mathrm{ER}_{\infty,dif} \le \mathrm{ER}_{\infty,con} 
\end{align}
\end{corollary}
\begin{proof}
See Appendix \ref{per_l}.
\end{proof}
In Corollary \ref{th_er_op}, the factors that help algorithms escape local minima are now seen to adversely affect the optimization performance. Again, the difference between decentralized and centralized methods is only significant in the large-batch regime. By integrating the findings of Theorem \ref{th_er} and Corollaries \ref{coro2} and \ref{th_er_op}, we deduce that network heterogeneity and graph structure inject additional noisy terms into decentralized methods and these facilitate their escape from sharp minima compared to the centralized method. However, these added noisy terms incorporate some deterioration into the long-term optimization performance. Furthermore, although consensus exhibits higher escaping efficiency than diffusion, this comes at the expense of reduced optimization performance. This observation
reveals an intrinsic trade-off between flatness and optimization within the context of multi-agent learning. {In addition, the results indicate a trade-off in the choice of the step size  $\mu$. While larger $\mu$
can help the algorithms escape from sharp minima, it also increases the risk that the algorithms may fail to converge altogether. This motivates the use of the piecewise learning-rate schedule \cite{HeZRS16, LinSPJ20} in practice. That is, a moderately small $\mu$ is applied during the initial stage of  training to assist in escaping sharp minima. Later, a smaller $\mu$ is adopted to ensure convergence.}

\section{Simulation results}\label{sec_ex}
\begin{figure}[ht]
\vskip 0.1in
\begin{center}
\subfigure[Random]{\includegraphics[width=.4\columnwidth]{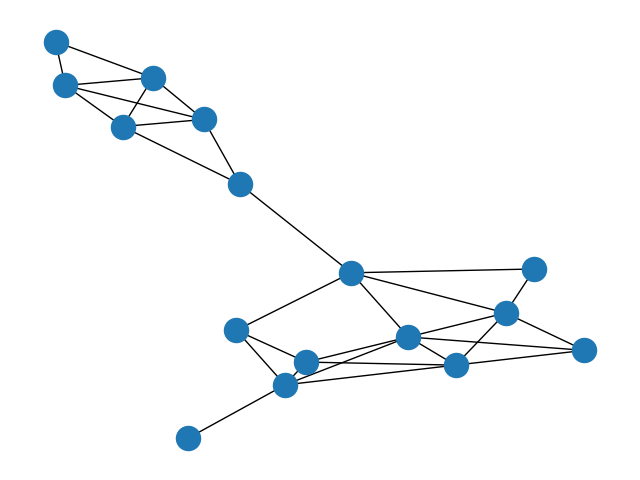}}
\subfigure[Ring]{\includegraphics[width=.4\columnwidth]{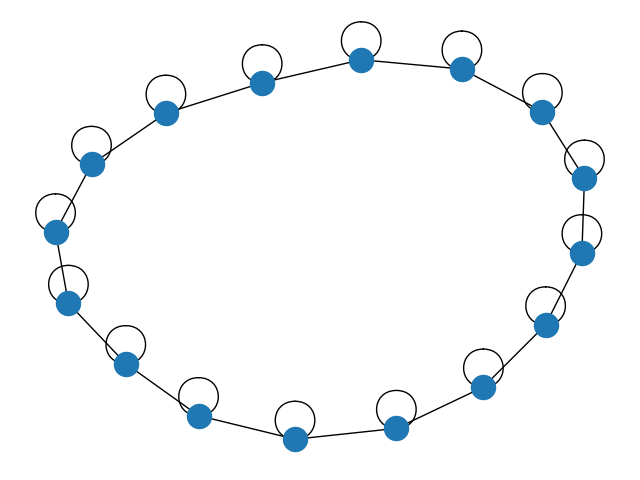}}
\caption{(a) The randomly-generated graph structure; (b) The ring structure.}
\label{graph16}
\end{center}
\vskip -0.1in
\end{figure}

 In this section, we compare the performance of the three distributed algorithms on CIFAR10 and CIFAR100 datasets across different neural network architectures. We consider two distinct training scenarios: (1) Random initialization. In this scenario, all models are initialized with random parameters, ensuring that they are trained from scratch. This setup allows us to assess the algorithms' performance in a general case without any prior knowledge. (2) Pretrained initialization. In the second scenario, the decentralized methods are initialized with models pretrained by the centralized algorithm. This approach investigates how decentralized algorithms perform when starting near a local minimum obtained through  centralized training, emphasizing their ability to converge to a flatter minimum. We provide a brief overview of the experimental setup here, with detailed description available in Appendix \ref{asr_ap}. 
  We choose the ResNet-18 \cite{HeZRS16}, WideResNet-28-10 \cite{zagoruyko2016wide} and DenseNet-121 \cite{HuangLMW17} as the base neural network structures for the two datasets. As for the graph structure of the multi-agent system, we use the Metropolis rule \cite{sayed2014adaptation} to randomly generate a doubly-stochastic graph with 16 nodes, and its structure is shown in Figure \ref{graph16}(a), {and the ring topology shown in Figure \ref{graph16}(b)}.  In the main text, we present the results for ResNet-18, while the results for the other two neural networks are included in Appendix \ref{asr_ap}. In the decentralized experiments, the full training dataset is divided into $K$ subsets, and each agent can only observe one subset. The centralized setting is the same as the traditional single-agent learning.  Moreover, we simulate three different local batch sizes including $B= 128, 256, 512$. Note that the distributed setting with local batch $B$ means the global batch is $KB$.  As for the learning scheme, we rely on the piecewise learning-rate schedule \cite{HeZRS16, LinSPJ20}, where the initial learning rate $0.2$ is divided by $10$ when the training process reaches $50\%$ and $75\%$ of the total epoch. That is, moderately small step sizes are applied first to search for flat minima, and then smaller ones are used to guarantee the stability (i.e., convergence) of algorithms.

\subsection{Random initialization: training from scratch.}\label{randominitial}
\begin{figure*}[ht]
\begin{center}
\subfigure[CIFAR10: B = 128]{\includegraphics[width=.3\linewidth]{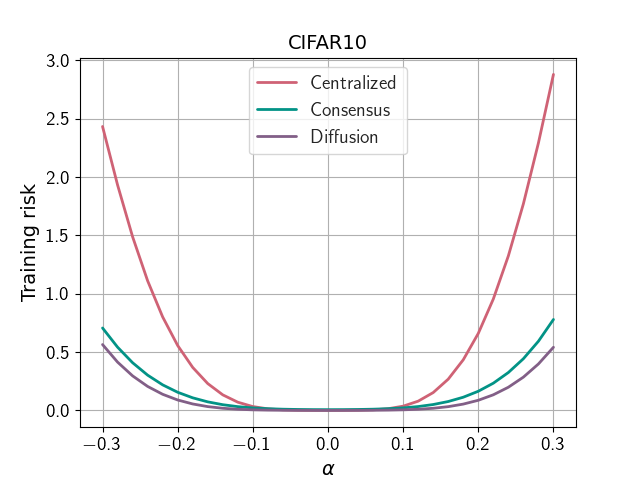}}
\subfigure[CIFAR10: B = 256]{\includegraphics[width=.3\linewidth]{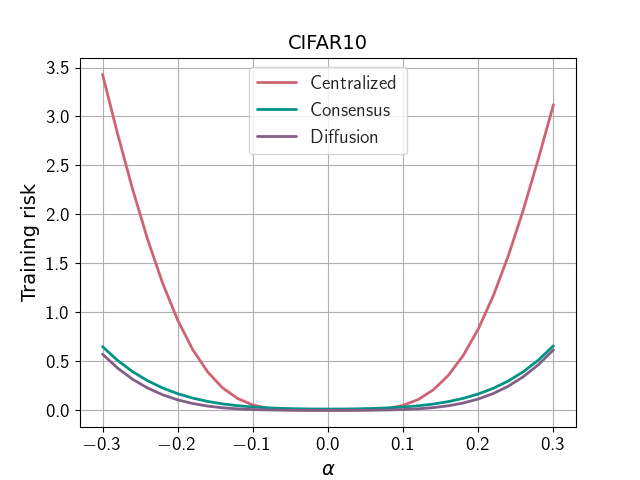}}
\subfigure[CIFAR10: B = 512]{\includegraphics[width=.3\linewidth]{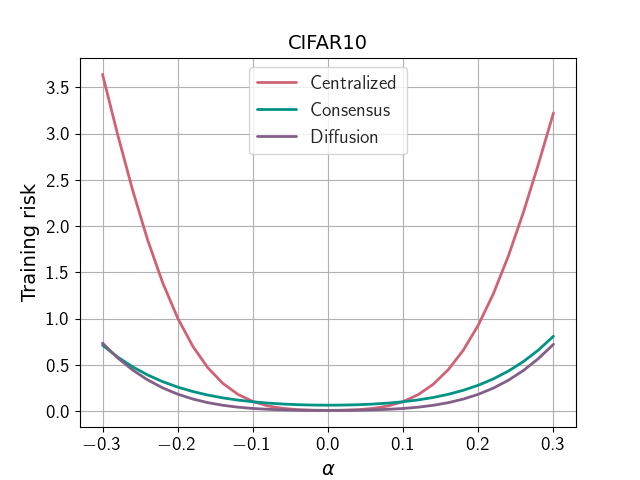}}
\subfigure[CIFAR100: B = 128]{\includegraphics[width=.3\linewidth]{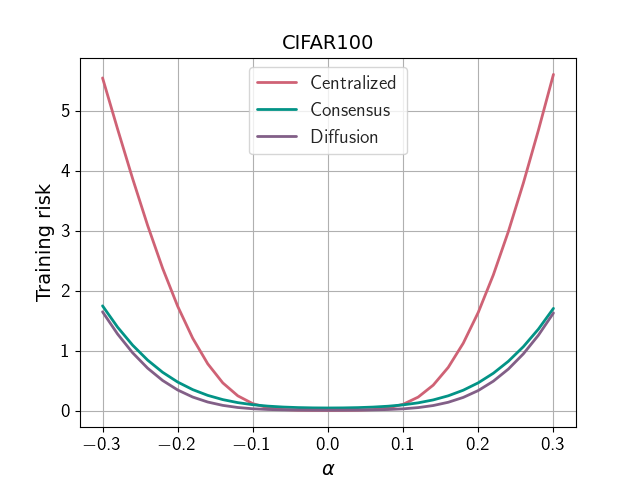}}
\subfigure[CIFAR100: B = 256]{\includegraphics[width=.3\linewidth]{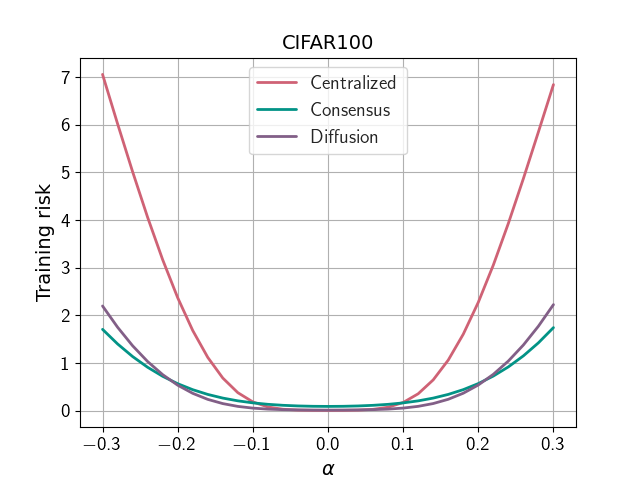}}
\subfigure[CIFAR100: B = 512]{\includegraphics[width=.3\linewidth]{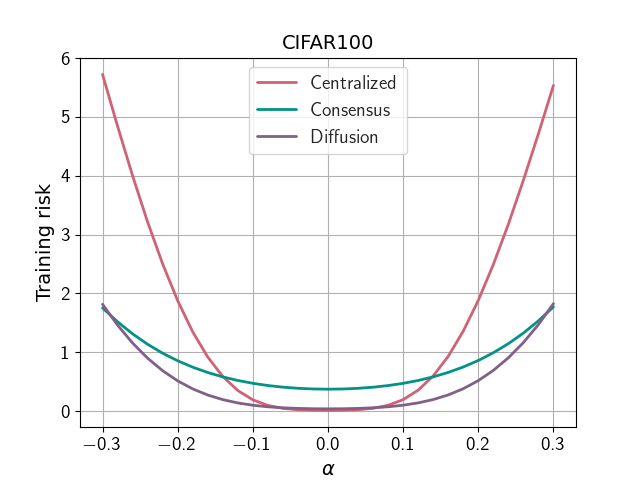}}
\caption{Flatness visualization of models trained from scratch with ResNet-18.  We use the visualization method from \cite{Li0TSG18}, according to which we compute the average training risk value $J(w + \alpha\boldsymbol{v})$ over different random directions $\boldsymbol{v}$ that match the norm of $w$. Wider valleys correspond to flatter minima.}
\label{loss_visual_cifar_scratch}
\end{center}
\end{figure*}

\begin{figure*}[ht]
\begin{center}
\subfigure[CIFAR10: B = 128]{\includegraphics[width=.3\linewidth]{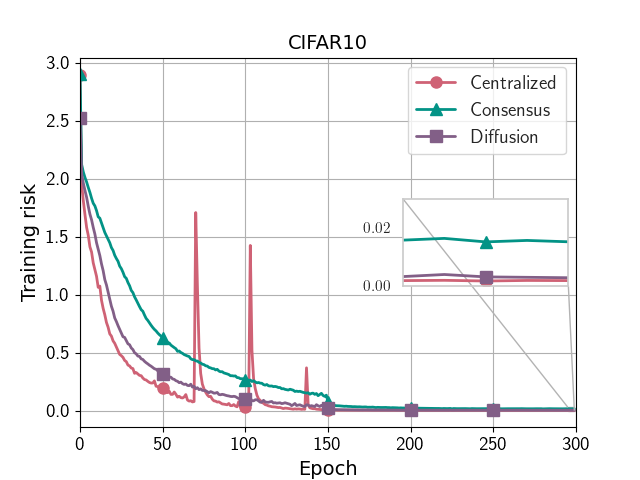}}
\subfigure[CIFAR10: B = 256]{\includegraphics[width=.3\linewidth]{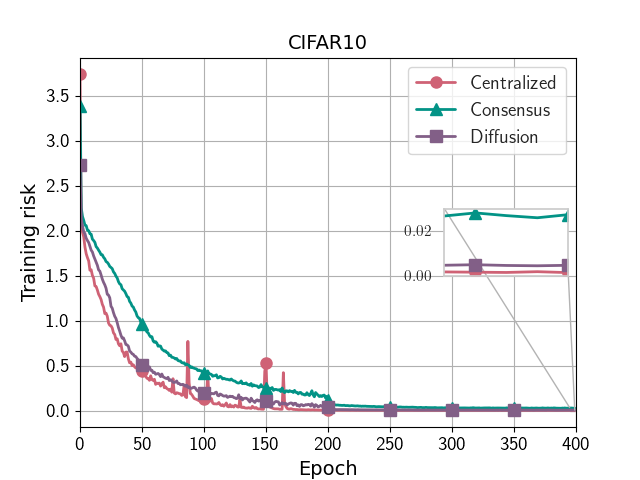}}
\subfigure[CIFAR10: B = 512]{\includegraphics[width=.3\linewidth]{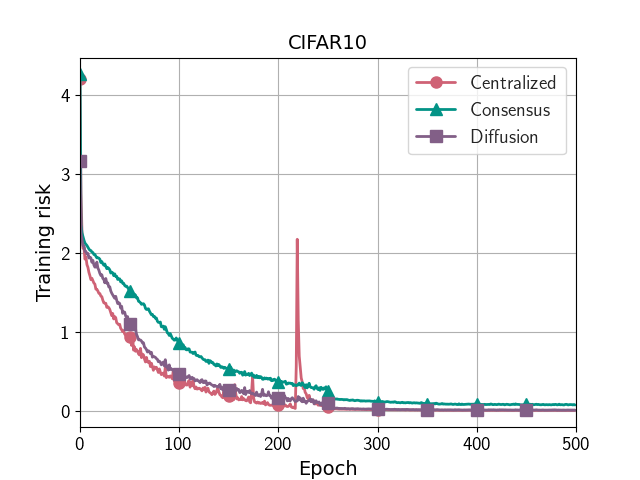}}
\subfigure[CIFAR100: B = 128]{\includegraphics[width=.3\linewidth]{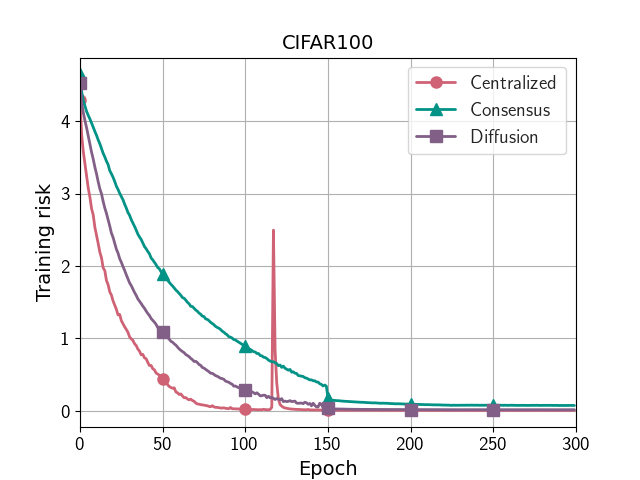}}
\subfigure[CIFAR100: B = 256]{\includegraphics[width=.3\linewidth]{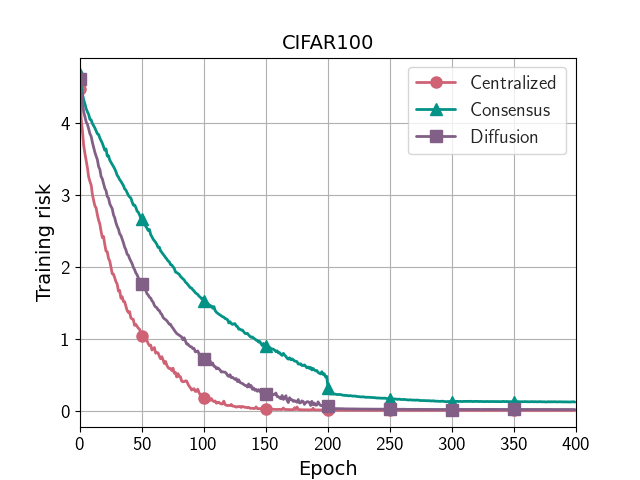}}
\subfigure[CIFAR100: B = 512]{\includegraphics[width=.3\linewidth]{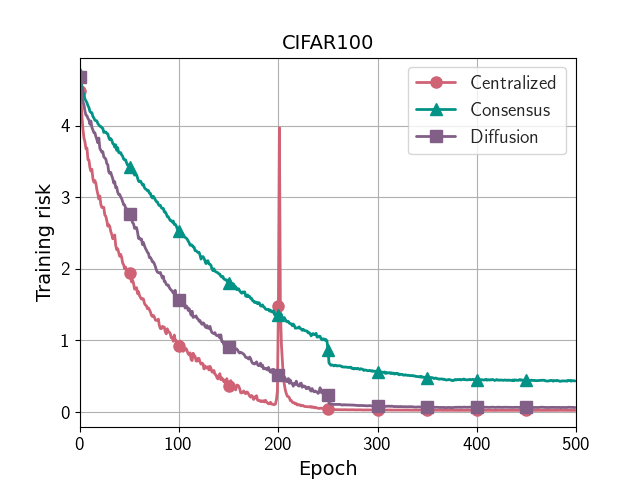}}
\caption{The evolution of training risk with ResNet-18 when training from scratch: (a)--(c) CIFAR10; (d)--(f) CIFAR100.}
\label{curve_cifar10_random_scratch}
\end{center}
\end{figure*}

In this section, we show the results of the first training scenario--random initialization--where all models are trained from scratch. We first illustrate the flatness and optimization performance of the three algorithms on CIFAR10 and CIFAR100. On one hand, we visualize the risk landscape around the obtained models in Figure \ref{loss_visual_cifar_scratch}. To do so, we use the visualization method from \cite{Li0TSG18} where for any model $w$ we compute the risk value $J(w + \alpha\boldsymbol{v})$ using some random directions $\boldsymbol{v}$ that match the norm of $w$. It can be observed from Figure \ref{loss_visual_cifar_scratch} that the decentralized methods arrive at models that are significantly flatter than centralized models. However, the difference in terms of flatness between diffusion and consensus is subtle. In other words, diffusion and consensus converge to models with similar flatness.  On the other hand, the evolution of the training risk is shown in Figure \ref{curve_cifar10_random_scratch}, from which we observe that all three methods converge after sufficient iterations. Also, consensus exhibits larger training risk in the steady state than diffusion especially when the local batch $B$ is $512$, while the optimization performance of diffusion and centralized are comparable. Therefore, diffusion enables flatter models than centralized without obviously sacrificing optimization performance. Note that we also observe some loss spikes in the training loss curves of the centralized method from Figure \ref{curve_cifar10_random_scratch}. One explanation for this phenomenon is that the centralized models may be oscillating around a sharp minimizer that is unstable when using moderately small step sizes. This suggests intuitively that decentralized methods aid in stabilizing the training process.

We next compare the test accuracy of the three methods on CIFAR10 and CIFAR100. To better show the performance of the three methods across different settings, we also simulate the \emph{ring} graph, shown in Figure \ref{graph16}(b), in addition to the random graph structure. Note that the \emph{ring} graph can be viewed as a specific case of a \emph{random} graph generated by the Metropolis rule. Thus the random graph is more general. The simulation results are shown in Table \ref{cad}. In each configuration, we run the simulation 3 times with different seeds, and the final results exhibit the mean of the last 10 epochs in these repetitions. We observe from Table \ref{cad} that diffusion outperforms the other two methods for each configuration when training from scratch. In particular, in the cases of $B = 512$, consensus even performs worse than centralized on the CIFAR100 dataset which is attributed to the optimization issue demonstrated in Figure \ref{curve_cifar10_random_scratch}. 

\begin{table*}[!t]
\footnotesize
\centering
\begin{spacing}{1.6}
\caption{Test accuracy of distributed algorithms trained from scratch on CIFAR10 and CIFAR100 under multiple configurations with ResNet-18.}
\label{cad}
\begin{tabular}{cccccc}
\toprule
Dataset& Method & Graph & $128 \times 16$ & $256 \times 16$ & $512\times16$ \\
\midrule
{\multirow{3}{*}{CIFAR10}} & Centralized & -- & $91.06\pm0.41\%$&$90.36\pm0.16\%$&$89.14\pm0.41\%$\\
\cline{2-6}
& \multirow{2}{*}{Consensus}& Random&$92.11\pm0.01\%$&$91.48\pm0.18\%$&$90.22\pm0.21\%$\\
                          && Ring &$91.67\pm0.15\%$&$91.30\pm0.04\%$&$89.24\pm0.40\%$\\
\cline{2-6}
& \multirow{2}{*}{Diffusion}& Random&$\textbf{92.77}\pm0.01\%$&$\textbf{92.03}\pm0.17\%$&$\textbf{91.49}\pm0.08\%$\\
                          && Ring &$\textbf{92.75}\pm0.20\%$&$\textbf{92.27}\pm0.09\%$&$\textbf{91.32}\pm0.17\%$\\
\midrule
\multirow{3}{*}{CIFAR100} & Centralized & -- & $70.20\pm 0.40\%$&$68.96\pm0.12$\%&$68.65\pm0.37\%$\\
\cline{2-6}
& \multirow{2}{*}{Consensus}& Random& $70.42\pm 0.59\%$&$69.79\pm0.14$\%&$67.38\pm0.45\%$\\
                          && Ring &$70.45\pm 0.67\%$&$68.03\pm 0.18\%$&$64.97\pm0.39\%$\\
\cline{2-6}
& \multirow{2}{*}{Diffusion}& Random&$\textbf{71.32}\pm0.50\%$&$\textbf{70.03}\pm0.28\%$&$\textbf{69.74}\pm0.83\%$\\
                          && Ring &$\textbf{71.30}\pm0.37\%$&$\textbf{69.64}\pm0.58\%$&$\textbf{69.38}\pm0.58\%$\\
\bottomrule
\end{tabular}
\end{spacing}
\end{table*}

\subsection{Pretrained initialization.}

\begin{figure*}[ht]
\begin{center}
\subfigure[CIFAR10: B = 128]{\includegraphics[width=.3\linewidth]{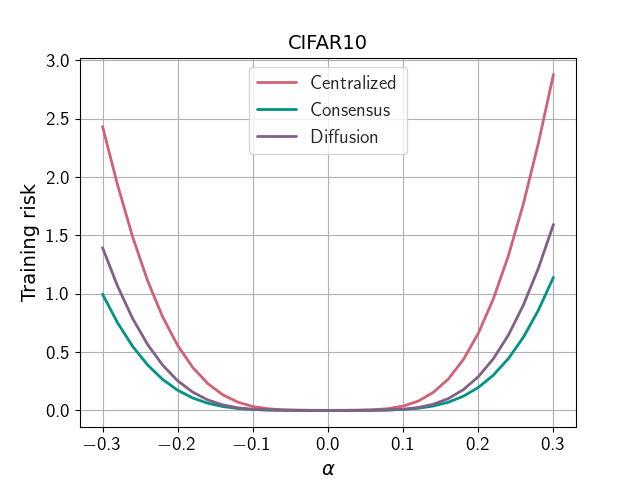}}
\subfigure[CIFAR10: B = 256]{\includegraphics[width=.3\linewidth]{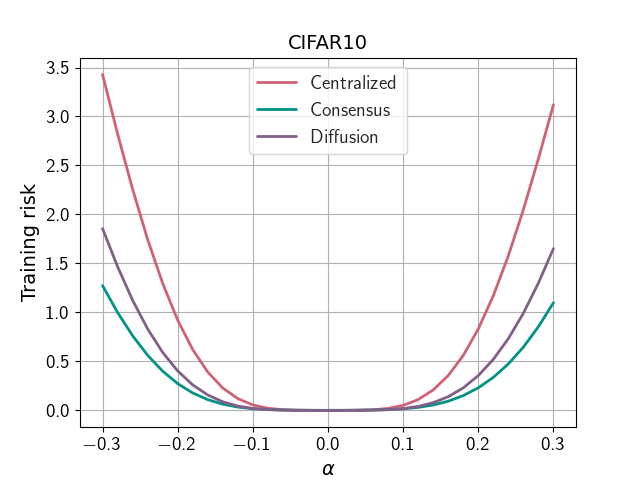}}
\subfigure[CIFAR10: B = 512]{\includegraphics[width=.3\linewidth]{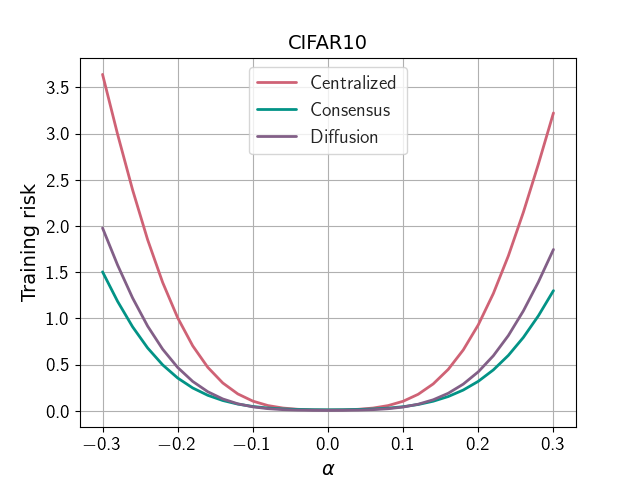}}
\subfigure[CIFAR100: B = 128]{\includegraphics[width=.3\linewidth]{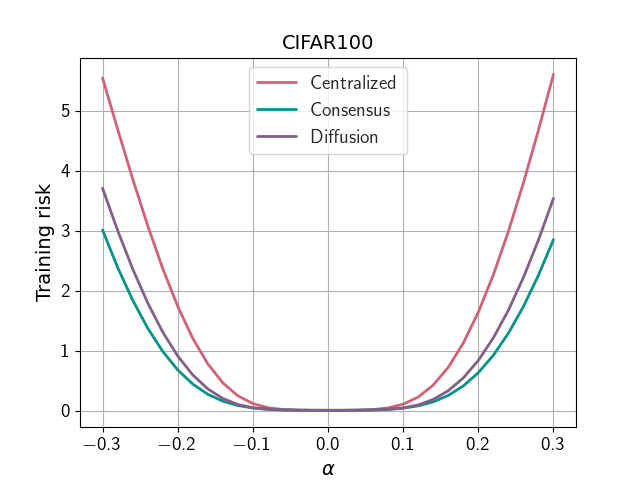}}
\subfigure[CIFAR100: B = 256]{\includegraphics[width=.3\linewidth]{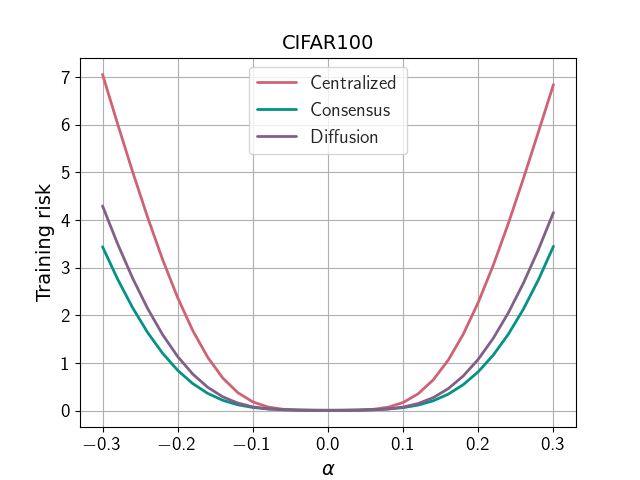}}
\subfigure[CIFAR100: B = 512]{\includegraphics[width=.3\linewidth]{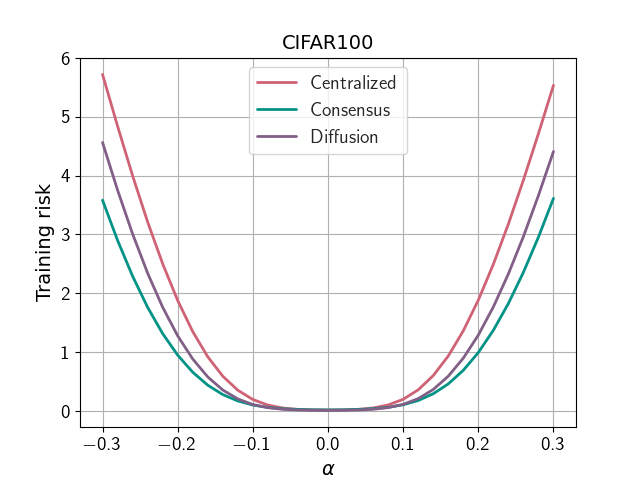}}
\caption{Flatness visualization of models with ResNet-18 using pretrained initializations in decentralized methods. The visulization method is the same with Figure \ref{loss_visual_cifar_scratch}.}
\label{loss_visual_cifar_pretrain}
\end{center}
\end{figure*}

In this section, we present the results of the second scenario, where consensus and diffusion methods are initialized with models pretrained by the centralized algorithm. The flatness of the models where the three methods converge to is visualized in Figure \ref{loss_visual_cifar_pretrain}. Similar to the results from the first scenario shown in Figure \ref{loss_visual_cifar_scratch}, we observe from Figure \ref{loss_visual_cifar_pretrain} that decentralized methods consistently converge to flatter minima than their centralized counterpart. Notably, in this scenario, consensus converges to flatter models than diffusion.

\begin{figure*}[ht]
\begin{center}
\subfigure[CIFAR10: B = 128]{\includegraphics[width=.3\linewidth]{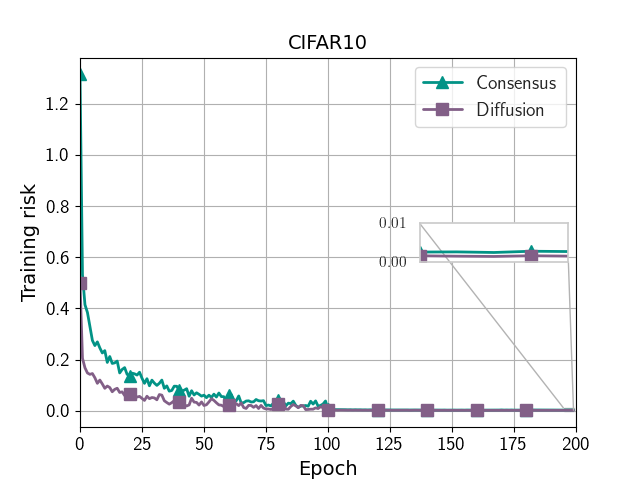}}
\subfigure[CIFAR10: B = 256]{\includegraphics[width=.3\linewidth]{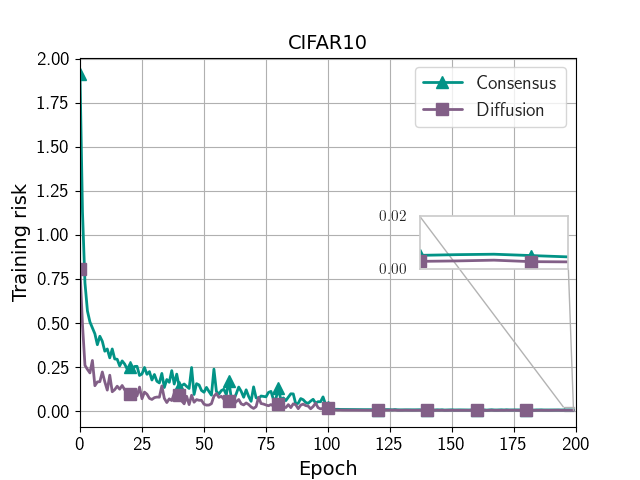}}
\subfigure[CIFAR10: B = 512]{\includegraphics[width=.3\linewidth]{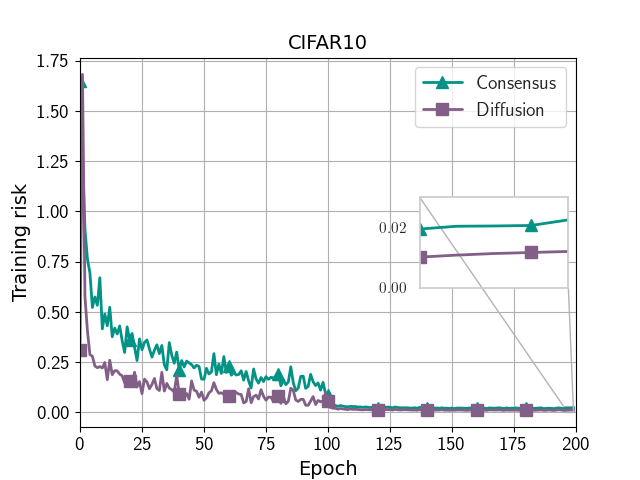}}
\subfigure[CIFAR100: B = 128]{\includegraphics[width=.3\linewidth]{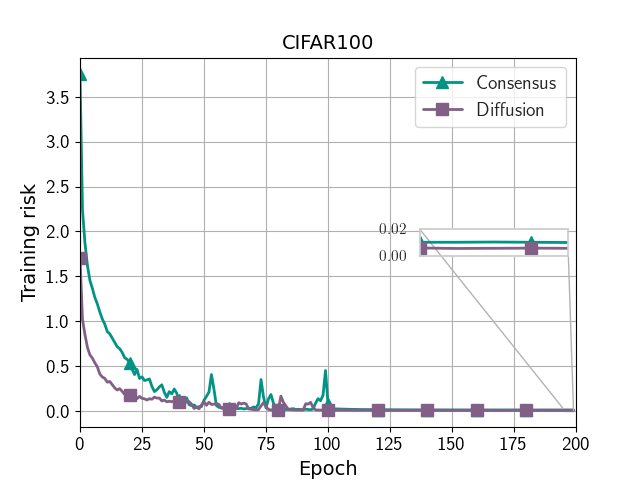}}
\subfigure[CIFAR100: B = 256]{\includegraphics[width=.3\linewidth]{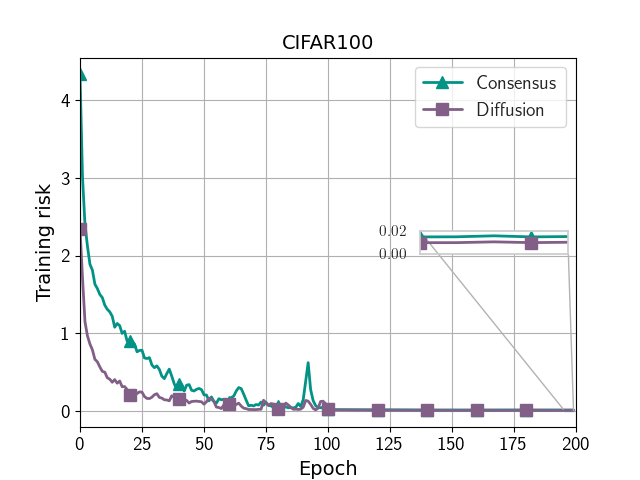}}
\subfigure[CIFAR100: B = 512]{\includegraphics[width=.3\linewidth]{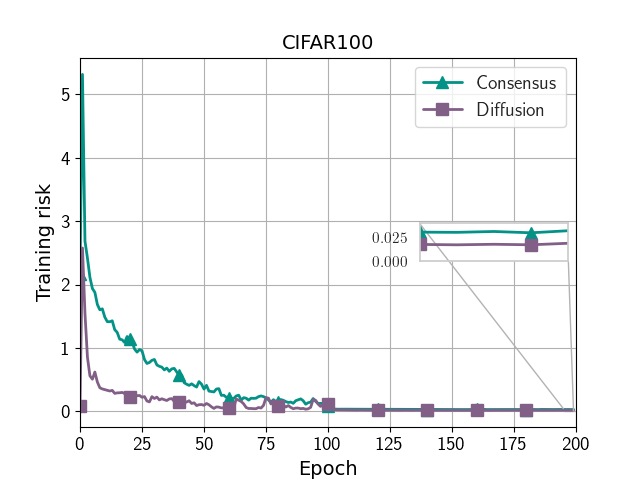}}
\caption{{The evolution of training risk with ResNet-18 using pretrained initializations in decentralized methods: (a)--(c) CIFAR10; (d)--(f) CIFAR100.}}
\label{curve_cifar10_random_pretrain}
\end{center}
\end{figure*}

We then illustrate the optimization performance of consensus and diffusion in Figure \ref{curve_cifar10_random_pretrain}. From this, we observe that although consensus exhibits slightly higher risk values than diffusion, their difference is more subtle compared to the results of the first scenario shown in Figure \ref{curve_cifar10_random_scratch}. This indicates that the optimization issue in the consensus method is alleviated in this scenario.

Finally, we show the test accuracy of the three methods in Table \ref{cad_pretrained}, where the final accuracy is averaged over three repetitions with different random seeds. In this scenario, models obtained by decentralized methods consistently achieve higher accuracy than those trained using the centralized algorithm. Moreover, consensus provides slightly better test accuracy than diffusion, mainly because the pretrained initialization alleviates the effect of the generally worse optimization performance by consensus.

\begin{table*}[!t]
\footnotesize
\centering
\begin{spacing}{1.6}
\caption{{Test accuracy of distributed algorithms with pretrained initializations on CIFAR10 and CIFAR100 under various configurations using ResNet-18.}}
\label{cad_pretrained}
\begin{tabular}{cccccc}
\toprule
Dataset& Method & Graph & $128 \times 16$ & $256 \times 16$ & $512\times16$ \\
\midrule
{\multirow{3}{*}{CIFAR10}} & Centralized & -- & $91.06\pm0.41\%$&$90.36\pm0.16\%$&$89.14\pm0.41\%$\\
\cline{2-6}
& \multirow{2}{*}{Consensus}& Random&$\textbf{92.21}\pm0.36\%$&$\textbf{91.55}\pm0.04\%$&$\textbf{90.57}\pm0.24\%$\\
                          && Ring &$\textbf{92.31}\pm0.37\%$&$\textbf{91.64}\pm0.13\%$&$\textbf{90.63}\pm0.21\%$\\
\cline{2-6}
& \multirow{2}{*}{Diffusion}& Random&$91.79\pm0.30\%$&$91.21\pm0.22\%$&$90.38\pm0.15\%$\\
                          && Ring &$92.03\pm0.36\%$&$91.26\pm0.14\%$&$90.34\pm0.14\%$\\
\midrule
\multirow{3}{*}{CIFAR100} & Centralized & -- & $70.20\pm 0.40\%$&$68.96\pm0.12$\%&$68.65\pm0.37\%$\\
\cline{2-6}
& \multirow{2}{*}{Consensus}& Random& $\textbf{72.08}\pm 0.31\%$&$\textbf{71.19}\pm0.27$\%&$\textbf{70.69}\pm0.33\%$\\
                          && Ring &$\textbf{72.23}\pm 0.30\%$&$\textbf{71.38}\pm 0.43\%$&$\textbf{70.71}\pm0.29\%$\\
\cline{2-6}
& \multirow{2}{*}{Diffusion}& Random&$71.35\pm0.46\%$&$70.24\pm0.28\%$&$69.85\pm0.18\%$\\
                          && Ring &$71.47\pm0.33\%$&$70.43\pm0.53\%$&$69.76\pm0.23\%$\\
\bottomrule
\end{tabular}
\end{spacing}
\end{table*}

\subsection{Discussion}
The simulation results in Table \ref{cad} and \ref{cad_pretrained} can be interpreted by the trade-off between generalization and optimization performance. Inspired by the findings from \cite{Lyu0A22,gatmiry2023inductive, WuS23, ZhuWYWM19, NacsonRSS22} that flatter models tend to generalize better, we present the generalization gap--which measures the difference between the training and test accuracy--in Table \ref{ggtt} and \ref{ggtt_pretrain}. These results demonstrate that flatter models found by decentralized methods result in smaller generalization gap than the centralized approach.

However, we emphasize that the final test accuracy is determined by both generalization and optimization performance. In the scenario of training from scratch,  when $B$ is 512 with CIFAR100 dataset, although consensus achieves better generalization, it simultaneously sacrifices too much optimization performance. Thus, its final test accuracy is worse than the centralized algorithm. Apart from this specific case ($B = 512$ with CIFAR100 dataset), consensus generally achieves a favorable balance than the centralized method, resulting in higher test accuracy.  Similarly, diffusion achieves a more favorable balance between flatness and optimization in the first scenario, thereby exhibiting slightly better test accuracy than the other two methods.

In the second scenario where decentralized methods are initialized at models pretrained by the centralized algorithm, consensus converges to flatter minima than the other two methods. Additionally, its optimization issue, which previously hindered its test accuracy, is alleviated in this scenario due to pretraining. As a result, consensus provides better accuracy than the other two methods.

In summary, decentralized algorithms in general deliver enhanced classification accuracy as they strike a more favorable balance between optimization and generalization compared to the centralized solution. Furthermore, a deeper comparison between consensus and diffusion reveals distinct strengths. In general, consensus demonstrates an advantage in generalization performance, whereas diffusion excels in optimization performance. These methods exhibit different trade-off abilities between optimization and flatness across different training scenarios.

\begin{table*}[t]
\footnotesize
\centering
\caption{{The generalization gap between training and test accuracy with ResNet-18 in the scenario of training from scratch.}}
\label{ggtt}
\begin{spacing}{1.5}
\begin{tabular}{cccccc}
\toprule
Dataset& Method & Graph & $128 \times 16$ & $256 \times 16$ & $512\times16$ \\
\midrule
{\multirow{3}{*}{CIFAR10}} & Centralized & -- & 0.0892&0.0960&0.1053\\
\cline{2-6}
& \multirow{2}{*}{Consensus}& Random& 0.0743&0.0777&0.0686\\
                          && Ring &0.0767&0.0760&0.0583\\
\cline{2-6}
& \multirow{2}{*}{Diffusion}& Random&0.0718&0.0789&0.0805\\
                          && Ring &0.0719&0.0764&0.0798\\
\midrule
\multirow{3}{*}{CIFAR100} & Centralized & -- & 0.2976&0.3099&0.3126\\
\cline{2-6}
& \multirow{2}{*}{Consensus}& Random& 0.2849&0.2817&0.2027\\
                          && Ring &0.2761&0.2764&0.1731\\
\cline{2-6}
& \multirow{2}{*}{Diffusion}& Random&0.2861&0.2989&0.2977\\
                          && Ring &0.2863&0.3029&0.2998 \\
\bottomrule
\end{tabular}
\end{spacing}
\end{table*}

\begin{table*}[t]
\footnotesize
\centering
\caption{The generalization gap between training and test accuracy with ResNet-18 in the scenario of pretrained initialization.}
\label{ggtt_pretrain}
\begin{spacing}{1.5}
\begin{tabular}{cccccc}
\toprule
Dataset& Method & Graph & $128 \times 16$ & $256 \times 16$ & $512\times16$ \\
\midrule
{\multirow{3}{*}{CIFAR10}} & Centralized & -- & 0.0892&0.0960&0.1053\\
\cline{2-6}
& \multirow{2}{*}{Consensus}& Random&0.0776&0.0837&0.0893\\
                          && Ring &0.0766&0.0827&0.0879\\
\cline{2-6}
& \multirow{2}{*}{Diffusion}& Random&0.0819&0.0874&0.0942\\
                          && Ring &0.0795&0.0869&0.0945\\
\midrule
\multirow{3}{*}{CIFAR100} & Centralized & -- & 0.2976&0.3099&0.3126\\
\cline{2-6}
& \multirow{2}{*}{Consensus}& Random& 0.2789&0.2875&0.2910\\
                          && Ring &0.2774&0.2855&0.2895\\
\cline{2-6}
& \multirow{2}{*}{Diffusion}& Random&0.2862&0.2972&0.3006\\
                          && Ring &0.2850&0.2953&0.3016 \\
\bottomrule
\end{tabular}
\end{spacing}
\end{table*}

\section{Conclusion}
We analyzed the learning behavior of three popular algorithms around local minima in nonconvex environments. The results show that higher escaping efficiency enables algorithms to favor flatter minima but comes at the expense of deteriorated optimization performance. In general, decentralized methods exhibit accelerated evasion from local minima in contrast to the centralized strategy, without significantly compromising optimization performance. This allows decentralized methods to achieve higher classification accuracy.
Furthermore, the consensus strategy has a larger excess-risk value than the diffusion one. While this performance measure is “worse” from an optimization perspective, it nevertheless comes with a benefit. It means that consensus has a higher chance of escaping from a local basin and favoring flat minima compared with diffusion.  Consequently, diffusion and consensus exhibit different behavior across different scenarios.   In the typical scenario of training from random initial conditions, diffusion demonstrates better test accuracy due to the more favorable balance between flatness and optimization. However, in the scenario where the decentralized methods are initialized near local minima pretrained by the centralized algorithm, consensus provides better test accuracy than diffusion. Overall, the results in this paper highlight an important trade-off between escaping efficiency and optimization performance in the context of multi-agent learning.

Although we focused on the traditional mini-batch gradient descent implementation, one useful extension would be to consider other types of optimizers, such as stochastic gradient momentum \cite{Yuan0HZPXY21} and ADAM \cite{ZhouF0XHE20}. In addition, it has been observed in the single-agent case that the structure of the gradient noise can influence the escaping efficiency and stability of stochastic gradient algorithms \cite{ZhuWYWM19, WuWS22}. It would be useful to examine the nature of this effect in the multi-agent case. 

\section*{Acknowledgments}
We acknowledge the assistance of ChatGPT in improving the English presentation in this paper. We also thank Dr. Elsa Rizk for her helpful suggestions.



 
%

\bibliographystyle{IEEEtran}
\bibliography{reference} 
\vskip -2\baselineskip plus -1fil 
\begin{IEEEbiography}[{\includegraphics[width=1in,height=1.25in,clip,keepaspectratio]{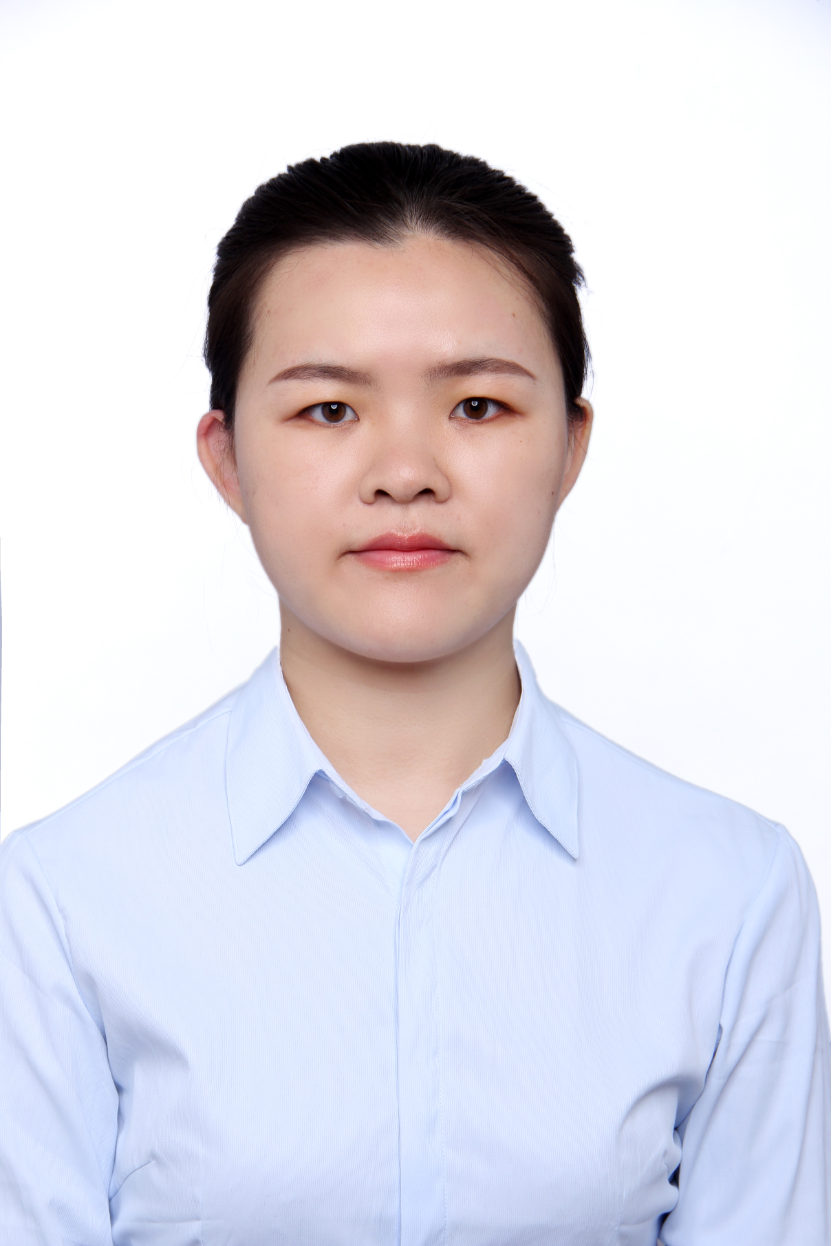}}]{Ying Cao}
is currently pursuing a Ph.D. in the Electrical Engineering Doctoral program of EPFL. She received her bachelor's and master’s degree in Electronic and Information Engineering from Northwestern Polytechnical University, Xi’an, China, in 2017 and 2020, respectively. She received the China National Scholarship in 2018 and 2019. Her research interests include decentralized learning and the robustness of machine learning. 
\end{IEEEbiography}
\vskip -2\baselineskip plus -1fil 
\begin{IEEEbiography}[{\includegraphics[width=1in,height=1.25in,clip,keepaspectratio]{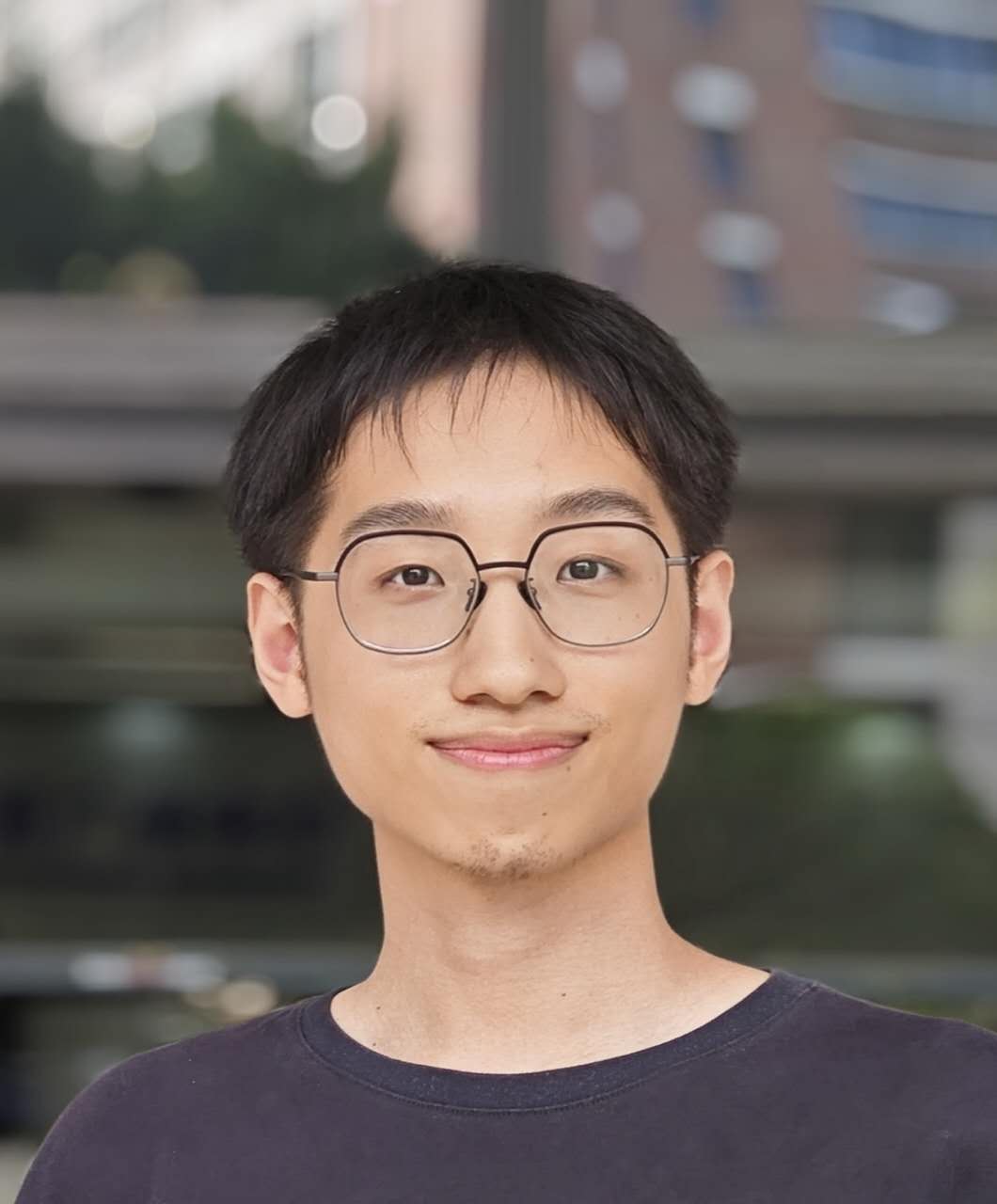}}]{Zhaoxian Wu}
is now pursuing his Ph.D. degree in the Department of Electrical, Computer, and Systems Engineering from Rensselaer Polytechnic Institute, Troy, US. He received his B.E. degree at the School of Data and Computer Science and his M.E degree at the School of Computer Science and Engineering from Sun Yat-sen University, Guangzhou, China, in 2020 and 2023, respectively. He received the China National Scholarship in 2016. His research focuses on the theory and application of optimization and signal processing to problems emerging in machine learning.
\end{IEEEbiography}
\vskip -2\baselineskip plus -1fil 
\begin{IEEEbiography}[{\includegraphics[width=1in,height=1.25in,clip,keepaspectratio]{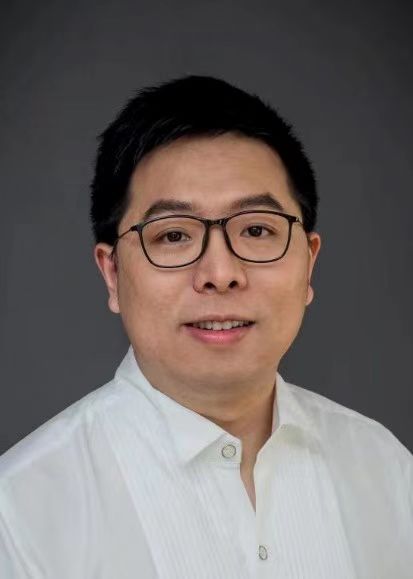}}]{Kun Yuan}
is an Assistant Professor at the Center for Machine Learning Research (CMLR) at Peking University. He completed his Ph.D. in Electrical and Computer Engineering at UCLA in 2019 and subsequently worked as a staff algorithm engineer at Alibaba (US) Group from 2019 to 2022. His research concentrates on the theoretical and algorithmic foundations of optimization, signal processing, machine learning, and data science. Dr. Yuan is a recipient of the 2017 IEEE Signal Processing Society Young Author Best Paper Award and the 2017 ICCM Distinguished Paper Award.
\end{IEEEbiography}
\vskip -2\baselineskip plus -1fil 
\begin{IEEEbiography}[{\includegraphics[width=1in,height=1.25in,clip,keepaspectratio]{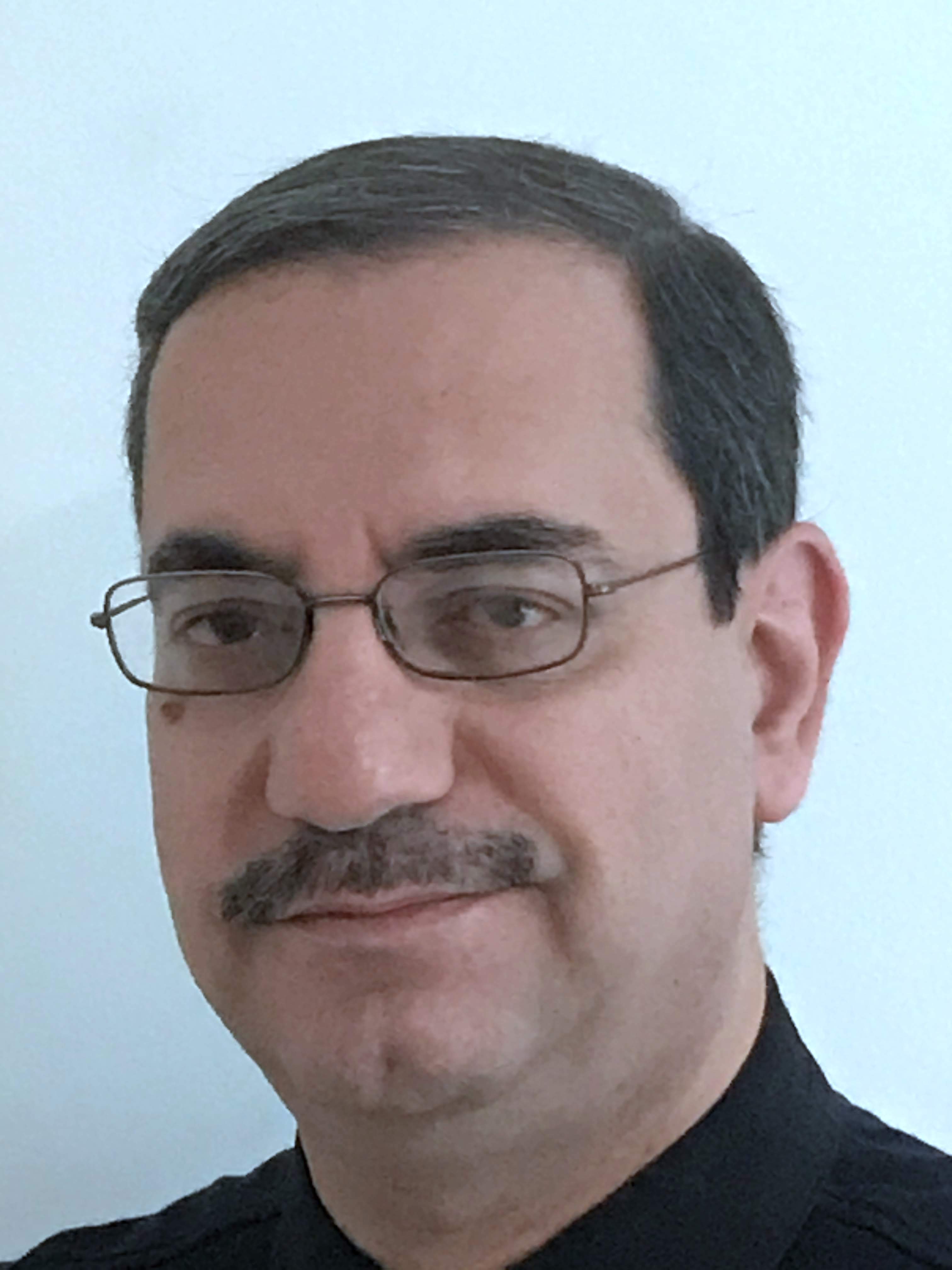}}]{Ali H. Sayed}
is Dean of Engineering at EPFL, Switzerland, where he also directs the Adaptive Systems Laboratory. He served before as Distinguished Professor and Chair of Electrical Engineering at UCLA. He is a member of the US National Academy of Engineering (NAE) and The World Academy of Sciences (TWAS). He served as President of the IEEE Signal Processing Society in 2018 and 2019. An author of over 650 scholarly publications and 10 books, his research involves several areas including adaptation and learning theories,  statistical inference, and multi-agent systems. His work has been recognized with several awards including more recently the 2022 IEEE Fourier Technical Field Award and the 2020 IEEE Wiener Society Award. He is a Fellow of IEEE, EURASIP, and the American Association for the Advancement of Science (AAAS).
\end{IEEEbiography}

\newpage
\appendices
\onecolumn
\begin{center}
   {\huge{On the Trade-off between Flatness and Optimization \\ in Distributed Learning }}
\end{center}
\begin{center}
    Ying Cao*, Zhaoxian Wu, Kun Yuan, Ali H. Sayed, \IEEEmembership{Fellow,~IEEE}\footnote{*\ Corresponding author. Authors Ying Cao and Ali H. Sayed are with the Institute of Electrical and Micro Engineering in EPFL,
Lausanne. Zhaoxian Wu was with the Sun Yat-sen University. Kun Yuan is with the Center for Machine Learning Research (CMLR) in Peking University.
Emails: \{ying.cao, ali.sayed\}@epfl.ch, wuzhx23@mail2.sysu.edu.cn, kunyuan@pku.edu.cn.}

\end{center}

\section{Poofs for Lemma \ref{lm_gn}: Properties associated with gradient noise}\label{pfgs}
Proofs in this section follow from \cite{sayed2014adaptation,kayaalp2022dif, sayed_2023}. We first note that the mini-batch gradient is an unbiased estimator of the true gradient. That is, for any $\boldsymbol{w} \in \mathcal{F}_{n-1}$:
\begin{align}\label{s0}
\mathds{E}\left[s_{k,n}^B(\boldsymbol{w})\vert\mathcal{F}_{n-1}\right] &= \frac{1}{B}\sum\limits_b \mathds{E}\left[ \nabla Q_k(\boldsymbol{w};\boldsymbol{x}_{k,n}^b) - \nabla J_{k}(\boldsymbol{w})\vert\mathcal{F}_{n-1} \right] = 0
\end{align}
Using (\ref{s0}) and the independence of data among agents, for any two agents $k$ and $\ell$, and $\boldsymbol{w}_k,  \boldsymbol{w}_\ell \in \mathcal{F}_{n-1}$, we have
\begin{equation}\label{sk_sl_0}
    \mathds{E}\left[\boldsymbol{s}_{k,n}(\boldsymbol{w}_k)\boldsymbol{s}_{\ell,n}^{\sf  T}(\boldsymbol{w}_\ell)\vert\mathcal{F}_{n-1}\right] = \mathds{E}\left[\boldsymbol{s}_{k,n}(\boldsymbol{w}_k)\vert\mathcal{F}_{n-1}\right]\times\mathds{E}\left[\boldsymbol{s}_{\ell,n}^{\sf T}(\boldsymbol{w}_\ell)\vert\mathcal{F}_{n-1}\right] = 0
\end{equation}
Then, for $B = 1$ we have
\begin{align}\label{sb12}
     \mathds{E}\left[\Vert \boldsymbol{s}_{k,n}(\boldsymbol{w})\Vert^2\vert\mathcal{F}_{n-1}\right]  = &  \mathds{E} \left[ \Vert\nabla Q_k(\boldsymbol{w};\boldsymbol{x}_{k,n}) - \nabla J_k(\boldsymbol{w})\Vert^2\vert\mathcal{F}_{n-1}\right]\notag\\
      \overset{(a)}{\le}& 2\mathds{E}\Vert\nabla Q_k(\boldsymbol{w};\boldsymbol{x}_{k,n})  - \nabla Q_k(w^\star;\boldsymbol{x}_{k,n})\Vert^2 +4\mathds{E}\Vert \nabla Q_k(w^\star;\boldsymbol{x}_{k,n})\Vert^2+8\Vert\nabla J_k(\boldsymbol{w}) - \nabla J_k(w^\star)\Vert^2\notag\\
      &+ 8\Vert\nabla J_k(w^\star)\Vert^2\notag\\
        \overset{(b)}{\le}& 10\mathds{E}\Vert\nabla Q_k(\boldsymbol{w};\boldsymbol{x}_{k,n})  - \nabla Q_k(w^\star;\boldsymbol{x}_{k,n})\Vert^2 + 12\mathds{E}\Vert \nabla Q_k(w^\star;\boldsymbol{x}_{k,n})\Vert^2\notag\\
        \overset{(c)}{\le}& 10L^2\Vert \boldsymbol{w} - w^\star\Vert^2 + 12\mathds{E}\Vert \nabla Q_k(w^\star;\boldsymbol{x}_{k,n})\Vert^2 \overset{\Delta}{=} a_1
\end{align}
where $(a)$ follows from Jensen's inequality, $(b)$ follows from the following inequality:
\begin{align}
\Vert\nabla J_k(w^\star)\Vert^2 = \Vert\mathds{E} \nabla Q_k(w^\star;\boldsymbol{x}_{k,n})\Vert^2\le \mathds{E} \Vert \nabla Q_k(w^\star;\boldsymbol{x}_{k,n})\Vert^2
\end{align}
\begin{align}
\Vert\nabla J_k(\boldsymbol{w}) - \nabla J_k(w^\star)\Vert^2 = \Vert \mathds{E} \nabla Q_k(\boldsymbol{w};\boldsymbol{x}_{k,n}) - \mathds{E} \nabla Q_k(w^\star;\boldsymbol{x}_{k,n})\Vert^2 \le \mathds{E}\Vert \nabla Q_k(\boldsymbol{w};\boldsymbol{x}_{k,n}) - \nabla Q_k(w^\star;\boldsymbol{x}_{k,n})\Vert^2
\end{align}
and $(c)$ follows from the Lipschitz condition in Assumption \ref{ass_smooth}. Similarly, for the fourth-order moment of the stochastic gradient noise, we have:
\begin{align}\label{sb14}
  &\; \mathds{E}\left[\Vert \boldsymbol{s}_{k,n}(\boldsymbol{w})\Vert^4\vert\mathcal{F}_{n-1}\right]\notag\\ 
  &= \mathds{E} \left[\Vert\nabla Q_k(\boldsymbol{w};\boldsymbol{x}_{k,n}) - \nabla J_k(\boldsymbol{w})\Vert^4\vert\mathcal{F}_{n-1}\right]   \notag\\
  &\le 8\mathds{E}\Vert\nabla Q_k(\boldsymbol{w};\boldsymbol{x}_{k,n})  - \nabla Q_k(w^\star;\boldsymbol{x}_{k,n})\Vert^4 + 64\mathds{E}\Vert \nabla Q_k(w^\star;\boldsymbol{x}_{k,n})\Vert^4 + 512\mathds{E}\Vert\nabla J_k(\boldsymbol{w}) - \nabla J_k(w^\star)\Vert^4 +512\mathds{E}\Vert\nabla J_k(w^\star)\Vert^4\notag\\
  &\le  520 L^4\Vert \boldsymbol{w} - w^\star\Vert^4 + 576\mathds{E}\Vert \nabla Q_k(w^\star;\boldsymbol{x}_{k,n})\Vert^4 \overset{\Delta}{=} a_2
\end{align}

We now verify that the size of terms related to gradient noise decreases with the increase of batch size. For any $\boldsymbol{w} \in \mathcal{F}_{n-1}$, we verify how the second and fourth-order conditional gradient noise varies with batch size. Recall (\ref{sb12}) and (\ref{sb14}), for $B=1$:
\begin{align}
    \mathds{E}[\Vert \boldsymbol{s}_{k,n}(\boldsymbol{w})\Vert^2\vert\mathcal{F}_{n-1} ] \le a_1,\quad \mathds{E}[\Vert \boldsymbol{s}_{k,n}(\boldsymbol{w})\Vert^4\vert\mathcal{F}_{n-1} ] \le a_2
\end{align}
where $a_1^2\le a_2$, while for general $B>1$:
\begin{align}\label{sb2l}
    {\mathds{E}[\Vert \boldsymbol{s}_{k,n}^B(\boldsymbol{w})\Vert^2\vert\mathcal{F}_{n-1} ]} =& \mathds{E}\left[\left\Vert\frac{1}{B}\sum\limits_b\nabla Q_k(\boldsymbol{w};\boldsymbol{x}_{k,n}^b) - \nabla J_{k}(\boldsymbol{w})\right\Vert^2\Bigg\vert\mathcal{F}_{n-1}\right] \notag\\
    \overset{(a)}{=}&\frac{1}{B^2}\sum\limits_b\mathds{E}[\Vert\nabla Q_k(\boldsymbol{w};\boldsymbol{x}_{k,n}^b) - \nabla J_{k}(\boldsymbol{w})\Vert^2\vert\mathcal{F}_{n-1} ]\notag\\
    =& \frac{1}{B}\mathds{E}[\Vert \boldsymbol{s}_{k,n}(\boldsymbol{w})\Vert^2\vert\mathcal{F}_{n-1}]\le \frac{1}{B} a_1
\end{align}
where $(a)$ follows from the independence among data. As for the  fourth-order conditional gradient noise, we prove it by induction. Assume
\begin{align}
   \mathds{E}[\Vert \boldsymbol{s}_{k,n}^{B-1}(\boldsymbol{w})\Vert^4\vert\mathcal{F}_{n-1}]  \le \frac{3a_2}{(B-1)^2}
\end{align}
then we have
\begin{align}\label{sb4l}
   &\mathds{E}[\Vert \boldsymbol{s}_{k,n}^{B}(\boldsymbol{w})\Vert^4\vert\mathcal{F}_{n-1}]   = \mathds{E}\left[\left\Vert \frac{B-1}{B}\cdot\frac{1}{B-1}\sum\limits_{b=1}^{B}\nabla Q_k(\boldsymbol{w};\boldsymbol{x}_{k,n}^b) - \nabla J_k(\boldsymbol{w})\right\Vert^4\Bigg\vert\mathcal{F}_{n-1} \right]\notag\\
   =&\mathds{E}\left\Vert \frac{B-1}{B}\cdot\frac{1}{B-1}\sum\limits_{b=1}^{B-1}\left(\nabla Q_k(\boldsymbol{w};\boldsymbol{x}_{k,n}^b) - \nabla J_k(\boldsymbol{w})\right)+\frac{1}{B}\Big(\nabla Q_k(\boldsymbol{w};\boldsymbol{x}_{k,n}) -  \nabla J_k(\boldsymbol{w})\Big)\right\Vert^4\notag\\
   =&\mathds{E}\left(\left\Vert \frac{B-1}{B}\cdot\frac{1}{B-1}\sum\limits_{b=1}^{B-1}\left(\nabla Q_k(\boldsymbol{w};\boldsymbol{x}_{k,n}^b) - \nabla J_k(\boldsymbol{w})\right) +\frac{1}{B}\Big(\nabla Q_k(\boldsymbol{w};\boldsymbol{x}_{k,n}) -  \nabla J_k(\boldsymbol{w})\Big)\right\Vert^2\right)^2\notag\\
   =&\mathds{E}\left\Vert \frac{B-1}{B}\cdot\frac{1}{B-1}\sum\limits_{b=1}^{B-1}\left(\nabla Q_k(\boldsymbol{w};\boldsymbol{x}_{k,n}^b) - \nabla J_k(\boldsymbol{w})\right) \right\Vert^4 + \mathds{E}\left\Vert\frac{1}{B}\Big(\nabla Q_k(\boldsymbol{w};\boldsymbol{x}_{k,n}) -  \nabla J_k(\boldsymbol{w})\Big)\right\Vert^4 \notag\\
   &\ +4\mathds{E}\left[\left(\frac{B-1}{B}\cdot\frac{1}{B-1}\sum\limits_{b=1}^{B-1}\left(\nabla Q_k(\boldsymbol{w};\boldsymbol{x}_{k,n}^b) - \nabla J_k(\boldsymbol{w})\right)\right)^{\sf T}\left( \frac{1}{B}\Big(\nabla Q_k(\boldsymbol{w};\boldsymbol{x}_{k,n}) -  \nabla J_k(\boldsymbol{w})\Big)\right)\right]^2 \notag\\
   &\ + 2\mathds{E}\left\Vert \frac{B-1}{B}\cdot\frac{1}{B-1}\sum\limits_{b=1}^{B-1}\left(\nabla Q_k(\boldsymbol{w};\boldsymbol{x}_{k,n}^b) - \nabla J_k(\boldsymbol{w})\right)\right\Vert^2\left\Vert\frac{1}{B}\Big(\nabla Q_k(\boldsymbol{w};\boldsymbol{x}_{k,n}) -  \nabla J_k(\boldsymbol{w})\Big)\right\Vert^2\notag\\
   \le & \mathds{E}\left\Vert \frac{B-1}{B}\cdot\frac{1}{B-1}\sum\limits_{b=1}^{B-1}\left(\nabla Q_k(\boldsymbol{w};\boldsymbol{x}_{k,n}^b) - \nabla J_k(\boldsymbol{w})\right) \right\Vert^4 + \mathds{E}\left\Vert\frac{1}{B}\Big(\nabla Q_k(\boldsymbol{w};\boldsymbol{x}_{k,n}) -  \nabla J_k(\boldsymbol{w})\Big)\right\Vert^4  \notag\\
   &\ + 6\mathds{E}\left\Vert \frac{B-1}{B}\cdot\frac{1}{B-1}\sum\limits_{b=1}^{B-1}\left(\nabla Q_k(\boldsymbol{w};\boldsymbol{x}_{k,n}^b) - \nabla J_k(\boldsymbol{w})\right)\right\Vert^2\mathds{E}\left\Vert\frac{1}{B}\Big(\nabla Q_k(\boldsymbol{w};\boldsymbol{x}_{k,n}) -  \nabla J_k(\boldsymbol{w})\Big)\right\Vert^2\notag\\
   \le &\frac{(B-1)^4}{B^4}\mathds{E}[\Vert \boldsymbol{s}_{k,n}^{B-1}(\boldsymbol{w})\Vert^4\vert\mathcal{F}_{n-1}] + \frac{a_2}{B^4} + 6\left(\frac{(B-1)^2}{B^4}\times\frac{1}{B-1}a_1^2\right)   \notag\\
    \le &  \frac{3(B-1)^2a_2}{B^4} + \frac{a_2}{B^4} + 6\frac{(B-1)}{B^4}a_2 = \frac{(3B^2 - 2)a_2}{B^4} \le \frac{3a_2}{B^2}
\end{align}

Combining (\ref{sb12}), (\ref{sb14}), (\ref{sb2l}) and (\ref{sb4l}), we can verify that the variance and fourth-order moment of the gradient noise is upper bounded by terms related to batch size $B$ and  the difference between the current model and the local minimizer denoted by $\boldsymbol{w} - w^\star$:
\begin{align}
\label{f_s2}
    \mathds{E}[\Vert \boldsymbol{s}_{k,n}^{B}(\boldsymbol{w})\Vert^2\vert\mathcal{F}_{n-1}]&\le O\left(\frac{1}{B}\right)\mathds{E}\Vert \boldsymbol{w} - w^\star\Vert^2  + O\left(\frac{1}{B}\right)\\
    \label{f_s4}
      \mathds{E}[\Vert \boldsymbol{s}_{k,n}^{B}(\boldsymbol{w})\Vert^4\vert\mathcal{F}_{n-1} ] & \le O\left(\frac{1}{B^2}\right)\mathds{E}\Vert \boldsymbol{w} - w^\star\Vert^4  + O\left(\frac{1}{B^2}\right)
\end{align}

As for the gradient covariance matrices, we have
\begin{align}\label{Rskb1}
    R_{s,k,n}^{B}(w) = &\mathds{E}\left\{ \left[\frac{1}{B}\sum\limits_{i}\nabla Q_k(w;\boldsymbol{x}_{k,n}^{i}) - \nabla J_k(w) \right]\left[\frac{1}{B}\sum\limits_{i}\nabla Q_k(w;\boldsymbol{x}_{k,n}^{i}) - \nabla J_k(w) \right]^{\sf T}\right\}\notag\\
    = &\frac{1}{B^2}\mathds{E}\left\{\sum\limits_{i}\sum\limits_{j}\left[\nabla Q_k(w;\boldsymbol{x}_{k,n}^{i}) - \nabla J_k(w)\right]\left[\nabla Q_k(w;\boldsymbol{x}_{k,n}^{j}) - \nabla J_k(w)\right]^{\sf T}\right\}\notag\\
 \overset{(a)}{=}&\frac{1}{B^2}\mathds{E}\left\{\sum\limits_{i} \left[\nabla Q_k(w;\boldsymbol{x}_{k,n}^{i}) - \nabla J_k(w)\right]\left[\nabla Q_k(w;\boldsymbol{x}_{k,n}^{i}) - \nabla J_k(w)\right]^{\sf T}\right\}\notag\\
 = & \frac{1}{B}\mathds{E}\left[\nabla Q_k(w;\boldsymbol{x}_{k}) - \nabla J_k(w)\right]\left[\nabla Q_k(w;\boldsymbol{x}_{k}) - \nabla J_k(w)\right]^{\sf T} = \frac{1}{B}R_{s,k,n}(w)
\end{align}
where $(a)$ is due to the fact that all data are sampled independently at each agent.

\section{Proof for Lemmas \ref{mse} and \ref{mse_centra}: {Upper bound for the second-order moment}}\label{mse2}
The proof in this section is similar to \cite{sayed2014adaptation, ChenS15a, ChenS15b} with 2 important differences: we now focus on nonconvex (as opposed to convex) objective functions and on the finite-horizon (as opposed to infinite-horizon) case. To examine the size of $\mathrm{ER}_n$, we first analyze the mean-square distance associated with  ${\widetilde{\boldsymbol{\scriptstyle\mathcal{W}}}}_{n}$ in (\ref{uni_dis}) denoted by $\mathds{E}\Vert{\widetilde{\boldsymbol{\scriptstyle\mathcal{W}}}}_{n}\Vert^2$ for later use, which can be verified  that is upper bounded by terms associated with $\mu$ and gradient noise. 

Since
\begin{align}\label{vvt}
    VV^{\sf T}=\left[\frac{1}{\sqrt{K}}\mathbbm{1} \quad V_\alpha\right]\left[
    \begin{array}{c}
    \frac{1}{\sqrt{K}}\mathbbm{1}^{\sf T}\\
    V_{\alpha}^{\sf T}
    \end{array}
    \right] = V^{\sf T}V = \left[
    \begin{array}{c}
    \frac{1}{\sqrt{K}}\mathbbm{1}^{\sf T}\\
    V_{\alpha}^{\sf T}
    \end{array}
    \right]\left[\frac{1}{\sqrt{K}}\mathbbm{1} \quad V_\alpha\right] = I_{K}
\end{align}
we have
\begin{align}\label{a_pro}
    \frac{1}{K}\mathbbm{1}\mathbbm{1}^{\sf  T} + V_\alpha V_\alpha^{\sf T} = I_K, \quad \mathbbm{1}^{\sf T}V_\alpha = 0,\quad V_\alpha^{\sf T}V_\alpha = I_{K-1}
\end{align}
Also, by the mean-value theorem \cite{sayed2014adaptation}, we have
\begin{align}\label{mvt}
    \nabla J_k(w^\star) - \nabla J_k(\boldsymbol{w}_{k,n}) = H_{k,n}(\boldsymbol{w}_{k,n})\tilde{\boldsymbol{w}}_{k,n}
\end{align}
Note that for the heterogeneous networks, we have 
\begin{equation}\label{g0}
    \nabla J_k(w^\star)\neq 0, \quad \nabla J(w^\star) = \frac{1}{K}\sum\limits_k  \nabla J_k(w^\star)= 0
\end{equation}
Recall the error recursion in (\ref{uni_dis}), since $A_1$ and $A_2$ are either $A$ or $I_K$, we have
\begin{align}\label{p_a1a2}
    A_1A_2 = A, \; A_1\mathbbm{1} = A_2\mathbbm{1} = \mathbbm{1}
\end{align}
from which  we have
\begin{align}\label{uni_dis_2}
    {\widetilde{\boldsymbol{\scriptstyle\mathcal{W}}}}_{n} &= \mathcal{A}_2^{}(\mathcal{A}_1^{} - \mu\mathcal{H}_{n-1}){\widetilde{\boldsymbol{\scriptstyle\mathcal{W}}}}_{n-1}  + \mu \mathcal{A}_2^{} d + \mu \mathcal{A}_2\boldsymbol{s}_n^B \notag\\
    & = (\mathcal{A}- \mu\mathcal{A}_2\mathcal{H}_{n-1}){\widetilde{\boldsymbol{\scriptstyle\mathcal{W}}}}_{n-1}  + \mu \mathcal{A}_2 d + \mu \mathcal{A}_2\boldsymbol{s}_n^B \notag\\
    &\overset{(a)}{=} \mathcal{V}(\mathcal{P} - \mu\mathcal{V}^{\sf T}\mathcal{A}_2\mathcal{H}_{n-1}\mathcal{V})\mathcal{V}^{\sf T}{\widetilde{\boldsymbol{\scriptstyle\mathcal{W}}}}_{n-1} + \mu \mathcal{A}_2 d + \mu \mathcal{A}_2\boldsymbol{s}_n^B
\end{align}
where $(a)$ follows from (\ref{decomp_A}). Multiplying (\ref{uni_dis_2}) from the left by $\mathcal{V}^{\sf T}$, the following recursion can be obtained:
\begin{align}\label{diff_2}
     \mathcal{V}^{\sf T}{\widetilde{\boldsymbol{\scriptstyle\mathcal{W}}}}_{n} =& \left[\begin{array}{c}
         (\frac{1}{\sqrt{K}}\mathbbm{1}^{\sf T} \otimes I_M){\widetilde{\boldsymbol{\scriptstyle\mathcal{W}}}}_{n} \\
         (V_\alpha^{\sf T}\otimes I_M){\widetilde{\boldsymbol{\scriptstyle\mathcal{W}}}}_{n}
    \end{array}\right] \overset{\Delta}{=} \left[\begin{array}{c}
         \bar{\boldsymbol{w}}_n  \\
          \check{\boldsymbol{w}}_n
    \end{array}\right] \notag\\
    =& (\mathcal{P} - \mu \mathcal{V}^{\sf T}\mathcal{A}_2\mathcal{H}_{n-1}\mathcal{V})\mathcal{V}^{\sf T}{\widetilde{\boldsymbol{\scriptstyle\mathcal{W}}}}_{n-1} + \mu \mathcal{V}^{\sf T}\mathcal{A}_2d + \mu\mathcal{V}^{\sf T}\mathcal{A}_2\boldsymbol{s}_{n}^{B}\notag\\
    =&  \left(\left[\begin{array}{cc}
        I_M & \boldsymbol{0} \\
        \boldsymbol{0} & P_{\alpha}\otimes I_M
    \end{array}\right] - \mu\left[\begin{array}{c}\frac{1}{\sqrt{K}}\mathbbm{1}^{\sf T}\otimes I_{M}\\
    V_{\alpha}^{\sf T}\otimes I_{M}
    \end{array}\right]\mathcal{A}_2\mathcal{H}_{n-1}\left[\frac{1}{\sqrt{K}}\mathbbm{1}\otimes I_M \quad V_\alpha\otimes{I_M}\right]\right) \left[\begin{array}{c}
         \bar{\boldsymbol{w}}_{n-1}  \\
          \check{\boldsymbol{w}}_{n-1}
    \end{array}\right] \notag\\
    & + \mu \left[\begin{array}{c}\frac{1}{\sqrt{K}}\mathbbm{1}^{\sf T}\otimes I_{M}\\
    V_{\alpha}^{\sf T}\otimes I_{M}
    \end{array}\right]\mathcal{A}_2 d + \mu \left[\begin{array}{c}\frac{1}{\sqrt{K}}\mathbbm{1}^{\sf T}\otimes I_{M}\\
    V_{\alpha}^{\sf T}\otimes I_{M}
    \end{array}\right]\mathcal{A}_2\boldsymbol{s}_{n}^{B}
\end{align}
Note that according to (\ref{g0}), we have 
\begin{equation}
(\frac{1}{\sqrt{K}}\mathbbm{1}^{\sf T}\otimes I_M)\mathcal{A}_2d =(\frac{1}{\sqrt{K}}\mathbbm{1}^{\sf T}\otimes I_M)d = \frac{1}{\sqrt{K}}\sum_k  \nabla J_k(w^\star) = 0
\end{equation}
Also, consider the average of the Hessian matrices of all agents denoted by
\begin{equation}
    \bar{H}_{n-1} = \frac{1}{K}\sum_k H_{k,n-1} 
\end{equation} 
then according to (\ref{p_a1a2}), the recursion (\ref{diff_2}) can be split as
\begin{align}\label{diff_3}
    \bar{\boldsymbol{w}}_{n} =& (I - \mu\bar{H}_{n-1})\bar{\boldsymbol{w}}_{n-1} - \mu(\frac{1}{\sqrt{K}}\mathbbm{1}^{\sf T}\otimes I_M)\mathcal{H}_{n-1}(V_\alpha\otimes I)\check{\boldsymbol{w}}_{n-1} + \mu(\frac{1}{\sqrt{K}}\mathbbm{1}^{\sf T}\otimes I_M) \boldsymbol{s}_{n}^B\\
    \check{\boldsymbol{w}}_{n} =& \left(P_{\alpha}\otimes I_M - \mu((V_\alpha^{\sf T}A_2)\otimes I_M) \mathcal{H}_{n-1}(V_\alpha\otimes I_M)\right)\check{\boldsymbol{w}}_{n-1} - \mu((V_\alpha^{\sf T}A_2)\otimes I_M)) \mathcal{H}_{n-1}(\frac{1}{\sqrt{K}}\mathbbm{1}\otimes{I_M})\bar{\boldsymbol{w}}_{n-1}\notag\\
    & + \mu((V_\alpha^{\sf T}A_2)\otimes I_M)d + \mu((V_\alpha^{\sf T}A_2)\otimes I_M)\boldsymbol{s}_n^B
\end{align}
{
then we have
\begin{align}\label{sum_mse_w}
\Vert{\widetilde{\boldsymbol{\scriptstyle\mathcal{W}}}}_{n}\Vert^2 \overset{(a)}{= }\Vert\mathcal{V}^{\sf T}{\widetilde{\boldsymbol{\scriptstyle\mathcal{W}}}}_{n}\Vert^2  = \Vert \bar{\boldsymbol{w}}_{n}  \Vert^2 + \Vert \check{\boldsymbol{w}}_{n} \Vert^2
\end{align}
where $(a)$ follows from (\ref{vvt}). Note that in the \emph{centralized} method, we have 
\begin{align}\label{pv0}
    P_\alpha = 0, \quad V_\alpha = 0
\end{align}
Thus, the distance to $w^\star$ for decentralized models includes the additional term, $\Vert \check{\boldsymbol{w}}_{n} \Vert^2$, compared to the centralized method. We now prove that this extra term measures the consensus distance, i.e., the distance between local models and their centroid. Consider the centroid of local models defined by 
\begin{align}
    \boldsymbol{w}_{c,n} \overset{\Delta}{=} \frac{1}{K}\sum_{k=1} \boldsymbol{w}_{k,n} 
\end{align}
and the associated error vector to $w^{\star}$:
\begin{align}
    \tilde{\boldsymbol{w}}_{c,n} \overset{\Delta}{=} w^{\star} - \boldsymbol{w}_{c,n}  = \frac{1}{K}\sum_{k=1} \tilde{\boldsymbol{w}}_{k,n} 
\end{align}
and we introduce their collection:
\begin{align}\label{w_cd}
{{\boldsymbol{\scriptstyle\mathcal{W}}}}_{c, n} \overset{\Delta}{=} \mathbbm{1}\otimes {\boldsymbol{w}}_{c,n},\quad
{\widetilde{\boldsymbol{\scriptstyle\mathcal{W}}}}_{c, n} \overset{\Delta}{=} \mathbbm{1}\otimes \tilde{\boldsymbol{w}}_{c,n} =  \frac{1}{K}((\mathbbm{1}\mathbbm{1}^{\sf T})\otimes I_M){\widetilde{\boldsymbol{\scriptstyle\mathcal{W}}}}_{n}
\end{align}
The consensus distance is defined by the square distance between the local models and their centroid, i.e., $\Vert {{\boldsymbol{\scriptstyle\mathcal{W}}}}_{n} -{{\boldsymbol{\scriptstyle\mathcal{W}}}}_{c, n} \Vert^2$. We first analyze the common term of decentralized and centralized methods in (\ref{sum_mse_w}). Specifically, we have
\begin{align}\label{r_bw_cw}
  \bar{\boldsymbol{w}}_{n} =  \left(\frac{1}{\sqrt{K}}\mathbbm{1}^{\sf T} \otimes I_M\right){\widetilde{\boldsymbol{\scriptstyle\mathcal{W}}}}_{n} = \frac{1}{\sqrt{K}} \sum_k  \tilde{\boldsymbol{w}}_{k,n} = \sqrt{K}\tilde{\boldsymbol{w}}_{c,n}
\end{align}
so that
\begin{align}\label{tilde_w_2_e}
    \Vert \bar{\boldsymbol{w}}_{n} \Vert^2 = K\Vert\tilde{\boldsymbol{w}}_{c,n}\Vert^2 = \Vert{\widetilde{\boldsymbol{\scriptstyle\mathcal{W}}}}_{c, n}\Vert^2
\end{align}
which means that the common term of decentralized and centralized methods denoted by $\Vert \bar{\boldsymbol{w}}_{n} \Vert^2$ actually measures the distance of the network centroid to the local minimizer $w^{\star}$. We next analyze the relationship between $\Vert \check{\boldsymbol{w}}_{n} \Vert^2$ and the consensus distance, for which we have the following equality for the difference between local models and their centroid:
\begin{align}\label{check_w_2_nov}
{{\boldsymbol{\scriptstyle\mathcal{W}}}}_{n} - {{\boldsymbol{\scriptstyle\mathcal{W}}}}_{c, n} = {\widetilde{\boldsymbol{\scriptstyle\mathcal{W}}}}_{n} - {\widetilde{\boldsymbol{\scriptstyle\mathcal{W}}}}_{c, n} \overset{(a)}{=} (I_{KM} - \frac{1}{K}(\mathbbm{1}\mathbbm{1}^{\sf T})\otimes I_M){\widetilde{\boldsymbol{\scriptstyle\mathcal{W}}}}_{n} \overset{(b)}{=} ((V_{\alpha}V_{\alpha}^{\sf T})\otimes I_M){\widetilde{\boldsymbol{\scriptstyle\mathcal{W}}}}_{n} \overset{(c)}{=} (V_{\alpha}\otimes I_M)\check{\boldsymbol{w}}_n
\end{align}
where $(a)$ follows from (\ref{w_cd}),  $(b)$ follows from (\ref{a_pro}), and $(c)$ follows from (\ref{diff_2}). Then applying the third equality of (\ref{a_pro}), we have
\begin{align}\label{check_w_2_e}
\Vert {\widetilde{\boldsymbol{\scriptstyle\mathcal{W}}}}_{n} - {\widetilde{\boldsymbol{\scriptstyle\mathcal{W}}}}_{c, n}\Vert^2 = \Vert (V_{\alpha}\otimes I_M)\check{\boldsymbol{w}}_n\Vert^2 = \Vert\check{\boldsymbol{w}}_n\Vert^2
\end{align}
Substituting (\ref{tilde_w_2_e}) and (\ref{check_w_2_e}) into (\ref{sum_mse_w}), we have
\begin{align}\label{w+w-c_sum}
\mathds{E}\Vert{\widetilde{\boldsymbol{\scriptstyle\mathcal{W}}}}_{n}\Vert^2 =  \mathds{E}\Vert{\widetilde{\boldsymbol{\scriptstyle\mathcal{W}}}}_{c, n}\Vert^2+ \mathds{E}\Vert {\widetilde{\boldsymbol{\scriptstyle\mathcal{W}}}}_{n} - {\widetilde{\boldsymbol{\scriptstyle\mathcal{W}}}}_{c, n}\Vert^2  
\end{align}
Thus, the mean-square distance of the network models to the local minimizer is composed of two components: the mean-square distance of the network centroid to the local minimizer and the consensus distance. Decentralized methods include the additional consensus distance term, which increases the overall distance from the local minimizer compared to the centralized method.
}

{We now analyze the upper bounds of the mean-square distance $\mathds{E}\Vert{\widetilde{\boldsymbol{\scriptstyle\mathcal{W}}}}_{n}\Vert^2$}. Here we start from \emph{decentralized methods}.
Conditioning both sides of (\ref{diff_3}) on $\mathcal{F}_{n-1}$, we now analyze the second-order moments of the two terms separately. First, for $\check{\boldsymbol{w}}_{n}$, consider
\begin{align}\label{ka3}
    \mathcal{P}_{\alpha} = P_\alpha \otimes I_M, \quad \mathcal{V}_\alpha = V_\alpha\otimes I_M, \quad \mathds{1} = \mathbbm{1}\otimes I_M
\end{align}
According to (\ref{diff_3}), we have
\begin{align}\label{checkw_2}
\mathds{E}[\Vert\check{\boldsymbol{w}}_{n}\Vert^2 \vert\mathcal{F}_{n-1}]\overset{(a)}{=}& \mathds{E}\Vert \left(\mathcal{P}_{\alpha} - \mu\mathcal{V}_\alpha^{\sf T}\mathcal{A}_2\mathcal{H}_{n-1}\mathcal{V}_\alpha\right)\check{\boldsymbol{w}}_{n-1} - \frac{\mu}{\sqrt{K}}\mathcal{V}_\alpha^{\sf T}\mathcal{A}_2\mathcal{H}_{n-1}\mathds{1}\bar{\boldsymbol{w}}_{n-1} + \mu\mathcal{V}_\alpha^{\sf T}\mathcal{A}_2d \Vert^2 + \mu^2 \mathds{E}[\Vert\mathcal{V}_\alpha^{\sf T}\mathcal{A}_2\boldsymbol{s}_n^B\Vert^2\vert \mathcal{F}_{n-1}]
\end{align}
where (a) follows from (\ref{s0}). 

We bound the first term on the right hand of (\ref{checkw_2}). Note that
\begin{equation}
    \Vert P_\alpha \check{\boldsymbol{w}}_{n-1}\Vert^2 \le \check{\boldsymbol{w}}_{n-1}^{\sf T} P_\alpha^2\check{\boldsymbol{w}}_{n-1} \le \lambda_{\max}(P_\alpha^2) \Vert\check{\boldsymbol{w}}_{n-1}\Vert^2
\end{equation}
and we know $0< \rho(P_\alpha) < 1$ from \cite{sayed2014adaptation} where $\rho(P_\alpha)$ denotes the spectral radius of $P_\alpha$.   Let $t = \rho(P_\alpha)$, we have
\begin{align}\label{check_1}
    & \mathds{E}\Vert \left(\mathcal{P}_{\alpha} - \mu\mathcal{V}_\alpha^{\sf T}\mathcal{A}_2\mathcal{H}_{n-1}\mathcal{V}_\alpha\right)\check{\boldsymbol{w}}_{n-1} - \frac{\mu}{\sqrt{K}}\mathcal{V}_\alpha^{\sf T}\mathcal{A}_2\mathcal{H}_{n-1}\mathds{1}\bar{\boldsymbol{w}}_{n-1} + \mu\mathcal{V}_\alpha^{\sf T}\mathcal{A}_2d \Vert^2\notag\\
     =& \mathds{E} \Vert t\cdot\frac{1}{t}\mathcal{P}_{\alpha}\check{\boldsymbol{w}}_{n-1} + (1-t)\cdot\frac{1}{1-t}\left(- \mu\mathcal{V}_\alpha^{\sf T}\mathcal{A}_2\mathcal{H}_{n-1}\mathcal{V}_\alpha\check{\boldsymbol{w}}_{n-1} - \frac{\mu}{\sqrt{K}}\mathcal{V}_\alpha^{\sf T}\mathcal{A}_2\mathcal{H}_{n-1}\mathds{1}\bar{\boldsymbol{w}}_{n-1} + \mu\mathcal{V}_\alpha^{\sf T}\mathcal{A}_2d\right )\Vert^2\notag\\
     \overset{(a)}{\le} & \frac{1}{t}\mathds{E}\Vert\mathcal{P}_{\alpha}\check{\boldsymbol{w}}_{n-1}\Vert^2 + \frac{1}{(1-t)}\mathds{E}\Vert- \mu\mathcal{V}_\alpha^{\sf T}\mathcal{A}_2\mathcal{H}_{n-1}\mathcal{V}_\alpha\check{\boldsymbol{w}}_{n-1} - \frac{\mu}{\sqrt{K}}\mathcal{V}_\alpha^{\sf T}\mathcal{A}_2\mathcal{H}_{n-1}\mathds{1}\bar{\boldsymbol{w}}_{n-1} + \mu\mathcal{V}_\alpha^{\sf T}\mathcal{A}_2d\Vert^2\notag\\
     \overset{(b)}{\le}&(t+O(\mu^2))\mathds{E}\Vert\check{\boldsymbol{w}}_{n-1}\Vert^2 + O(\mu^2)\mathds{E}\Vert\bar{\boldsymbol{w}}_{n-1}\Vert^2 + O(\mu^2)
\end{align}
where $(a)$ and $(b)$ follow from Jensen's inequality, {and the last constant term $O(\mu^2)$ arises from the network heterogeneity which guarantees $d\neq 0$, otherwise this term is 0.}

We now bound the second term of (\ref{checkw_2}), which is related to the gradient noise. Consider
\begin{align}\label{checks2}
    \quad \check{\boldsymbol{s}}_{n}^B  = {V}^{\sf T}_{\alpha}\mathcal{A}_2\boldsymbol{s}_n^B
 \end{align}
 we have
\begin{align}\label{check_2}
    \mu^2\mathds{E}\left[\Vert \check{\boldsymbol{s}}_{n}^B\Vert^2 \vert\mathcal{F}_{n-1}\right]\le \mu^2\Vert\mathcal{V}_\alpha^{\sf T}\mathcal{A}_2\Vert^2\mathds{E}\left[\Vert\boldsymbol{s}_n^B\Vert^2\vert\mathcal{F}_{n-1}\right]
\end{align}
Thus, we should bound $\mathds{E}\left[\Vert\boldsymbol{s}_n^B\Vert^2\vert\mathcal{F}_{n-1}\right]$. Fortunately, we have
\begin{align}\label{snb}
\mathds{E}\left[\Vert\boldsymbol{s}_n^B\Vert^2\vert\mathcal{F}_{n-1}\right] &=\sum\limits_k \mathds{E}\left[\Vert\boldsymbol{s}_{k,n}^B(\boldsymbol{w}_{k,n-1})\Vert^2\vert\mathcal{F}_{n-1}\right] \notag\\
&\overset{(a)}{\le} O(\frac{1}{B})\sum\limits_k \mathds{E}\Vert \tilde{\boldsymbol{w}}_{k,n-1}\Vert^2 + O(\frac{1}{B})\notag\\
&\le O(\frac{1}{B})\mathds{E}\Vert {\widetilde{\boldsymbol{\scriptstyle\mathcal{W}}}}_{n} \Vert^2 + O(\frac{1}{B}) \notag\\
&=  O(\frac{1}{B})(\mathds{E}\Vert\mathcal{V}\mathcal{V}^{\sf T}{\widetilde{\boldsymbol{\scriptstyle\mathcal{W}}}}_{n}\Vert^2) + O(\frac{1}{B}) \notag\\
&\le O(\frac{1}{B})(\mathds{E}\Vert\check{\boldsymbol{w}}_{n-1}\Vert^2 +\mathds{E}\Vert\bar{\boldsymbol{w}}_{n-1}\Vert^2 )+O(\frac{1}{B})
\end{align}
where $(a)$ follows from (\ref{f_s2}).  Combining (\ref{checkw_2}), (\ref{check_1}), (\ref{snb}) and (\ref{check_2}), we obtain
\begin{align}\label{checkw2}
\mathds{E}\Vert\check{\boldsymbol{w}}_n\Vert^2 \le \left(\rho(P_\alpha)+O(\mu^2)\right)\mathds{E}\Vert\check{\boldsymbol{w}}_{n-1}\Vert^2  + O(\mu^2)\mathds{E}\Vert \bar{\boldsymbol{w}}_{n-1}\Vert^2 + O(\mu^2)
\end{align}

We now analyse the size of $\mathds{E}\Vert \bar{\boldsymbol{w}}_{n}\Vert^2$. Consider
\begin{align}\label{s_bar_ss}
    \bar{\boldsymbol{s}}^B_n \overset{\Delta}{=} \frac{1}{\sqrt{K}}\mathds{1}^{\sf T} \boldsymbol{s}_{n}^B =\frac{1}{\sqrt{K}} \sum\limits_k \boldsymbol{s}_{k,n}^{B}
\end{align}
for which we have
\begin{align}\label{bars2}
    \mathds{E}\left[\Vert\frac{1}{\sqrt{K}} \sum\limits_k \boldsymbol{s}_{k,n}^{B}\Vert^2 \vert\mathcal{F}_{n-1}\right] \overset{(a)}{\le} \frac{1}{K}\sum\limits_k\mathds{E}[\Vert\boldsymbol{s}_{k,n}^B\Vert^2\vert\mathcal{F}_{n-1}] =\frac{1}{K}\mathds{E}[\Vert\boldsymbol{s}_{n}^B\Vert^2\vert\mathcal{F}_{n-1}]
\end{align}
where $(a)$ follows from the sampling independence among agents.

Also, consider $0 < t = 1 - O(\mu) < 1$,
with (\ref{diff_3}) we have
\begin{align}\label{barw2}
    \mathds{E}\left[\Vert \bar{\boldsymbol{w}}_{n}\Vert^2 \vert \mathcal{F}_{n-1}\right] \overset{(a)}{=}&\mathds{E}\Vert (I - \mu \bar{H}_{n-1})\bar{\boldsymbol{w}}_{n-1} - \frac{\mu}{\sqrt{K}}\mathds{1}^{\sf T}\mathcal{H}_{n-1}\mathcal{V}_{\alpha}\check{\boldsymbol{w}}_{n-1}\Vert^2 + \frac{\mu^2}{K}\mathds{E}[\Vert\mathds{1}^{\sf T} \boldsymbol{s}_{n}^B\Vert^2\vert\mathcal{F}_{n-1}]\notag\\
   \overset{(b)}{\le} & \frac{1}{t}\mathds{E}\Vert(I - \mu \bar{H}_{n-1})\bar{\boldsymbol{w}}_{n-1}\Vert^2 + \frac{\mu^2}{K(1-t)}\mathds{E}\Vert \mathds{1}^{\sf T}\mathcal{H}_{n-1}\mathcal{V}_{\alpha}\check{\boldsymbol{w}}_{n-1}\Vert^2 + \frac{\mu^2}{K}\mathds{E}[\Vert\boldsymbol{s}_{n}^B\Vert^2\vert\mathcal{F}_{n-1}]\notag\\
   \overset{(c)}{\le}& \left(\frac{(1+\mu L)^2}{1-O(\mu)}+O(\frac{\mu^2}{B})\right)\mathds{E}\Vert\bar{\boldsymbol{w}}_{n-1}\Vert^2+ O(\mu)\mathds{E}\Vert\check{\boldsymbol{w}}_{n-1}\Vert^2 + O(\frac{\mu^2}{B})
\end{align}
where $(a)$ follows from (\ref{s0}), $(b)$ follows from the Jensen's inequality and (\ref{bars2}), and $(c)$ follows from (\ref{snb}) and the Lipschitz condition in Assumption \ref{ass_smooth}. 

Combining  (\ref{checkw2}) and (\ref{barw2}), we obtain
\begin{align}\label{tildew2}
    \mathds{E}\left[\begin{array}{c}
         \Vert\bar{\boldsymbol{w}}_{n}\Vert^2\\
          \Vert\check{\boldsymbol{w}}_{n}\Vert^2
    \end{array}\right]\le \left[\begin{array}{cc}
       \frac{(1+\mu L)^2}{1-O(\mu)} & O(\mu) \\
         O(\mu^2)&  \rho(P_\alpha)+O(\mu^2)
    \end{array}\right]\left[\begin{array}{c}
         \mathds{E}\Vert\bar{\boldsymbol{w}}_{n-1}\Vert^2\\
         \mathds{E} \Vert\check{\boldsymbol{w}}_{n-1}\Vert^2
    \end{array}\right] + \left[\begin{array}{c}
         O(\frac{\mu^2}{B})\\
         O(\mu^2)
    \end{array}\right]
\end{align}
Let
\begin{equation}
    \Gamma_1 = \left[\begin{array}{cc}
       \frac{(1+\mu L)^2}{1-O(\mu)} & O(\mu) \\
         O(\mu^2)&  \rho(P_\alpha)+O(\mu^2)
    \end{array}\right]
\end{equation}
and by iterating (\ref{tildew2}), we obtain
\begin{align}\label{ftw}
    \mathds{E}\left[\begin{array}{c}
         \Vert\bar{\boldsymbol{w}}_{n}\Vert^2\\
          \Vert\check{\boldsymbol{w}}_{n}\Vert^2
    \end{array}\right]\le \Gamma_1^{n+1}\mathds{E}\left[\begin{array}{c}
         \Vert\bar{\boldsymbol{w}}_{-1}\Vert^2\\
          \Vert\check{\boldsymbol{w}}_{-1}\Vert^2
    \end{array}\right] + \sum\limits_{i = 0}^{n} \Gamma_1^{i}\left[\begin{array}{c}
         O(\frac{\mu^2}{B})\\
         O(\mu^2)
    \end{array}\right] = \Gamma_1^{n+1}\mathds{E}\left[\begin{array}{c}
         \Vert\bar{\boldsymbol{w}}_{-1}\Vert^2\\
          \Vert\check{\boldsymbol{w}}_{-1}\Vert^2
    \end{array}\right]  + (I - \Gamma_1)^{-1}(I-\Gamma_1^{n+1})\left[\begin{array}{c}
         O(\frac{\mu^2}{B})\\
         O(\mu^2)
    \end{array}\right]
\end{align}
To proceed, it is necessary to compute $(I - \Gamma_1)^{-1}(I-\Gamma_1^{n+1})$. Basically, since
\begin{align}
     1 - \frac{(1+\mu L)^2}{1-O(\mu)} = \frac{(1-O(\mu))-(1+\mu L)^2}{1 - \mu L} = -O(\mu)
\end{align}
we have
\begin{align}\label{mi}
    (I - \Gamma_1)^{-1} =& \left[\begin{array}{cc}
        1 - \frac{(1+\mu L)^2}{1-O(\mu)} & -O(\mu) \\
        -O(\mu^2) &  1 - \rho(P_\alpha)+O(\mu^2) + O(\mu^2)
    \end{array}\right]^{-1} = \left[\begin{array}{cc}
        -O(\mu) & -O(\mu) \\
        -O(\mu^2) &  O(1)
    \end{array}\right]^{-1} \notag\\
    =& -\frac{1}{O(\mu)}\left[\begin{array}{cc}
        O(1) & O(\mu) \\
        O(\mu^2) &  -O(u)
    \end{array}\right] = \left[\begin{array}{cc}
        -O(\frac{1}{\mu}) & -O(1) \\
        -O(\mu) &  O(1)
    \end{array}\right]
\end{align}

As for the matrix power $\Gamma_1^{n+1}$, 
\begin{align}\label{gamma1_n}
    \Gamma_1^{n+1} = \left[\begin{array}{cc}
      (\frac{(1+\mu L)^2}{1-O(\mu)})^{n+1}   & O(\mu) \\
        O(\mu^2) & \rho^{n+1}(P_\alpha)
    \end{array}\right] 
\end{align}
for which by resorting to Lemma 2 in \cite{VlaskiS21}, with $n\le O(\frac{1}{\mu})$, we have
\begin{align}\label{o1ns}
   \left(\frac{(1+\mu L)^2}{1-O(\mu)}\right)^{n+1}  = O(1), \quad\quad 1 -  \left(\frac{(1+\mu L)^2}{1-O(\mu)}\right)^{n+1}  = -O(1)
\end{align}
then we have
\begin{align}\label{mp}
    I - \Gamma_1^{n+1}= \left[\begin{array}{cc}
      1 - \left(\frac{(1+\mu L)^2}{1-O(\mu)}\right)^{n+1}   & -O(\mu) \\
        -O(\mu^2) & 1 - \rho^{n+1}(P_\alpha)
    \end{array}\right] = \left[\begin{array}{cc}
     - O(1)  & -O(\mu) \\
        -O(\mu^2) & O(1) 
    \end{array}\right] 
\end{align}
Also, according to assumption \ref{ass_ori}, we have
\begin{align}
  &\mathds{E}\Vert\bar{\boldsymbol{w}}_{-1}\Vert^2 = \mathds{E}\Vert \frac{1}{\sqrt{K}}\sum\limits_{k=1}^{K}\tilde{\boldsymbol{w}}_{k,-1}\Vert^2  \le \sum\limits_{k=1}^{K}\mathds{E}\Vert\tilde{\boldsymbol{w}}_{k,-1}\Vert^2 \le o(\frac{\mu}{B}) \\
  &\mathds{E}\Vert\check{\boldsymbol{w}}_{-1}\Vert^2 =\mathds{E}\Vert\mathcal{V}^{\sf T}_{\alpha}{\widetilde{\boldsymbol{\scriptstyle\mathcal{W}}}}_{-1}\Vert^2 \le \Vert\mathcal{V}^{\sf T}_{\alpha}\Vert^2 \mathds{E}\Vert{\widetilde{\boldsymbol{\scriptstyle\mathcal{W}}}}_{-1}\Vert^2 \le o(\frac{\mu}{B}) 
\end{align}
Then with (\ref{gamma1_n}) and (\ref{o1ns}), we obtain
\begin{align}\label{w02}
\Gamma_1^{n+1}\mathds{E}\left[\begin{array}{c}
         \Vert\bar{\boldsymbol{w}}_{-1}\Vert^2\\
          \Vert\check{\boldsymbol{w}}_{-1}\Vert^2
    \end{array}\right] \le \left[\begin{array}{c}
          o(\frac{\mu}{B})\\
           o(\frac{\mu}{B})
    \end{array}\right] 
\end{align}

Then, substituting (\ref{mi}), (\ref{mp}) and (\ref{w02}) into (\ref{ftw}), we obtain
\begin{align}\label{fbar_check_2_s}
     \mathds{E}\left[\begin{array}{c}
         \Vert\bar{\boldsymbol{w}}_{n}\Vert^2\\
          \Vert\check{\boldsymbol{w}}_{n}\Vert^2
    \end{array}\right] \le \left[\begin{array}{c}
          o(\frac{\mu}{B})\\
           o(\frac{\mu}{B})
    \end{array}\right]  + \left[\begin{array}{cc}
        -O(\frac{1}{\mu}) & -O(1) \\
        -O(\mu) &  O(1)
    \end{array}\right]\left[\begin{array}{cc}
     -O(1) & -O(\mu) \\
        -O(\mu^2) & O(1) 
    \end{array}\right]\left[\begin{array}{c}
         O(\frac{\mu^2}{B})\\
         O(\mu^2)
    \end{array}\right] \le \left[\begin{array}{c}
         O(\frac{\mu}{B})+O(\mu^2)\\
         O(\mu^2)
    \end{array}\right]
\end{align}
from which we have
\begin{align}\label{wn2}
\mathds{E}\Vert{\widetilde{\boldsymbol{\scriptstyle\mathcal{W}}}}_{n}\Vert^2 \le \Vert\mathcal{V}\Vert^2\mathds{E}\Vert{\mathcal{V}^{\sf T}\widetilde{\boldsymbol{\scriptstyle\mathcal{W}}}}_{n}\Vert^2 = \Vert\mathcal{V}\Vert^2 \mathds{E}(\Vert\bar{\boldsymbol{w}}_{n}\Vert^2+\Vert\check{\boldsymbol{w}}_{n}\Vert^2)\le O(\frac{\mu}{B}) + O( \mu^2)
\end{align}
According to (\ref{cbmu}), we obtain 
\begin{align}\label{wn2_2}
\mathds{E}\Vert{\widetilde{\boldsymbol{\scriptstyle\mathcal{W}}}}_{n}\Vert^2 \le O(\mu^{1+\eta}) + O(\mu^2) = O(\mu^\gamma) 
\end{align}
where $\gamma = \min\{1+\eta, 2\}$.

We finally examine the bounds of the \emph{centralized method}. As mentioned via (\ref{pv0}), in the centralized method, $\check{\boldsymbol{w}}_{n}$ is always $0$. Thus, we only need to analyze $\mathds{E}\Vert\bar{\boldsymbol{w}}_{n}\Vert^2$, for which according to (\ref{barw2}), we have:
\begin{align}
     \mathds{E}\Vert \bar{\boldsymbol{w}}_{n}\Vert^2 \le \left(\frac{(1+\mu L)^2}{1-O(\mu)}+O(\frac{\mu^2}{B})\right)^{n+1}\mathds{E}\Vert\bar{\boldsymbol{w}}_{-1}\Vert^2+ \frac{1- \left(\frac{(1+\mu L)^2}{1-O(\mu)}+O(\frac{\mu^2}{B})\right)^{n+1}}{1- \left(\frac{(1+\mu L)^2}{1-O(\mu)}+O(\frac{\mu^2}{B})\right)}\times O(\frac{\mu^2}{B}) 
\end{align}
By following the same logic with the derivation process from (\ref{gamma1_n})--(\ref{wn2_2}),  for the centralized method, we have
\begin{align}\label{wn2_centra}
  \mathds{E}\Vert{\widetilde{\boldsymbol{\scriptstyle\mathcal{W}}}}_{n}\Vert^2 \le O(\frac{\mu}{B}) = O(\mu^{1+\eta})  
\end{align}
By comparing (\ref{wn2_2}) and (\ref{wn2_centra}), we see that the decentralized methods have extra $O(\mu^2)$ noisy terms related to the network heterogeneity and graph structure. {According to (\ref{w+w-c_sum}), the extra $O(\mu^2)$ noisy terms also relate to the consensus distance.}

\section{Proof for Lemma \ref{mse} and \ref{mse_centra}: Upper bound for the fourth-order moment}\label{mse4}
In this section, we analyze the size of $\mathds{E}\Vert{\widetilde{\boldsymbol{\scriptstyle\mathcal{W}}}}_{n}\Vert^4$ for later use. We recall the relation of $\bar{\boldsymbol{w}}_n$ to $\check{\boldsymbol{w}}_n$ in (\ref{diff_3}), and use the following equality:
\begin{align}
    \Vert \boldsymbol{a} + \boldsymbol{b}\Vert^4 = \Vert \boldsymbol{a}\Vert^4 + \Vert \boldsymbol{b}\Vert^4 + 2\Vert \boldsymbol{a}\Vert^2 \Vert \boldsymbol{b}\Vert^2 + 4\boldsymbol{a}^{\sf T}\boldsymbol{b}\Vert \boldsymbol{a}\Vert^2+ 4\boldsymbol{a}^{\sf T}\boldsymbol{b}\Vert \boldsymbol{b}\Vert^2 + 4(\boldsymbol{a}^{\sf T}\boldsymbol{b})^2
\end{align}
where $\boldsymbol{a}$ and $\boldsymbol{b}$ are two column vectors. When $\mathds{E}\boldsymbol{b} = 0$, with the Cauchy–Schwarz inequality, we have:
\begin{align}\label{sum4}
    \mathds{E}\Vert\boldsymbol{a}+\boldsymbol{b}\Vert^4 \le \mathds{E}\Vert\boldsymbol{a}\Vert^4 + 3\mathds{E}\Vert\boldsymbol{b}\Vert^4+ 8\mathds{E}\Vert\boldsymbol{a}\Vert^2\Vert\boldsymbol{b}\Vert^2
\end{align}

We first analyze the size of $\mathds{E}\Vert\bar{\boldsymbol{w}}_n\Vert^4$. Let $t = 1 - O(\mu)$, we have
\begin{align}\label{barw4}
\mathds{E}\Vert\bar{\boldsymbol{w}}_{n}\Vert^4 \overset{(a)}{=}&\mathds{E}\Vert (I - \mu \bar{H}_{n-1})\bar{\boldsymbol{w}}_{n-1} - \frac{\mu}{\sqrt{K}}\mathds{1}^{\sf T}\mathcal{H}_{n-1}\mathcal{V}_{\alpha}\check{\boldsymbol{w}}_{n-1} \Vert^4 + 3\mu^4\mathds{E}\Vert\frac{1}{\sqrt{K}}\mathds{1}^{\sf T} \boldsymbol{s}_{n}^B \Vert^4 \notag\\
& + 8\mu^2\mathds{E}\Vert(I - \mu \bar{H}_{n-1})\bar{\boldsymbol{w}}_{n-1} - \mu\frac{1}{\sqrt{K}}\mathds{1}^{\sf T}\mathcal{H}_{n-1}\mathcal{V}_{\alpha}\check{\boldsymbol{w}}_{n-1} \Vert^2\Vert\frac{1}{\sqrt{K}}\mathds{1}^{\sf T} \boldsymbol{s}_{n}^B \Vert^2\notag\\
\overset{(b)}{\le} &\frac{1}{t^3}\mathds{E}\Vert (I - \mu \bar{H}_{n-1})\bar{\boldsymbol{w}}_{n-1}\Vert^4 + \frac{O(\mu^4)}{(1-t)^3}\mathds{E}\Vert\check{\boldsymbol{w}}_{n-1}\Vert^4 + 3\mu^4\mathds{E}\Vert\frac{1}{\sqrt{K}}\mathds{1}^{\sf T} \boldsymbol{s}_{n}^B \Vert^4 \notag\\
&+8\mu^2\mathds{E}\left[\left(\frac{1}{t}\Vert (I - \mu \bar{H}_{n-1})\bar{\boldsymbol{w}}_{n-1}\Vert^2 + \frac{O(\mu^2)}{1-t}\Vert\check{\boldsymbol{w}}_{n-1}\Vert^2\right)\cdot\Vert\frac{1}{\sqrt{K}}\mathds{1}^{\sf T} \boldsymbol{s}_{n}^B \Vert^2\right]\notag\\
\le & \frac{(1+\mu L)^4}{(1-O(\mu))^3}\mathds{E}\Vert\bar{\boldsymbol{w}}_{n-1}\Vert^4+O(\mu)\mathds{E}\Vert\check{\boldsymbol{w}}_{n-1}\Vert^4+O(\mu^4)\mathds{E}\Vert\frac{1}{\sqrt{K}}\mathds{1}^{\sf T} \boldsymbol{s}_{n}^B \Vert^4 + O(\mu^2)\mathds{E}\Vert\bar{\boldsymbol{w}}_{n-1}\Vert^2\Vert\frac{1}{\sqrt{K}}\mathds{1}^{\sf T} \boldsymbol{s}_{n}^B \Vert^2 \notag\\
&+ O(\mu^3)\mathds{E}\Vert\check{\boldsymbol{w}}_{n-1}\Vert^2\Vert\frac{1}{\sqrt{K}}\mathds{1}^{\sf T} \boldsymbol{s}_{n}^B \Vert^2
\end{align}
where $(a)$ follows from $(\ref{s0})$ and (\ref{sum4}), and  $(b)$ follows from Jensen's inequality.

Then for $\check{\boldsymbol{w}}_{n-1}$, similar to the proof when analyzing the second-order error in Appendix \ref{mse2}, we consider $t = \rho(P_\alpha)$ which is the spectral radius of $P_\alpha$ and obtain
\begin{align}\label{checkw4}
    \mathds{E}\Vert\check{\boldsymbol{w}}_n\Vert^4 \overset{(a)}{\le}&\mathds{E}\Vert \left(\mathcal{P}_{\alpha} - \mu\mathcal{V}_\alpha^{\sf T}\mathcal{A}_2\mathcal{H}_{n-1}\mathcal{V}_\alpha\right)\check{\boldsymbol{w}}_{n-1} - \frac{\mu}{\sqrt{K}}\mathcal{V}_\alpha^{\sf T}\mathcal{A}_2\mathcal{H}_{n-1}\mathds{1}\bar{\boldsymbol{w}}_{n-1} + \mu\mathcal{V}_\alpha^{\sf T}\mathcal{A}_2d \Vert^4 + 3\mu^4\mathds{E}\Vert\mathcal{V}^{\sf T}_{\alpha}\mathcal{A}_2\boldsymbol{s}_n^B\Vert^4\notag\\
    & + 8\mu^2\mathds{E}\left[\Vert\mathcal{V}^{\sf T}_{\alpha}\mathcal{A}_2\boldsymbol{s}_n^B\Vert^2\Vert\left(\mathcal{P}_{\alpha} - \mu\mathcal{V}_\alpha^{\sf T}\mathcal{A}_2\mathcal{H}_{n-1}\mathcal{V}_\alpha\right)\check{\boldsymbol{w}}_{n-1} - \frac{\mu}{\sqrt{K}}\mathcal{V}_\alpha^{\sf T}\mathcal{A}_2\mathcal{H}_{n-1}\mathds{1}\bar{\boldsymbol{w}}_{n-1} + \mu\mathcal{V}_\alpha^{\sf T}\mathcal{A}_2d \Vert^2\right]\notag\\
    \overset{(b)}{\le}&(\rho(P_\alpha)+O(\mu^4))\mathds{E}\Vert\check{\boldsymbol{w}}_{n-1}\Vert^4 + O(\mu^4)\mathds{E}\Vert\bar{\boldsymbol{w}}_{n-1}\Vert^4 + O(\mu^4)\mathds{E}\Vert d \Vert^4 + 3\mu^4\mathds{E}\Vert\mathcal{V}^{\sf T}_{\alpha}\mathcal{A}_2\boldsymbol{s}_n^B\Vert^4\notag\\
    & + 8\mu^2\mathds{E}\left[\Vert\mathcal{V}^{\sf T}_{\alpha}\mathcal{A}_2\boldsymbol{s}_n^B\Vert^2\left((\rho(P_\alpha)+O(\mu^2))\Vert\check{\boldsymbol{w}}_{n-1}\Vert^2+ O(\mu^2)\Vert\bar{\boldsymbol{w}}_{n-1}\Vert^2 + O(\mu^2)\Vert d\Vert^2\right)\right]
\end{align}
where $(a)$ follows from $(\ref{s0})$ and (\ref{sum4}), and $(b)$ follows from Jensen's inequality. 

To proceed with 
the analysis for $(\ref{barw4})$ and $(\ref{checkw4})$, we should analyze the fourth-order moment associated with gradient noise. Recall (\ref{checks2}) and (\ref{s_bar_ss}), we have
\begin{align}\label{sum_sn}
    \mathds{E}\Vert\bar{\boldsymbol{s}}^B_n \Vert^4 + \mathds{E}\Vert\check{\boldsymbol{s}}_{n}^B\Vert^4
\le\mathds{E}[\Vert\bar{\boldsymbol{s}}^B_n\Vert^2+\Vert\check{\boldsymbol{s}}^B_n \Vert^2 ]^2 = \mathds{E}\Vert\mathcal{V}^{\sf T}\mathcal{A}_2\boldsymbol{s}_{n}^B\Vert^4 \le O(1)\mathds{E}\Vert\boldsymbol{s}_n^B\Vert^4
\end{align}
Also,
\begin{align}\label{snb4}
    \mathds{E}\Vert\boldsymbol{s}^B_n\Vert^4 = &\mathds{E}[\Vert\boldsymbol{s}^B_n\Vert^2]^2 =\mathds{E}\left[\sum\limits_k \Vert\boldsymbol{s}_{k,n}^B\Vert^2\right]^2 = \mathds{E}\left[\sum\limits_k \frac{1}{K}\cdot K \Vert\boldsymbol{s}_{k,n}^B\Vert^2\right]^2\overset{(a)}{\le}K\sum\limits_k \mathds{E}\Vert\boldsymbol{s}_{k,n}^B\Vert^4\notag\\
    \overset{(b)}{\le}& O(\frac{1}{B^2})\sum\limits_k \mathds{e}\Vert\tilde{\boldsymbol{w}}_{k,n}\Vert^4 + O(\frac{1}{B^2})
\end{align}
where $(a)$ follows from Jensen's inequality, and $(b)$ follows from $(\ref{sb4l})$. Moreover, we have
\begin{align}\label{wn4}
    \sum\limits_k\Vert\tilde{\boldsymbol{w}}_{k,n}\Vert^4 \le (\sum\limits_k\Vert\tilde{\boldsymbol{w}}_{k,n}\Vert^2)^2 = \Vert\widetilde{\boldsymbol{\scriptstyle\mathcal{W}}}_{n}\Vert^4
\end{align}
Substituting $(\ref{wn4})$ into $(\ref{snb4})$, we have
\begin{align}\label{f_snb4}
    \mathds{E}\Vert\boldsymbol{s}^B_n\Vert^4 &\le O(\frac{1}{B^2})\mathds{E}\Vert\widetilde{\boldsymbol{\scriptstyle\mathcal{W}}}_{n}\Vert^4 + O(\frac{1}{B^2})\notag\\
    &\le O(\frac{1}{B^2})\Vert\mathcal{V}\Vert^4\mathds{E}\Vert\mathcal{V}^{\sf T}\widetilde{\boldsymbol{\scriptstyle\mathcal{W}}}_{n}\Vert^4 + O(\frac{1}{B^2})\notag\\
    & \overset{(a)}{\le} O(\frac{1}{B^2})(\mathds{E}\Vert\bar{\boldsymbol{w}}_{n-1}\Vert^4+\mathds{E}\Vert\check{\boldsymbol{w}}_{n-1}\Vert^4) + O(\frac{1}{B^2})
\end{align}
where $(a)$ follows from the Jensen's inequality such that
\begin{align}
\mathds{E}\Vert\mathcal{V}^{\sf T}\widetilde{\boldsymbol{\scriptstyle\mathcal{W}}}_{n}\Vert^4 = \mathds{E}[\Vert\bar{\boldsymbol{w}}_{n-1}\Vert^2 + \Vert\check{\boldsymbol{w}}_{n-1}\Vert^2]^2 \le 2\mathds{E}\Vert\bar{\boldsymbol{w}}_{n-1}\Vert^4 +2\mathds{E}\Vert\check{\boldsymbol{w}}_{n-1}\Vert^4
\end{align}
Substituting $(\ref{sum_sn})$, $(\ref{snb4})$, $(\ref{f_snb4})$ and $(\ref{snb})$ into (\ref{barw4}), we have
\begin{align}\label{f_barw}
    \mathds{E}\Vert\bar{\boldsymbol{w}}_n\Vert^4 \le&\frac{(1+\mu L)^4}{(1-O(\mu))^3}\mathds{E}\Vert\bar{\boldsymbol{w}}_{n-1}\Vert^4 + O(\mu)\mathds{E}\Vert\check{\boldsymbol{w}}_{n-1}\Vert^4 + O(\frac{\mu^4}{B^2})\mathds{E}\Vert\bar{\boldsymbol{w}}_{n-1}\Vert^4 + O(\frac{\mu^4}{B^2})\mathds{E}\Vert\check{\boldsymbol{w}}_{n-1}\Vert^4 + O(\frac{\mu^4}{B^2})\notag\\
    & +O(\frac{\mu^2}{B})\mathds{E}\Vert\bar{\boldsymbol{w}}_{n-1}\Vert^2(\Vert\check{\boldsymbol{w}}_{n-1}\Vert^2 +\Vert\bar{\boldsymbol{w}}_{n-1}\Vert^2 +O(1)) + O(\frac{\mu^3}{B})\mathds{E}\Vert\check{\boldsymbol{w}}_{n-1}\Vert^2(\Vert\check{\boldsymbol{w}}_{n-1}\Vert^2 +\Vert\bar{\boldsymbol{w}}_{n-1}\Vert^2 +O(1))\notag\\
    \overset{(a)}{\le}& \left(\frac{(1+\mu L)^4}{(1-O(\mu))^3} + O(\frac{\mu^2}{B})\right)\mathds{E}\Vert\bar{\boldsymbol{w}}_{n-1}\Vert^4 + O(\mu)\mathds{E}\Vert\check{\boldsymbol{w}}_{n-1}\Vert^4 + O(\frac{\mu^3}{B^2}) + {O(\frac{\mu^5}{B})}
\end{align}
where in $(a)$ we apply the result of (\ref{fbar_check_2_s}) and the following inequality:
\begin{align}\label{jensen4}
    \mathds{E}\Vert\bar{\boldsymbol{w}}_{n-1}\Vert^2\Vert\check{\boldsymbol{w}}_{n-1}\Vert^2 \le \frac{1}{2}\mathds{E}\Vert\bar{\boldsymbol{w}}_{n-1}\Vert^4 + \frac{1}{2}\mathds{E}\Vert\check{\boldsymbol{w}}_{n-1}\Vert^4
\end{align}

Similarly,  for $\mathds{E}\Vert\check{\boldsymbol{w}}_n\Vert^4$, we substitute $(\ref{sum_sn})$, $(\ref{snb4})$ and $(\ref{snb})$ into (\ref{checkw4}), and obtain
\begin{align}\label{f_checkw}
    \mathds{E}\Vert\check{\boldsymbol{w}}_n\Vert^4 \le &(\rho(P_\alpha)+O(\mu^4))\mathds{E}\Vert\check{\boldsymbol{w}}_{n-1}\Vert^4 + O(\mu^4)\mathds{E}\Vert\bar{\boldsymbol{w}}_{n-1}\Vert^4 + O(\mu^4) + O(\frac{\mu^4}{B^2})\mathds{E}\Vert\bar{\boldsymbol{w}}_{n-1}\Vert^4 +O(\frac{\mu^4}{B^2})\mathds{E}\Vert\check{\boldsymbol{w}}_{n-1}\Vert^4 + O(\frac{\mu^4}{B^2})\notag\\
    & + O(\frac{\mu^2}{B})\mathds{E}\left[\left(\Vert\bar{\boldsymbol{w}}_{n-1}\Vert^2+ \Vert\check{\boldsymbol{w}}_{n-1} \Vert^2 +O(1)\right)\left(\Vert\check{\boldsymbol{w}}_{n-1} \Vert^2 + O(\mu^2)\Vert\bar{\boldsymbol{w}}_{n-1} \Vert^2 + O(\mu^2)\right)\right]\notag\\
     \le & (\rho(P_\alpha)+O(\frac{\mu^2}{B}))\mathds{E}\Vert\check{\boldsymbol{w}}_{n-1}\Vert^4 + O(\frac{\mu^2}{B})\mathds{E}\Vert\bar{\boldsymbol{w}}_{n-1}\Vert^4 + O(\mu^4)
\end{align}

Combining  (\ref{f_barw}) and (\ref{f_checkw}), we obtain
\begin{align}\label{f_w4}
\mathds{E}\left[\begin{array}{c}
         \Vert\bar{\boldsymbol{w}}_{n}\Vert^4\\
          \Vert\check{\boldsymbol{w}}_{n}\Vert^4
    \end{array}\right]\le \left[\begin{array}{cc}
       \frac{(1+\mu L)^4}{(1-O(\mu))^3} & O(\mu) \\
         O(\frac{\mu^2}{B})&  \rho(P_\alpha)+O(\frac{\mu^2}{B})
    \end{array}\right]\left[\begin{array}{c}
         \mathds{E}\Vert\bar{\boldsymbol{w}}_{n-1}\Vert^4\\
         \mathds{E} \Vert\check{\boldsymbol{w}}_{n-1}\Vert^4
    \end{array}\right] + \left[\begin{array}{c}
         O(\frac{\mu^3}{B^2}) + {O(\frac{\mu^5}{B})}\\
         O(\mu^4)
    \end{array}\right]
\end{align}
Let 
\begin{align}\label{gamma_2_m}
    \Gamma_2 = \left[\begin{array}{cc}
       \frac{(1+\mu L)^4}{(1-O(\mu))^3} & O(\mu) \\
         O(\frac{\mu^2}{B})&  \rho(P_\alpha)+O(\frac{\mu^2}{B})
    \end{array}\right]
\end{align}
and by iterating (\ref{f_w4}), we have
\begin{align}\label{f_w42}
    \mathds{E}\left[\begin{array}{c}
         \Vert\bar{\boldsymbol{w}}_{n}\Vert^4\\
          \Vert\check{\boldsymbol{w}}_{n}\Vert^4
    \end{array}\right]\le \Gamma_2^{n+1}\left[\begin{array}{c}
         \Vert\bar{\boldsymbol{w}}_{-1}\Vert^4\\
          \Vert\check{\boldsymbol{w}}_{-1}\Vert^4
    \end{array}\right] + (I-\Gamma_2)^{-1}(I - \Gamma_2^{n+1})\left[\begin{array}{c}
          O(\frac{\mu^3}{B^2}) + {O(\frac{\mu^5}{B})}\\
         O(\mu^4)
    \end{array}\right]
\end{align}
for which with Assumption \ref{ass_ori} we have
\begin{align}\label{wb4}
    \mathds{E}\Vert\boldsymbol{\bar{{w}}}_{-1}\Vert^4 = \frac{1}{K} \Vert\sum\limits_k \tilde{\boldsymbol w}_{k,-1}\Vert^4 \overset{(a)}{\le} \frac{1}{K} \sum\limits_k\Vert\tilde{\boldsymbol w}_{k,-1}\Vert^4 \le o(\frac{\mu^2}{B^2})
\end{align}
where $(a)$ follows from Jensen's inequality. Also,
\begin{align}\label{wc4}
    \mathds{E}\Vert\check{\boldsymbol{w}}_{-1}\Vert^4 = \mathds{E}\Vert \mathcal{V}_\alpha^{\sf T}\widetilde{\boldsymbol{\scriptstyle\mathcal{W}}}_{-1}\Vert^4 \le \mathds{E}\Vert\mathcal{V}_\alpha^{\sf T}\Vert^4\left(\Vert\widetilde{\boldsymbol{\scriptstyle\mathcal{W}}}_{-1}\Vert^2\right)^2 \le o(\frac{\mu^2}{B^2})
\end{align}
Note that
\begin{align}\label{gamma_21}
   (I-\Gamma_2)^{-1} =  \left[\begin{array}{cc}
       1 - \frac{(1+\mu L)^4}{(1-O(\mu))^3} & -O(\mu) \\
         -O(\frac{\mu^2}{B})&  1 - \rho(P_\alpha) + O(\frac{\mu^2}{B})
    \end{array}\right]^{-1} =  \left[\begin{array}{cc}
       -O(\mu) & -O(\mu) \\
         -O(\frac{\mu^2}{B})&  O(1)
    \end{array}\right]^{-1} =  \left[\begin{array}{cc}
       -O(\frac{1}{\mu}) & -O(1) \\
        -O(\frac{\mu}{B})&  O(1)
    \end{array}\right]
\end{align}
and
\begin{align}\label{gamm2n}
    \Gamma_2^{n+1}= \left[\begin{array}{cc}
       \frac{(1+\mu L)^4}{(1-O(\mu))^3} & O(\mu) \\
         O(\frac{\mu^2}{B})&   \rho(P_\alpha)+O(\frac{\mu^2}{B})
    \end{array}\right]^{n+1} =  \left[\begin{array}{cc}
       (\frac{(1+\mu L)^4}{(1-O(\mu))^3})^{n+1} +o(\mu) & O(\mu) \\
         O(\frac{\mu^2}{B})&  ( \rho(P_\alpha)+O(\frac{\mu^2}{B}))^{n+1} + o(\mu) 
    \end{array}\right] 
\end{align}
so that
\begin{align}\label{gamma_22}
    I - \Gamma_2^{n+1} =  \left[\begin{array}{cc}
       1 - (\frac{(1+\mu L)^4}{(1-O(\mu))^3})^{n+1} +o(\mu) & -O(\mu) \\
         -O(\frac{\mu^2}{B})&  1 - ( \rho(P_\alpha)+O(\frac{\mu^2}{B}))^{n+1} + o(\mu) 
    \end{array}\right] 
\end{align}

Similar to $\Gamma_1$, by resorting to Lemma 2 in \cite{VlaskiS21}, and with $n\le O(\frac{1}{\mu})$, we have
\begin{align}\label{ste_w4}
    \left(\frac{(1+\mu L)^4}{(1-O(\mu))^3}\right)^{n+1}   = O(1), \quad\quad 1- \left(\frac{(1+\mu L)^4}{(1-O(\mu))^3}\right)^{n+1}   = -O(1)
\end{align}
so that
\begin{align}\label{gammns}
    I - \Gamma_2^{n+1}  =  \left[\begin{array}{cc}
       -O(1) & -O(\mu) \\
         -O(\frac{\mu^2}{B})&  O(1) 
    \end{array}\right]
\end{align}
\begin{align}\label{gammans_2}
    \Gamma_2^{n+1}= \left[\begin{array}{cc}
       O(1) & O(\mu) \\
         O(\frac{\mu^2}{B})&  O(1)
    \end{array}\right] 
\end{align}

Substituting (\ref{wb4}), (\ref{wc4}), (\ref{gamma_21}), (\ref{gammns}) and (\ref{gammans_2}) into (\ref{f_w42}), we obtain
\begin{align}
    \mathds{E}\left[\begin{array}{c}
         \Vert\bar{\boldsymbol{w}}_{n}\Vert^4\\
          \Vert\check{\boldsymbol{w}}_{n}\Vert^4
    \end{array}\right]\le o(\frac{\mu^2}{B^2}) + \left[\begin{array}{cc}
       -O(\frac{1}{\mu}) & -O(1) \\
         -O(\frac{\mu}{B})&  O(1)
    \end{array}\right]\left[\begin{array}{cc}
       -O(1) & -O(\mu) \\
        - O(\frac{\mu^2}{B})&  O(1) 
    \end{array}\right]\left[\begin{array}{c}
         O(\frac{\mu^3}{B^2}) + {O(\frac{\mu^5}{B})}\\
         O(\mu^4)
    \end{array}\right]\le \left[\begin{array}{c}
         O(\frac{\mu^2}{B^2}) + O(\mu^4)\\
         O(\mu^4)
    \end{array}\right]
\end{align}
and, therefore,
\begin{align}\label{ffw4}
\mathds{E}\Vert{\widetilde{\boldsymbol{\scriptstyle\mathcal{W}}}}_{n}\Vert^4 &\le \Vert\mathcal{V}\Vert^4\mathds{E}\Vert{\mathcal{V}^{\sf T}\widetilde{\boldsymbol{\scriptstyle\mathcal{W}}}}_{n}\Vert^4 = \Vert\mathcal{V}\Vert^4 \mathds{E}\left(\Vert\bar{\boldsymbol{w}}_{n}\Vert^2 + \Vert\check{\boldsymbol{w}}_{n}\Vert^2\right)^2 \overset{(a)}{\le} 2\Vert\mathcal{V}\Vert^4 \mathds{E}\left(\Vert\bar{\boldsymbol{w}}_{n}\Vert^4 + \Vert\check{\boldsymbol{w}}_{n}\Vert^4\right) \notag\\
   &\le O(\frac{\mu^2}{B^2}) + O(\frac{\mu^4}{B}) = O(\mu^{2\gamma})
\end{align}
where $(a)$ follows from Jensen's inequality, and we observe from (\ref{wn2_2}) and (\ref{ffw4}) that
\begin{align}\label{ew2}
\mathds{E}\Vert{\widetilde{\boldsymbol{\scriptstyle\mathcal{W}}}}_{n}\Vert^2 = O((\mathds{E}\Vert{\widetilde{\boldsymbol{\scriptstyle\mathcal{W}}}}_{n}\Vert^4)^{\frac{1}{2}})
\end{align}
Also, by using the Jensen's inequality again, we have
 \begin{align}
\mathds{E}\Vert{\widetilde{\boldsymbol{\scriptstyle\mathcal{W}}}}_{n}\Vert^3 = \mathds{E}(\Vert{\widetilde{\boldsymbol{\scriptstyle\mathcal{W}}}}_{n}\Vert^4)^{\frac{3}{4}} \le (\mathds{E}\Vert{\widetilde{\boldsymbol{\scriptstyle\mathcal{W}}}}_{n}\Vert^4)^{\frac{3}{4}} = O(\mu^{1.5\gamma})
 \end{align}

As for the \emph{centralized method}, we substitute (\ref{wn2_centra}) into (\ref{f_barw}) and iterate it, then we obtain
\begin{align}
\label{f_bar_s4_centra}
    \mathds{E}\Vert\bar{\boldsymbol{w}}_n\Vert^4 {\le} \left(\frac{(1+\mu L)^4}{(1-O(\mu))^3} + O(\frac{\mu^2}{B})\right)^{n+1}\mathds{E}\Vert\bar{\boldsymbol{w}}_{-1}\Vert^4  + \frac{1 - \left(\frac{(1+\mu L)^4}{(1-O(\mu))^3}\right)^{n+1} }{1 - \frac{(1+\mu L)^4}{(1-O(\mu))^3}}\times O(\frac{\mu^3}{B^2})
\end{align}
with which and (\ref{ste_w4}), for the centralized method, it holds that
\begin{align}
\label{wn4_centra}
&\mathds{E}\Vert{\widetilde{\boldsymbol{\scriptstyle\mathcal{W}}}}_{n}\Vert^4  \le O(\frac{\mu^2}{B^2}) = O(\mu^{2(1+\eta)})\\
\label{wn3_centra}
&\mathds{E}\Vert{\widetilde{\boldsymbol{\scriptstyle\mathcal{W}}}}_{n}\Vert^3 \le (\mathds{E}\Vert{\widetilde{\boldsymbol{\scriptstyle\mathcal{W}}}}_{n}\Vert^4)^{\frac{3}{4}} \le O(\mu^{1.5(1+\eta)})
\end{align}

\section{Proof for Lemma \ref{mse} and \ref{mse_centra}: Approximation error of the short-term model}\label{aest}
The argument is similar to \cite{sayed2014adaptation} except that we now focus on nonconvex (as opposed to convex) risk functions. To clarify how far the algorithm can escape from a local minimum, it is necessary to assess the size of the distance between $\boldsymbol{w}_{k,n}$ and $w^\star$ rather than upper bound it. However, the dependence of $\mathcal{H}_{n-1}$ on ${{\boldsymbol{\scriptstyle\mathcal{W}}}}_{n-1}$ makes the analysis difficult.  This motivates the short-term model in (\ref{uni_dis_a}). Consider 
\begin{align}
   \mathcal{C} \overset{\Delta}{=}  \mathcal{A}_2(\mathcal{A}_1- \mu \mathcal{H}) = \mathcal{A} - \mu\mathcal{A}_2\mathcal{H} 
\end{align}
and recall the short-term model in (\ref{uni_dis_a}):
\begin{align}\label{diff_ba}
{\widetilde{\boldsymbol{\scriptstyle\mathcal{W}}}}_{n}' = \mathcal{C}{\widetilde{\boldsymbol{\scriptstyle\mathcal{W}}}}_{n-1}' + \mu \mathcal{A}_2 d + \mu \mathcal{A}_2\boldsymbol{s}_n^B 
\end{align}
where the Hessian matrix $H_{k,n}(\boldsymbol{w}_{k,n})$ is approximated by the Hessian matrix at $w^\star$, and ${\widetilde{\boldsymbol{\scriptstyle\mathcal{W}}}}_{-1}' = {\widetilde{\boldsymbol{\scriptstyle\mathcal{W}}}}_{-1}$. 

In this section, we analyze the approximation error caused by the short-term model. Consider $\boldsymbol{z}_n =\widetilde{\boldsymbol{\scriptstyle\mathcal{W}}}_n' - \widetilde{\boldsymbol{\scriptstyle\mathcal{W}}}_n$ which measures the difference between the true model and the short-term model. Subtracting (\ref{uni_dis}) and (\ref{diff_ba}), we obtain:
\begin{align}\label{zn}
    \boldsymbol{z}_n = \mathcal{C}\boldsymbol{z}_{n-1}+ \mu \mathcal{A}_2(\mathcal{H}_{n-1} - \mathcal{H})\widetilde{\boldsymbol{\scriptstyle\mathcal{W}}}_{n-1} 
\end{align}
Note that according to the Taylor expansion technique, when $\boldsymbol{w}_{k,n}$ is sufficiently close to $w^\star$, we have:
\begin{align}
    H_{k,n}(\boldsymbol{w}_{k,n}) - H_k^\star = - \nabla H_k^\star\tilde{\boldsymbol{w}}_{k,n} + O(\Vert\tilde{\boldsymbol{w}}_{k,n}\Vert^2)
\end{align}
from which we get
\begin{align}
    \Vert(H_{k,n}(\boldsymbol{w}_{k,n}) - H_k^\star)\tilde{\boldsymbol{w}}_{k,n}\Vert \le \Vert \nabla H_k^\star\Vert\Vert\tilde{\boldsymbol{w}}_{k,n}\Vert^2 +O(\Vert\tilde{\boldsymbol{w}}_{k,n}\Vert^3)
\end{align}
so that
\begin{align}\label{h_e}
   \Vert (\mathcal{H}_{n-1} - \mathcal{H})\widetilde{\boldsymbol{\scriptstyle\mathcal{W}}}_{n-1}\Vert \le O(\Vert\widetilde{\boldsymbol{\scriptstyle\mathcal{W}}}_{n-1}\Vert^2)
\end{align}

We now analyze the size of $\mathds{E}\Vert\boldsymbol{z}_n \Vert^2$. Substituting (\ref{decomp_A}) into (\ref{zn}), we have
\begin{align}\label{de_z}
    \boldsymbol{z}_n &= (\mathcal{V}\mathcal{P}\mathcal{V}^{\sf T} - \mu \mathcal{A}_2\mathcal{H})\boldsymbol{z}_{n-1}+\mu\mathcal{A}_2(\mathcal{H}_{n-1} - \mathcal{H})\widetilde{\boldsymbol{\scriptstyle\mathcal{W}}}_{n-1} \notag\\
    &=  \mathcal{V}(\mathcal{P} - \mu \mathcal{V}^{\sf T}\mathcal{A}_2\mathcal{H}\mathcal{V})\mathcal{V}^{\sf T}\boldsymbol{z}_{n-1}+\mu\mathcal{A}_2(\mathcal{H}_{n-1} - \mathcal{H})\widetilde{\boldsymbol{\scriptstyle\mathcal{W}}}_{n-1} 
\end{align}
Similar to appendix \ref{mse2}, we multiply both sides of (\ref{de_z}) by $\mathcal{V}^{\sf T}$ so that it can be decomposed as
\begin{align}
         \mathcal{V}^{\sf T}{{\boldsymbol{{z}}}}_{n} = &\left[\begin{array}{c}
         \frac{1}{\sqrt{K}}\mathds{1}^{\sf  T}\boldsymbol{z}_n \\
         \mathcal{V}_\alpha^{\sf T}\boldsymbol{z}_n 
    \end{array}\right] \overset{\Delta}{=} \left[\begin{array}{c}
         \bar{\boldsymbol{z}}_n  \\
          \check{\boldsymbol{z}}_n
    \end{array}\right] \notag\\
    =&  \left(\left[\begin{array}{cc}
        I_M & 0 \\
        0 & \mathcal{P}_\alpha
    \end{array}\right] - \mu\left[\begin{array}{c}\frac{1}{\sqrt{K}}\mathds{1}^{\sf  T}\\
    \mathcal{V}_\alpha^{\sf T}
    \end{array}\right]\mathcal{A}_2\mathcal{H}\left[\frac{1}{\sqrt{K}}\mathds{1} \quad \mathcal{V}_\alpha\right]\right) \left[\begin{array}{c}
    \bar{\boldsymbol{z}}_{n-1}  \\
          \check{\boldsymbol{z}}_{n-1}
    \end{array}\right] + \mu \mathcal{V}^{\sf T}\mathcal{A}_2(\mathcal{H}_{n-1} - \mathcal{H})\widetilde{\boldsymbol{\scriptstyle\mathcal{W}}}_{n-1}
\end{align}
Consider
\begin{equation}
    \bar{H} = \frac{1}{K}\mathds{1}^{\sf  T}\mathcal{H}\mathds{1} = \frac{1}{K}\sum_k H_{k}(w^\star)
\end{equation} 
which is positive-definite since it is the Hessian matrix of the global risk $J$ at a local minimizer. Then we have
\begin{align}
    \bar{\boldsymbol{z}}_n &= (I_M - \mu\bar{H})\bar{\boldsymbol{z}}_{n-1} - \frac{\mu}{\sqrt{K}}\mathds{1}^{\sf  T}\mathcal{H}\mathcal{V}_\alpha\check{\boldsymbol{z}}_{n-1}+ \frac{\mu}{\sqrt{K}}\mathds{1}^{\sf  T}(\mathcal{H}_{n-1} - \mathcal{H})\widetilde{\boldsymbol{\scriptstyle\mathcal{W}}}_{n-1}\notag\\
    \check{\boldsymbol{z}}_{n} &= (\mathcal{P}_{\alpha} - \mu\mathcal{V}_{\alpha}^{\sf T}\mathcal{A}_2\mathcal{H}\mathcal{V}_\alpha)\check{\boldsymbol{z}}_{n-1} - \frac{\mu}{\sqrt{K}}\mathcal{V}_{\alpha}^{\sf T}\mathcal{A}_2\mathcal{H}\mathds{1}\bar{\boldsymbol{z}}_{n-1} + \mu\mathcal{V}_{\alpha}^{\sf T}\mathcal{A}_2(\mathcal{H}_{n-1} - \mathcal{H})\widetilde{\boldsymbol{\scriptstyle\mathcal{W}}}_{n-1}
\end{align}
Still, for the centralized method, we only need to analyze $\bar{\boldsymbol{z}}_n $  as now $\check{\boldsymbol{z}}_{n}$ is always $0$.

We first analyze $\mathds{E}\Vert\bar{\boldsymbol{z}}_n\Vert^2$. Let $t = \Vert I_M - \mu \bar{H}\Vert = 1 - O(\mu) > 0$, we have
\begin{align}\label{s_barz}
\mathds{E}\Vert \bar{\boldsymbol{z}}_{n}\Vert^2 = &\mathds{E}\Vert (I_M - \mu \bar{H})\bar{\boldsymbol{z}}_{n-1} - \frac{\mu}{\sqrt{K}}\mathds{1}^{\sf T}\mathcal{H}\mathcal{V}_{\alpha}\check{\boldsymbol{z}}_{n-1} + \mu\frac{1}{\sqrt{K}}\mathds{1}^{\sf  T}(\mathcal{H}_{n-1} - \mathcal{H})\widetilde{\boldsymbol{\scriptstyle\mathcal{W}}}_{n-1}\Vert^2 \notag\\
\overset{(a)}{\le} & \frac{1}{t}\mathds{E}\Vert(I_M - \mu \bar{H})\bar{\boldsymbol{z}}_{n-1} \Vert^2+ \frac{O(\mu^2)}{1-t}\mathds{E}\Vert\check{\boldsymbol{z}}_{n-1}\Vert^2+ \frac{O(\mu^2)}{1-t}\mathds{E}\Vert\widetilde{\boldsymbol{\scriptstyle\mathcal{W}}}_{n-1}\Vert^4\notag\\
\overset{(b)}{=} & (1 - O(\mu))\mathds{E}\Vert\bar{\boldsymbol{z}}_{n-1}\Vert^2 + O(\mu)\mathds{E}\Vert\check{\boldsymbol{z}}_{n-1}\Vert^2 + O(\mu^{1+2\gamma})
\end{align}
where $(a)$ follows from Jensen's inequality and (\ref{h_e}), and $(b)$ follows from (\ref{ffw4}).

Next, we analyze $\mathds{E}\Vert\check{\boldsymbol{z}}_n\Vert^2$. Let $t = \rho(P_\alpha)$ which is the spectral radius of $P_\alpha$, we have
\begin{align}\label{s_checkz}
    \mathds{E}\Vert\check{\boldsymbol{z}}_{n} \Vert^2 =& \mathds{E}\Vert (\mathcal{P}_{\alpha} - \mu\mathcal{V}_{\alpha}^{\sf T}\mathcal{A}_2\mathcal{H}\mathcal{V}_\alpha)\check{\boldsymbol{z}}_{n-1} - \frac{\mu}{\sqrt{K}}\mathcal{V}_{\alpha}^{\sf T}\mathcal{A}_2\mathcal{H}\mathds{1}\bar{\boldsymbol{z}}_{n-1} + \mu\mathcal{V}_{\alpha}^{\sf T}\mathcal{A}_2(\mathcal{H}_{n-1} - \mathcal{H})\widetilde{\boldsymbol{\scriptstyle\mathcal{W}}}_{n-1}\Vert^2\notag\\
    \overset{(a)}{\le}& t\mathds{E}\Vert \check{\boldsymbol{z}}_{n-1}\Vert^2 + O(\mu^2)\mathds{E}\Vert\check{\boldsymbol{z}}_{n-1}\Vert^2 + O(\mu^2)\mathds{E}\Vert\bar{\boldsymbol{z}}_{n-1}\Vert^2+ O(\mu^2)\mathds{E}\Vert\widetilde{\boldsymbol{\scriptstyle\mathcal{W}}}_{n-1}\Vert^4\notag\\
\le &  (\rho(P_\alpha)+O(\mu^2))\mathds{E}\Vert \check{\boldsymbol{z}}_{n-1}\Vert^2 + O(\mu^2)\mathds{E}\Vert\bar{\boldsymbol{z}}_{n-1}\Vert^2+ O(\mu^{2+2\gamma})
\end{align}
where $(a)$ follows from Jensen's inequality, and $(b)$ follows from (\ref{ffw4}). By combining (\ref{s_barz}) and (\ref{s_checkz}), we obtain the recursion associated with the size of  $\mathds{E}\Vert\mathcal{V}_\alpha^{\sf T}{{\boldsymbol{{z}}}}_{n}\Vert^2$:
\begin{align}\label{szn}
   \mathds{E}\left[\begin{array}{c}
         \Vert\bar{\boldsymbol{z}}_{n}\Vert^2\\
          \Vert\check{\boldsymbol{z}}_{n}\Vert^2
    \end{array}\right]\le \Gamma' \mathds{E}\left[\begin{array}{c}
         \Vert\bar{\boldsymbol{z}}_{n-1}\Vert^2\\
          \Vert\check{\boldsymbol{z}}_{n-1}\Vert^2
    \end{array}\right] + \left[\begin{array}{c}
        O(\mu^{1+2\gamma}) \\
          O(\mu^{2+2\gamma})
    \end{array}\right] 
\end{align}
where $\boldsymbol{z}_{-1} = 0$, and
\begin{align}\label{gammap}
    \Gamma' = \left[\begin{array}{cc}
       1 - O(\mu) & O(\mu) \\
         O(\mu^2)&  \rho(P_\alpha)+O(\mu^2)
    \end{array}\right]
\end{align}
By iterating (\ref{szn}), we have
\begin{align}\label{szn_2}
    \mathds{E}\left[\begin{array}{c}
         \Vert\bar{\boldsymbol{z}}_{n}\Vert^2\\
          \Vert\check{\boldsymbol{z}}_{n}\Vert^2
    \end{array}\right] = (I - \Gamma')^{-1}(I - \Gamma'^{n+1})\left[\begin{array}{c}
        O(\mu^{1+2\gamma}) \\
          O(\mu^{2+2\gamma})
    \end{array}\right]  
\end{align}
for which it can be verified that
\begin{align}\label{gammapin}
    (I - \Gamma')^{-1} =  \left[\begin{array}{cc}
       O(\mu) & -O(\mu) \\
         -O(\mu^2)&  1 - \rho(P_\alpha)+O(\mu^2)
    \end{array}\right]^{-1} = \left[\begin{array}{cc}
       O(\frac{1}{\mu}) & O(1) \\
         O(\mu)&  O(1)
    \end{array}\right]
\end{align}
and 
\begin{align}\label{gammapn}
    I - \Gamma'^{n+1} =  \left[\begin{array}{cc}
       1 - (1 - O(\mu))^{n+1} & -O(\mu) \\
         -O(\mu^2)&  1 - (\rho(P_\alpha)+O(\mu^2))^{n+1}
    \end{array}\right] = \left[\begin{array}{cc}
       1 - (1 - O(\mu))^{n+1} & -O(\mu) \\
         -O(\mu^2)&  O(1)
    \end{array}\right]
\end{align}
Again, similar to (\ref{o1ns}), we have:
\begin{align}\label{1omu}
   1 - (1 - O(\mu))^{n+1} < 1 = O(1) 
\end{align}
Then we substitute (\ref{gammapin}), (\ref{gammapn}) and (\ref{1omu}) into (\ref{szn_2}), and obtain
\begin{align}\label{ew4}
    \mathds{E}\left[\begin{array}{c}
         \Vert\bar{\boldsymbol{z}}_{n}\Vert^2\\
          \Vert\check{\boldsymbol{z}}_{n}\Vert^2
    \end{array}\right]\le&  \left[\begin{array}{cc}
       O(\frac{1}{\mu}) & O(1) \\
         O(\mu)&  O(1)
    \end{array}\right] \left[\begin{array}{cc}
       O(1) & -O(\mu) \\
         -O(\mu^2)&  O(1)
    \end{array}\right] \left[\begin{array}{c}
        O(\mu^{1+2\gamma})\\
          O(\mu^{2+2\gamma})\end{array}\right] \le \left[\begin{array}{c}
        O(\mu^{2\gamma})\\
          O(\mu^{2+2\gamma})\end{array}\right] 
\end{align}
from which we have
\begin{align}\label{ew4_s}
\mathds{E}\Vert\widetilde{\boldsymbol{\scriptstyle\mathcal{W}}}_n' - \widetilde{\boldsymbol{\scriptstyle\mathcal{W}}}_n \Vert^2 = \mathds{E}  \Vert\mathcal{V}^{-\sf T}\mathcal{V}^{\sf T}{{\boldsymbol{{z}}}}_{n}\Vert^2 \le \Vert\mathcal{V}^{-\sf T}\Vert^2 \mathds{E}(\Vert\check{\boldsymbol{z}}_{n}\Vert^2+\Vert\bar{\boldsymbol{z}}_{n}\Vert^2) = O(\mu^{2\gamma})
\end{align}

Since
\begin{align}\label{dian}
\mathds{E}\Vert\widetilde{\boldsymbol{\scriptstyle\mathcal{W}}}_n' \Vert^2 &= \mathds{E}\Vert\widetilde{\boldsymbol{\scriptstyle\mathcal{W}}}_n' -\widetilde{\boldsymbol{\scriptstyle\mathcal{W}}}_n+\widetilde{\boldsymbol{\scriptstyle\mathcal{W}}}_n\Vert^2\notag\\
&\le \mathds{E}\Vert\widetilde{\boldsymbol{\scriptstyle\mathcal{W}}}_n'- \widetilde{\boldsymbol{\scriptstyle\mathcal{W}}}_n\Vert^2 + \mathds{E}\Vert\widetilde{\boldsymbol{\scriptstyle\mathcal{W}}}_n\Vert^2 + 2\vert\mathds{E}(\widetilde{\boldsymbol{\scriptstyle\mathcal{W}}}_n' -\widetilde{\boldsymbol{\scriptstyle\mathcal{W}}}_n)^{\sf T}\widetilde{\boldsymbol{\scriptstyle\mathcal{W}}}_n\vert\notag\\
    &\le \mathds{E}\Vert\widetilde{\boldsymbol{\scriptstyle\mathcal{W}}}_n'- \widetilde{\boldsymbol{\scriptstyle\mathcal{W}}}_n\Vert^2 + \mathds{E}\Vert\widetilde{\boldsymbol{\scriptstyle\mathcal{W}}}_n\Vert^2 + 2\sqrt{\mathds{E}\Vert\widetilde{\boldsymbol{\scriptstyle\mathcal{W}}}_n'- \widetilde{\boldsymbol{\scriptstyle\mathcal{W}}}_n\Vert^2\mathds{E}\Vert\widetilde{\boldsymbol{\scriptstyle\mathcal{W}}}_n\Vert^2}\notag\\
    & \overset{(a)}{\le} O(\mu^\gamma)
\end{align}
where in $(a)$ we use the results of (\ref{wn2_2}) and (\ref{ew4_s}), then we have
\begin{align}\label{ewwp}
\mathds{E}\Vert\widetilde{\boldsymbol{\scriptstyle\mathcal{W}}}_n' \Vert^2 - \mathds{E}\Vert\widetilde{\boldsymbol{\scriptstyle\mathcal{W}}}_n\Vert^2 &\le \mathds{E}\Vert\widetilde{\boldsymbol{\scriptstyle\mathcal{W}}}_n'- \widetilde{\boldsymbol{\scriptstyle\mathcal{W}}}_n\Vert^2 + 2\sqrt{\mathds{E}\Vert\widetilde{\boldsymbol{\scriptstyle\mathcal{W}}}_n'- \widetilde{\boldsymbol{\scriptstyle\mathcal{W}}}_n\Vert^2\mathds{E}\Vert\widetilde{\boldsymbol{\scriptstyle\mathcal{W}}}_n\Vert^2} \notag\\
    & \overset{(a)}{=} O(\mathds{E}\Vert\widetilde{\boldsymbol{\scriptstyle\mathcal{W}}}_{n-1}\Vert^4) + O(\sqrt{\mathds{E}\Vert\widetilde{\boldsymbol{\scriptstyle\mathcal{W}}}_{n-1}\Vert^4\cdot(\mathds{E}\Vert\widetilde{\boldsymbol{\scriptstyle\mathcal{W}}}_{n-1}\Vert^4)^{\frac{1}{2}}}) \notag\\
    &= O(\mu^{1.5\gamma})
\end{align}
where in $(a)$ we use the results of $(\ref{ew2})$ and $(\ref{ew4})$. 
Similar to (\ref{dian}), we have
\begin{align}\label{dian_2}
\mathds{E}\Vert\widetilde{\boldsymbol{\scriptstyle\mathcal{W}}}_n \Vert^2 &= \mathds{E}\Vert\widetilde{\boldsymbol{\scriptstyle\mathcal{W}}}_n -\widetilde{\boldsymbol{\scriptstyle\mathcal{W}}}_n'+\widetilde{\boldsymbol{\scriptstyle\mathcal{W}}}_n'\Vert^2\notag\\
&\le \mathds{E}\Vert\widetilde{\boldsymbol{\scriptstyle\mathcal{W}}}_n- \widetilde{\boldsymbol{\scriptstyle\mathcal{W}}}_n'\Vert^2 + \mathds{E}\Vert\widetilde{\boldsymbol{\scriptstyle\mathcal{W}}}_n'\Vert^2 + 2\vert\mathds{E}(\widetilde{\boldsymbol{\scriptstyle\mathcal{W}}}_n -\widetilde{\boldsymbol{\scriptstyle\mathcal{W}}}_n')^{\sf T}\widetilde{\boldsymbol{\scriptstyle\mathcal{W}}}_n'\vert\notag\\
    &\le \mathds{E}\Vert\widetilde{\boldsymbol{\scriptstyle\mathcal{W}}}_n'- \widetilde{\boldsymbol{\scriptstyle\mathcal{W}}}_n\Vert^2 + \mathds{E}\Vert\widetilde{\boldsymbol{\scriptstyle\mathcal{W}}}_n'\Vert^2 + 2\sqrt{\mathds{E}\Vert\widetilde{\boldsymbol{\scriptstyle\mathcal{W}}}_n'- \widetilde{\boldsymbol{\scriptstyle\mathcal{W}}}_n\Vert^2\mathds{E}\Vert\widetilde{\boldsymbol{\scriptstyle\mathcal{W}}}_n'\Vert^2}
\end{align}
from which we have
\begin{align}\label{ewwp_2}
\mathds{E}\Vert\widetilde{\boldsymbol{\scriptstyle\mathcal{W}}}_n \Vert^2 - \mathds{E}\Vert\widetilde{\boldsymbol{\scriptstyle\mathcal{W}}}_n'\Vert^2 &\le \mathds{E}\Vert\widetilde{\boldsymbol{\scriptstyle\mathcal{W}}}_n'- \widetilde{\boldsymbol{\scriptstyle\mathcal{W}}}_n\Vert^2 + 2\sqrt{\mathds{E}\Vert\widetilde{\boldsymbol{\scriptstyle\mathcal{W}}}_n'- \widetilde{\boldsymbol{\scriptstyle\mathcal{W}}}_n\Vert^2\mathds{E}\Vert\widetilde{\boldsymbol{\scriptstyle\mathcal{W}}}_n'\Vert^2} \le O(\mu^{1.5\gamma})
\end{align}
Combining (\ref{ewwp}) and (\ref{ewwp_2}), we have
\begin{align}\label{ewwp_4}
    \vert \mathds{E}\Vert\widetilde{\boldsymbol{\scriptstyle\mathcal{W}}}_n' \Vert^2 - \mathds{E}\Vert\widetilde{\boldsymbol{\scriptstyle\mathcal{W}}}_n\Vert^2 \vert \le O(\mu^{1.5\gamma})
\end{align}
which means that the approximation error of replacing $\mathds{E}\Vert\widetilde{\boldsymbol{\scriptstyle\mathcal{W}}}_n \Vert^2$ by $\mathds{E}\Vert\widetilde{\boldsymbol{\scriptstyle\mathcal{W}}}_n' \Vert^2$ can be omitted compared with the size of $\mathds{E}\Vert\widetilde{\boldsymbol{\scriptstyle\mathcal{W}}}_n \Vert^2$. In other words, $\mathds{E}\Vert\widetilde{\boldsymbol{\scriptstyle\mathcal{W}}}_n' \Vert^2$ carries sufficient information of $\mathds{E}\Vert\widetilde{\boldsymbol{\scriptstyle\mathcal{W}}}_n\Vert^2$  .

Similarly, for the \emph{centralized method}, we only need to analyze $\mathds{E}\Vert \bar{\boldsymbol{z}}_{n}\Vert^2$. Iterating  (\ref{s_barz}) gives
\begin{align}\label{s_barz_centra}
\mathds{E}\Vert \bar{\boldsymbol{z}}_{n}\Vert^2 
=  \frac{1- (1-O(\mu))^{n+1}}{1- (1 - O(\mu))} \times O(\mu)\mathds{E}\Vert\widetilde{\boldsymbol{\scriptstyle\mathcal{W}}}_{n-1}\Vert^4
\end{align}
We substitute (\ref{wn2_centra}), (\ref{wn4_centra}), (\ref{1omu}) into (\ref{s_barz_centra}), and use (\ref{dian}) and (\ref{dian_2}), then we obtain
\begin{align}
\label{z_centra_1}
    &\mathds{E}\Vert\widetilde{\boldsymbol{\scriptstyle\mathcal{W}}}_n' - \widetilde{\boldsymbol{\scriptstyle\mathcal{W}}}_n \Vert^2 \le O(\mu^{2(1+\eta)})\\
    \label{z_centra_2}
    &\vert\mathds{E}\Vert\widetilde{\boldsymbol{\scriptstyle\mathcal{W}}}_n' \Vert^2 - \mathds{E}\Vert\widetilde{\boldsymbol{\scriptstyle\mathcal{W}}}_n\Vert^2\vert \le O(\mu^{1.5(1+\eta)})
\end{align}

\section{Proof for Theorem \ref{th_er}}\label{per}
In this section, we derive closed-form expressions for the excess-risk performance of centralized, consensus and diffusion strategies over a finite time horizon $n\le O(\frac{1}{\mu})$. Specifically, {we verify how far the algorithms can escape from the local minimum $J(w^\star)$ on average}. To do this, we need to use the upper bounds in Lemmas \ref{mse} and \ref{mse_centra}, where centralized and decentralized methods have different expressions. For simplicity, we use $\gamma'$ to unify the results of decentralized and centralized methods in Lemmas \ref{mse} and \ref{mse_centra}, namely,
\begin{align}\label{3uw}
\mathds{E}\Vert{\widetilde{\boldsymbol{\scriptstyle\mathcal{W}}}}_{n}\Vert^2 \le O(\mu^{\gamma'}), \quad  \mathds{E}\Vert{\widetilde{\boldsymbol{\scriptstyle\mathcal{W}}}}_{n}\Vert^4 \le O(\mu^{2\gamma'}),\quad \vert\mathds{E}\Vert\widetilde{\boldsymbol{\scriptstyle\mathcal{W}}}_n' \Vert^2 - \mathds{E}\Vert\widetilde{\boldsymbol{\scriptstyle\mathcal{W}}}_n\Vert^2\vert \le O(\mu^{1.5\gamma'})
\end{align}
where in the centralized method $\gamma' = {1+\eta}$, while in decentralized methods $\gamma' = \gamma = \min\{1+\eta, 2\}$.    

For each agent, the excess risk corresponding to its model is:
\begin{align}
    \mathrm{ER}_{k,n} &= \mathds{E}J(\boldsymbol{w}_{k,n}) - J(w^\star) \notag\\
    &= \frac{1}{K}\sum\limits_{\ell=1}^{K}  \left(\mathds{E}J_\ell(\boldsymbol{w}_{k,n}) -  J_\ell(w^\star) \right)\notag\\
    & \overset{(a)}{=} \frac{1}{K}\sum\limits_{\ell=1}^{K}  \left(\nabla J_\ell(w^\star) + \frac{1}{2}\mathds{E}(\boldsymbol{w}_{k,n} - w^\star)^{\sf T}H_\ell^\star(\boldsymbol{w}_{k,n} - w^\star) \pm O(\mathds{E}\Vert\boldsymbol{w}_{k,n} - w^\star\Vert^3)\right)\notag\\
    & \overset{(b)}{=} {\frac{1}{2}}\mathds{E}\Vert\boldsymbol{w}_{k,n} - w^\star\Vert^2_{\bar{H}} \pm O(\mathds{E}\Vert\boldsymbol{w}_{k,n} - w^\star\Vert^3)
\end{align}
where $(a)$ follows from the Taylor expansion, and $(b)$ follows from (\ref{g0}). {Then we have the following average excess risk (or escaping efficiency) across the network involving models of all agents:
\begin{align}\label{erave_p_prime}
    \mathrm{ER}_n = \frac{1}{K}\sum\limits_k \mathrm{ER}_{k,n} \overset{(a)}{=} \frac{1}{2K}\mathds{E}\Vert{\widetilde{\boldsymbol{\scriptstyle\mathcal{W}}}}_{n}\Vert^2_{I_K\otimes\bar{H}}\pm O((\mathds{E}\Vert{\widetilde{\boldsymbol{\scriptstyle\mathcal{W}}}}_{n}\Vert^4)^\frac{3}{4}) =  \frac{1}{2K}\mathds{E}\Vert{\widetilde{\boldsymbol{\scriptstyle\mathcal{W}}}}_{n}\Vert^2_{I_K\otimes\bar{H}} \pm O(\mu^{1.5\gamma'})
\end{align}
where the $O(\mu^{1.5\gamma'})$ term can be omitted according to (\ref{3uw}), and $(a)$ follows from the Taylor expansion and the following inequality:
\begin{align}
    \mathds{E}\Vert\boldsymbol{w}_k - w^\star\Vert^3 = \mathds{E}(\Vert\boldsymbol{w}_k - w^\star\Vert^4)^\frac{3}{4} \le (\mathds{E}\Vert\boldsymbol{w}_k - w^\star\Vert^4)^\frac{3}{4} \le (\mathds{E}\Vert{\widetilde{\boldsymbol{\scriptstyle\mathcal{W}}}}_{n}\Vert^4)^\frac{3}{4}\le O(\mu^{1.5\gamma'})
\end{align}}

{Similar to (\ref{w+w-c_sum}), we first clarify the relationship between $\mathrm{ER}_n$ and the consensus distance. Recalling equality (\ref{diff_2}), we have
\begin{align}\label{t_w_sum_h}
\Vert{\widetilde{\boldsymbol{\scriptstyle\mathcal{W}}}}_{n}\Vert^2_{I_K\otimes\bar{H}} = \left\Vert\left[\begin{array}{c}
\bar{\boldsymbol{w}}_n  \\
     \check{\boldsymbol{w}}_n 
\end{array}\right]\right\Vert^2 _{I_K\otimes\bar{H}} = \Vert\bar{\boldsymbol{w}}_n \Vert^2_{\bar{H}} + \Vert\check{\boldsymbol{w}}_n\Vert^2_{I_{K-1}\otimes\bar{H}}
\end{align}
For the term of $\bar{\boldsymbol{w}}_n$, we have
\begin{align}\label{w_bar_h}
  \Vert\bar{\boldsymbol{w}}_n \Vert^2_{\bar{H}} \overset{(a)}{=} \Vert\sqrt{K}\tilde{\boldsymbol{w}}_{c,n}\Vert^2_{\bar{H}} = K\Vert \tilde{\boldsymbol{w}}_{c,n} \Vert^2_{\bar{H}} = \Vert {\widetilde{\boldsymbol{\scriptstyle\mathcal{W}}}}_{c,n}\Vert_{I_K\otimes \bar{H}} 
\end{align}
where $(a)$ follows from (\ref{r_bw_cw}), and the block vector ${\widetilde{\boldsymbol{\scriptstyle\mathcal{W}}}}_{c,n}$ defined in (\ref{w_cd}) is a collection of $\tilde{\boldsymbol{w}}_{c,n}$. Also, for $\check{\boldsymbol{w}}_n$, we obtain
\begin{align}\label{w_check_h}
    \Vert\check{\boldsymbol{w}}_n\Vert^2_{I_{K-1}\otimes\bar{H}} \overset{(a)}{=}\Vert\check{\boldsymbol{w}}_n\Vert^2_{(V_{\alpha}^{\sf T}V_{\alpha})\otimes\bar{H}} = \Vert (V_\alpha\otimes I_M)\check{\boldsymbol{w}}_n\Vert^2_{I_K\otimes \bar{H}} \overset{(b)}{=}\Vert{\widetilde{\boldsymbol{\scriptstyle\mathcal{W}}}}_{n}-{\widetilde{\boldsymbol{\scriptstyle\mathcal{W}}}}_{c,n}\Vert^2_{I_K\otimes\bar{H}}
\end{align}
where $(a)$ follows from (\ref{a_pro}), and $(b)$ follows from (\ref{check_w_2_nov}). Substituting  (\ref{t_w_sum_h}), (\ref{w_bar_h}) and (\ref{w_check_h}) into (\ref{erave_p_prime}), we obtain
\begin{align}\label{er_w_c_w_n}
 \mathrm{ER}_n = \frac{1}{2K}\mathds{E}\Vert{\widetilde{\boldsymbol{\scriptstyle\mathcal{W}}}}_{n}\Vert^2_{I_K\otimes\bar{H}} \pm O(\mu^{1.5\gamma'})  = \frac{1}{2K}\mathds{E}\{\Vert {\widetilde{\boldsymbol{\scriptstyle\mathcal{W}}}}_{c,n}\Vert_{I_K\otimes \bar{H}} + \Vert{\widetilde{\boldsymbol{\scriptstyle\mathcal{W}}}}_{n}-{\widetilde{\boldsymbol{\scriptstyle\mathcal{W}}}}_{c,n}\Vert^2_{I_K\otimes\bar{H}}\} \pm O(\mu^{1.5\gamma'})
\end{align}
Thus, decentralized methods include the additional term caused by consensus distance, which further increases the escaping efficiency compared to the centralized method.
}

{We next analyze the exact expression of $\mathrm{ER}_n$. To do this, we resort to the short-term model ${\widetilde{\boldsymbol{\scriptstyle\mathcal{W}}}}_{n}'$ which can well approximate the true model ${\widetilde{\boldsymbol{\scriptstyle\mathcal{W}}}}_{n}$. Basically, we have}
\begin{align}\label{erave_p}
    \mathrm{ER}_n = \frac{1}{2K}\mathds{E}\Vert{\widetilde{\boldsymbol{\scriptstyle\mathcal{W}}}}_{n}\Vert^2_{I\otimes\bar{H}}\pm O((\mathds{E}\Vert{\widetilde{\boldsymbol{\scriptstyle\mathcal{W}}}}_{n}\Vert^4)^\frac{3}{4}) \overset{(a)}{=} \frac{1}{2K}\mathds{E}\Vert{\widetilde{\boldsymbol{\scriptstyle\mathcal{W}}}}_{n}'\Vert^2_{I\otimes\bar{H}} \pm O(\mu^{1.5\gamma'})
\end{align}
where $(a)$ follows from the following inequality which is similar to (\ref{ewwp}) and (\ref{ewwp_2}):
\begin{align}\label{ewwph}
\mathds{E}\Vert\widetilde{\boldsymbol{\scriptstyle\mathcal{W}}}_n' \Vert^2_{I\otimes\bar{H}} - \mathds{E}\Vert\widetilde{\boldsymbol{\scriptstyle\mathcal{W}}}_n\Vert^2_{I\otimes\bar{H}} &\le \mathds{E}\Vert\widetilde{\boldsymbol{\scriptstyle\mathcal{W}}}_n'- \widetilde{\boldsymbol{\scriptstyle\mathcal{W}}}_n\Vert^2_{I\otimes\bar{H}} + 2\Vert\bar{H}\Vert\sqrt{\mathds{E}\Vert\widetilde{\boldsymbol{\scriptstyle\mathcal{W}}}_n'- \widetilde{\boldsymbol{\scriptstyle\mathcal{W}}}_n\Vert^2\mathds{E}\Vert\widetilde{\boldsymbol{\scriptstyle\mathcal{W}}}_n\Vert^2} \notag\\
&\le \Vert\bar{H}\Vert \mathds{E}\Vert\widetilde{\boldsymbol{\scriptstyle\mathcal{W}}}_n'- \widetilde{\boldsymbol{\scriptstyle\mathcal{W}}}_n\Vert^2 + 2\Vert\bar{H}\Vert\sqrt{\mathds{E}\Vert\widetilde{\boldsymbol{\scriptstyle\mathcal{W}}}_n'- \widetilde{\boldsymbol{\scriptstyle\mathcal{W}}}_n\Vert^2\mathds{E}\Vert\widetilde{\boldsymbol{\scriptstyle\mathcal{W}}}_n\Vert^2}\notag\\
    &= O(\mu^{1.5\gamma'})
\end{align}
and
\begin{align}\label{ewwph_2}
\mathds{E}\Vert\widetilde{\boldsymbol{\scriptstyle\mathcal{W}}}_n \Vert^2_{I\otimes\bar{H}} - \mathds{E}\Vert\widetilde{\boldsymbol{\scriptstyle\mathcal{W}}}_n'\Vert^2_{I\otimes\bar{H}} \le  O(\mu^{1.5\gamma'})
\end{align}
so that 
\begin{align}
\mathds{E}\Vert\widetilde{\boldsymbol{\scriptstyle\mathcal{W}}}_n'\Vert^2_{I\otimes\bar{H}} - O(\mu^{1.5\gamma'}) \le \mathds{E}\Vert\widetilde{\boldsymbol{\scriptstyle\mathcal{W}}}_n \Vert^2_{I\otimes\bar{H}} \le \mathds{E}\Vert\widetilde{\boldsymbol{\scriptstyle\mathcal{W}}}_n'\Vert^2_{I\otimes\bar{H}} + O(\mu^{1.5\gamma'})
\end{align}

To proceed, we examine the size of $\mathds{E}\Vert{\widetilde{\boldsymbol{\scriptstyle\mathcal{W}}}}_{n}'\Vert^2_{I\otimes\bar{H}}$.
We start from the short-term model in (\ref{diff_ba}) and iterate it, 
\begin{align}
{\widetilde{\boldsymbol{\scriptstyle\mathcal{W}}}}_{n}' = \mathcal{C}^{n+1}{\widetilde{\boldsymbol{\scriptstyle\mathcal{W}}}}_{-1}' + \mu\sum\limits_{i=0}^{n} \mathcal{C}^{i}\mathcal{A}_2 d + \mu \sum\limits_{i=0}^{n} \mathcal{C}^{i}\mathcal{A}_2\boldsymbol{s}_i^B 
\end{align}
from which we proceed to examine the size of $\mathds{E}\Vert{\widetilde{\boldsymbol{\scriptstyle\mathcal{W}}}}_{n}'\Vert^2_{I\otimes\bar{H}}$ as follows:
\begin{align}\label{wnph1}
\mathds{E}\left\Vert{\widetilde{\boldsymbol{\scriptstyle\mathcal{W}}}}_{n}'\right\Vert^2_{I\otimes\bar{H}} = \mathds{E}\left\Vert\mathcal{C}^{n+1}{\widetilde{\boldsymbol{\scriptstyle\mathcal{W}}}}_{-1}' + \mu\sum\limits_{i=0}^{n} \mathcal{C}^{i}\mathcal{A}_2 d \right\Vert^2_{I\otimes\bar{H}} + \mu^2\mathds{E}\left\Vert\sum\limits_{i=0}^{n} \mathcal{C}^{i}\mathcal{A}_2\boldsymbol{s}_i^B \right\Vert^2_{I\otimes\bar{H}} 
\end{align}
We first examine the term $\mathds{E}\left\Vert\mathcal{C}^{n+1}{\widetilde{\boldsymbol{\scriptstyle\mathcal{W}}}}_{-1}' + \mu\sum\limits_{i=0}^{n} \mathcal{C}^{i}\mathcal{A}_2 d \right\Vert^2_{I\otimes\bar{H}}$. 
To do so, it is necessary to compute $\sum\limits_{i=0}^{n} \mathcal{C}^{i}$:
\begin{align}\label{sumci}
    \sum\limits_{i=0}^{n} \mathcal{C}^{i} = (I - \mathcal{C})^{-1}(I - \mathcal{C}^{n+1})
\end{align}
Recall that 
\begin{align}
    \mathcal{C} =  \mathcal{A} - \mu \mathcal{A}_2\mathcal{H} = \mathcal{V}\mathcal{P}\mathcal{V}^{\sf T} - \mu\mathcal{A}_2\mathcal{H} = \mathcal{V}(\mathcal{P} - \mu\mathcal{V}^{\sf T}\mathcal{A}_2\mathcal{H}\mathcal{V})\mathcal{V}^{\sf T}
\end{align}
and consider
\begin{align}\label{Bbar}
    \bar{\mathcal{C}} = \mathcal{P} - \mu\mathcal{V}^{\sf T}\mathcal{A}_2\mathcal{H}\mathcal{V}^{-\sf T }
\end{align}
then, it holds that
\begin{align}
\label{cm11}
    (I - \mathcal{C})^{-1} &= \mathcal{V}(I - \bar{\mathcal{C}})^{-1}\mathcal{V}^{\sf T}\\
    \label{cn11}
    (I - \mathcal{C}^{n+1}) & = \mathcal{V}(I - \bar{\mathcal{C}}^{n+1})\mathcal{V}^{\sf T}
\end{align}
Substituting  (\ref{vpv1}) and (\ref{ka3})  into (\ref{Bbar}), we obtain
\begin{align}\label{mcbar}
    \bar{\mathcal{C}} = \left[\begin{array}{cc}
        I & 0 \\
        0 & \mathcal{P}_{\alpha}
    \end{array}\right]- \mu\left[\begin{array}{c}
         \frac{1}{\sqrt{K}}\mathds{1}^{\sf T} \\
          \mathcal{V}_\alpha^{\sf T}
\end{array}\right]\mathcal{A}_2\mathcal{H}\left[\begin{array}{cc}
       \frac{1}{\sqrt{K}}\mathds{1}  & \mathcal{V}_\alpha 
    \end{array}\right] = \left[\begin{array}{cc}
        I - \mu\bar{H} & -\mu\frac{1}{\sqrt{K}}\mathds{1}^{\sf T}\mathcal{H}\mathcal{V}_\alpha\\
       -\mu\mathcal{V}_\alpha^{\sf T}\mathcal{A}_2\mathcal{H}\frac{1}{\sqrt{K}}\mathds{1} & \mathcal{P}_{\alpha} - \mu\mathcal{V}_\alpha^{\sf T}\mathcal{A}_2\mathcal{H}\mathcal{V}_\alpha
    \end{array}\right]
\end{align}
from which we have
\begin{align}\label{cbar1}
    (I - \bar{\mathcal{C}})^{-1} = \left[\begin{array}{cc}
       \mu\bar{H} & \mu\frac{1}{\sqrt{K}}\mathds{1}^{\sf T}\mathcal{H}\mathcal{V}_\alpha\\
       \mu\mathcal{V}_\alpha^{\sf T}\mathcal{A}_2\mathcal{H}\frac{1}{\sqrt{K}}\mathds{1} & I - \mathcal{P}_{\alpha} - \mu\mathcal{V}_\alpha^{\sf T}\mathcal{A}_2\mathcal{H}\mathcal{V}_\alpha
    \end{array}\right]^{-1}
\end{align}
We appeal to the block matrix inversion formula:
\begin{align}\label{bmi}
    \left[\begin{array}{cc}
       A  & B \\
       C  & D
    \end{array}\right]^{-1}=  \left[\begin{array}{cc}
       A^{-1}  & 0 \\
       0  & 0
    \end{array}\right] +  \left[\begin{array}{cc}
       A^{-1}B\Delta^{-1}CA^{-1}  & -A^{-1}B\Delta^{-1} \\
       -\Delta^{-1}CA^{-1} & \Delta^{-1}
    \end{array}\right]
\end{align}
where the Schur complement $\Delta$ is defined by
\begin{align}
    \Delta = D - CA^{-1}B
\end{align}
Applying this formula to (\ref{cbar1}), we have
\begin{align}
    \Delta = I - \mathcal{P}_{\alpha} + O(\mu)
\end{align}
and
\begin{align}
    A^{-1} & = \frac{1}{\mu}\bar{H}^{-1}\\
    \Delta^{-1} &= ( I - \mathcal{P}_{\alpha} + O(\mu))^{-1}\\
    A^{-1}B\Delta^{-1}CA^{-1} & = \frac{1}{K}\bar{H}^{-1}\mathds{1}^{\sf T}\mathcal{H}\mathcal{V}_\alpha( I - \mathcal{P}_{\alpha} + O(\mu))^{-1}\mathcal{V}_\alpha^{\sf T}\mathcal{A}_2\mathcal{H}\mathds{1}\bar{H}^{-1}\\
    -A^{-1}B\Delta^{-1} &= -\frac{1}{\sqrt{K}}\bar{H}^{-1}\mathds{1}^{\sf T}\mathcal{H}\mathcal{V}_\alpha( I - \mathcal{P}_{\alpha} + O(\mu))^{-1}\\
    -\Delta^{-1}CA^{-1}   &= -\frac{1}{\sqrt{K}}( I - \mathcal{P}_{\alpha} + O(\mu))^{-1}\mathcal{V}_\alpha^{\sf T}\mathcal{A}_2\mathcal{H}\mathds{1}\bar{H}^{-1}
\end{align}
so that
\begin{align}\label{C1}
     (I - \bar{\mathcal{C}})^{-1} = \left[\begin{array}{cc}
        \frac{1}{\mu}\bar{H}^{-1} + \frac{1}{K}\bar{H}^{-1}\mathds{1}^{\sf T}\mathcal{H}\mathcal{V}_\alpha( I - \mathcal{P}_{\alpha} + O(\mu))^{-1}\mathcal{V}_\alpha^{\sf T}\mathcal{A}_2\mathcal{H}\mathds{1}\bar{H}^{-1} & -\frac{1}{\sqrt{K}}\bar{H}^{-1}\mathds{1}^{\sf T}\mathcal{H}\mathcal{V}_\alpha( I - \mathcal{P}_{\alpha} + O(\mu))^{-1}\\
        -\frac{1}{\sqrt{K}}( I - \mathcal{P}_{\alpha} + O(\mu))^{-1}\mathcal{V}_\alpha^{\sf T}\mathcal{A}_2\mathcal{H}\mathds{1}\bar{H}^{-1} &( I - \mathcal{P}_{\alpha} + O(\mu))^{-1}
    \end{array}\right]
\end{align}
As for $\bar{\mathcal{C}}^{n+1}$, we have
\begin{align}\label{cnp1}
    \bar{\mathcal{C}}^{n+1} = \left[\begin{array}{cc}
        I - \mu\bar{H} & -\mu\frac{1}{\sqrt{K}}\mathds{1}^{\sf T}\mathcal{H}\mathcal{V}_\alpha\\
       -\mu\mathcal{V}_\alpha^{\sf T}\mathcal{A}_2\mathcal{H}\frac{1}{\sqrt{K}}\mathds{1} & \mathcal{P}_{\alpha} - \mu\mathcal{V}_\alpha^{\sf T}\mathcal{A}_2\mathcal{H}\mathcal{V}_\alpha
    \end{array}\right]^{n+1} = \left[\begin{array}{cc}
        (I - \mu \bar{H})^{n+1} + O(\mu^2) &  O(\mu)  \\
          O(\mu)& \mathcal{P}_{\alpha}^{n+1} + O(\mu^2)
    \end{array}\right]
\end{align}
We therefore obtain
\begin{align}\label{Cn}
    I - \bar{\mathcal{C}}^{n+1}  = \left[\begin{array}{cc}
        I - (I- \mu\bar{H})^{n+1} + O(\mu^2) &  O(\mu)  \\
         O(\mu) & I - \mathcal{P}_{\alpha}^{n+1} + O(\mu^2)
    \end{array}\right]
\end{align}

Note that for any $n$ and sufficiently small $\mu$, we have
\begin{align}
   (I- \mu\bar{H})^{n+1}\le I, \quad\; I - (I- \mu\bar{H})^{n+1} \le I, \quad\; \mathcal{P}_{\alpha}^{n+1} \le I,\quad\; I - \mathcal{P}_{\alpha}^{n+1} \le I
\end{align}
which means the right-hand matrices minus the left-hand side ones gives non-negative definite matrices. Thus we can write
\begin{align}\label{no1}
 (I- \mu\bar{H})^{n+1} = O(1),\quad\; I - (I- \mu\bar{H})^{n+1} = O(1), \quad\;  \mathcal{P}_{\alpha}^{n+1} = O(1),\quad\; I - \mathcal{P}_{\alpha}^{n+1} = O(1)
\end{align}
with which and (\ref{cnp1}), we obtain 
\begin{align}\label{l_cn}
    \mathcal{C}^{n+1}  = \mathcal{V}\bar{\mathcal{C}}^{n+1}\mathcal{V}^{\sf T} = O(1)
\end{align}
where $O(1)$ means constant-bounded terms whose norms do not affected by small $\mu$.

Note also that for any two invertible matrices $X$ and $Y$, where $X= O(1)$ which is a constant and $Y = o(1)$ which is related to a small variable, i.e., $X$ dominates $Y$, we have
\begin{align}\label{in_s}
    (X + Y)^{-1} = X^{-1} - X^{-1}Y(I + X^{-1}Y)^{-1}X^{-1} =  X^{-1} + o(1)
\end{align}
which means that the inverse of the sum of two matrices can be well expressed by the inverse of the dominate matrix. Also, for any two vectors $x$ and $y$, assume $x = O(1)$ and $y = o(1)$, i.e., $x$ dominates $y$, we have
\begin{align}\label{s_sum}
    \Vert x + y\Vert^2 = \Vert x \Vert^2 + \Vert y \Vert^2 + 2x^{\sf T}y = \Vert x \Vert^2 \pm o(1)
\end{align}
which means that the square of the sum of any two vectors can be well expressed by the square of the dominate vector. 

We recall the term $\mathds{E}\left\Vert\mathcal{C}^{n+1}{\widetilde{\boldsymbol{\scriptstyle\mathcal{W}}}}_{-1}' + \mu\sum\limits_{i=0}^{n} \mathcal{C}^{i}\mathcal{A}_2 d \right\Vert^2_{I\otimes\bar{H}}$, and obtain:
\begin{align}\label{mucad}
    \mathds{E}\left\Vert\mathcal{C}^{n+1}{\widetilde{\boldsymbol{\scriptstyle\mathcal{W}}}}_{-1}' + \mu\sum\limits_{i=0}^{n} \mathcal{C}^{i}\mathcal{A}_2 d \right\Vert^2_{I\otimes\bar{H}} {=} \mu^2\left\Vert\sum\limits_{i=0}^{n} \mathcal{C}^{i}\mathcal{A}_2 d \right\Vert^2_{I\otimes\bar{H}} + \mu\mathds{E} \left(\sum\limits_{i=0}^{n}\mathcal{C}^{i}\mathcal{A}_2 d\right)^{\sf T}(I\otimes\bar{H})(\mathcal{C}^{n+1}{\widetilde{\boldsymbol{\scriptstyle\mathcal{W}}}}_{-1}') + \mathds{E}\Vert\mathcal{C}^{n+1}{\widetilde{\boldsymbol{\scriptstyle\mathcal{W}}}}_{-1}'\Vert^2_{I\otimes\bar{H}}
\end{align}

We now examine $\mu^2\left\Vert\sum\limits_{i=0}^{n} \mathcal{C}^{i}\mathcal{A}_2 d \right\Vert^2_{I\otimes\bar{H}}$. To do so, we substitute  (\ref{sumci}), (\ref{cm11}), (\ref{cn11}), (\ref{C1}),  (\ref{Cn}), (\ref{in_s}) and (\ref{s_sum}) into $\Vert\sum\limits_{i=0}^{n} \mathcal{C}^{i}\mathcal{A}_2 d \Vert^2_{I\otimes\bar{H}}$, and obtain
\begin{align}\label{wnphd1}
    \left\Vert\sum\limits_{i=0}^{n} \mathcal{C}^{i}\mathcal{A}_2 d\right\Vert^2_{I\otimes\bar{H}} &= d^{\sf T}\mathcal{A}_2(I - \mathcal{C})^{-\sf T}(I - \mathcal{C}^{n+1})^{\sf T}({I\otimes\bar{H}})(I - \mathcal{C}^{n+1})(I - \mathcal{C})^{-1}\mathcal{A}_2d\notag\\
    &\overset{(a)}{=} d^{\sf T}\mathcal{A}_2\mathcal{V}(I - \bar{\mathcal{C}})^{-\sf T}(I - \bar{\mathcal{C}}^{n+1})^{\sf T}\mathcal{V}^{\sf T}({I\otimes\bar{H}})\mathcal{V}(I - \bar{\mathcal{C}}^{n+1})(I - \bar{\mathcal{C}})^{-1}\mathcal{V}^{\sf T}\mathcal{A}_2 d\notag\\
& \overset{(b)}{=} \Vert d^{\sf T}\mathcal{A}_2\mathcal{V}(I - \bar{\mathcal{C}})^{-\sf T}(I - \bar{\mathcal{C}}^{n+1})^{\sf T}\Vert^2_{I\otimes\bar{H}} \notag\\
& \overset{(c)}{=}  \Vert d^{\sf T}\mathcal{A}_2 \mathcal{V}_\alpha( I - \mathcal{P}_{\alpha})^{-1}(I - \mathcal{P}_{\alpha}^{n+1})\Vert^2_{I\otimes\bar{H}} \pm o(1)
\end{align}
where $(a)$ follows from (\ref{cm11}) and (\ref{cn11}), $(b)$ follows from the following equality:
\begin{align}
    \mathcal{V}^{\sf T}({I\otimes\bar{H}})\mathcal{V} = (V^{\sf T}\otimes I)({I\otimes\bar{H}})(V\otimes I) = (V^{\sf T} V)\otimes \bar{H} = {I\otimes\bar{H}}
\end{align}
and  $(c)$ follows from the following equality:
\begin{align}
&\Vert d^{\sf T}\mathcal{A}_2\mathcal{V}(I - \bar{\mathcal{C}})^{-\sf T}(I - \bar{\mathcal{C}}^{n+1})^{\sf T}\Vert^2_{I\otimes\bar{H}} \notag\\
&= \Bigg\Vert d^{\sf T}\mathcal{A}_2 \left[\begin{array}{cc}
       \frac{1}{\sqrt{K}}\mathds{1}  & \mathcal{V}_\alpha 
    \end{array}\right] \left[\begin{array}{cc}
        \frac{1}{\mu}\bar{H}^{-1} + O(1) & -\frac{1}{\sqrt{K}}\bar{H}^{-1}\mathds{1}^{\sf T}\mathcal{H}\mathcal{V}_\alpha( I - \mathcal{P}_{\alpha} + O(\mu))^{-1}\\
        -\frac{1}{\sqrt{K}}( I - \mathcal{P}_{\alpha} + O(\mu))^{-1}\mathcal{V}_\alpha^{\sf T}\mathcal{A}_2\mathcal{H}\mathds{1}\bar{H}^{-1} &( I - \mathcal{P}_{\alpha} + O(\mu))^{-1}
    \end{array}\right]^{\sf T}\notag\\
    &\quad\;\times \left[\begin{array}{cc}
        I - (I- \mu\bar{H})^{n+1} + O(\mu^2) & O(\mu)  \\
         O(\mu) & I - \mathcal{P}_{\alpha}^{n+1} + O(\mu^2)
    \end{array}\right]^{\sf T}\Bigg\Vert^2_{I\otimes\bar{H}}\notag\\
    &\overset{(d)}{=} \Bigg\Vert \left[\begin{array}{cc}
       0 & d^{\sf T}\mathcal{A}_2 \mathcal{V}_\alpha  
    \end{array}\right] \left[\begin{array}{cc}
        \frac{1}{\mu}\bar{H}^{-1} + O(1) & -\frac{1}{\sqrt{K}}\bar{H}^{-1}\mathds{1}^{\sf T}\mathcal{H}\mathcal{V}_\alpha( I - \mathcal{P}_{\alpha} + O(\mu))^{-1}\\
        -\frac{1}{\sqrt{K}}( I - \mathcal{P}_{\alpha} + O(\mu))^{-1}\mathcal{V}_\alpha^{\sf T}\mathcal{A}_2\mathcal{H}\mathds{1}\bar{H}^{-1} &( I - \mathcal{P}_{\alpha} + O(\mu))^{-1}
    \end{array}\right]^{\sf T}\notag\\
    &\quad\;\times \left[\begin{array}{cc}
        I - (I- \mu\bar{H})^{n+1} + O(\mu^2) & O(\mu)  \\
         O(\mu) & I - \mathcal{P}_{\alpha}^{n+1} + O(\mu^2)
    \end{array}\right]^{\sf T}\Bigg\Vert^2_{I\otimes\bar{H}}\notag\\
    &= \frac{1}{K}\Vert d^{\sf T}\mathcal{A}_2 \mathcal{V}_\alpha ( I - \mathcal{P}_{\alpha} + O(\mu))^{-1}\mathcal{V}_\alpha^{\sf T}\mathcal{H}\mathds{1}\bar{H}^{-1}(I - (I- \mu\bar{H})^{n+1}) + O(\mu)\Vert^2_{I\otimes\bar{H}}\notag\\
    &\quad\;+ \Vert d^{\sf T}\mathcal{A}_2 \mathcal{V}_\alpha( I - \mathcal{P}_{\alpha} + O(\mu))^{-1}(I - \mathcal{P}_{\alpha}^{n+1}) + O(\mu)\Vert^2 _{I\otimes\bar{H}}\notag\\
    &\overset{(e)} = \frac{1}{K}\Vert d^{\sf T}\mathcal{A}_2 \mathcal{V}_\alpha ( I - \mathcal{P}_{\alpha})^{-1}\mathcal{V}_\alpha^{\sf T}\mathcal{H}\mathds{1}\bar{H}^{-1}(I - (I- \mu\bar{H})^{n+1})\Vert^2_{\bar{H}} + \Vert d^{\sf T}\mathcal{A}_2 \mathcal{V}_\alpha( I - \mathcal{P}_{\alpha})^{-1}(I - \mathcal{P}_{\alpha}^{n+1})\Vert^2_{I\otimes\bar{H}} \pm O(\mu)\notag\\
    & \overset{(f)} = \Vert d^{\sf T}\mathcal{A}_2 \mathcal{V}_\alpha( I - \mathcal{P}_{\alpha})^{-1}(I - \mathcal{P}_{\alpha}^{n+1})\Vert^2_{I\otimes\bar{H}} + O(\epsilon^2) \pm O(\mu)
\end{align}
where $(d)$ follows from (\ref{g0}) and (\ref{p_a1a2}), and $(e)$ follows from (\ref{in_s}) and (\ref{s_sum}), and in $(f)$
we apply the following inequality:
\begin{align}
    &\Vert d^{\sf T}\mathcal{A}_2 \mathcal{V}_\alpha ( I - \mathcal{P}_{\alpha})^{-1}\mathcal{V}_\alpha^{\sf T}\mathcal{H}\mathds{1}\bar{H}^{-1}(I - (I- \mu\bar{H})^{n+1})\Vert^2_{\bar{H}} \notag\\
    & \overset{(g)}{=} \Vert d^{\sf T}\mathcal{A}_2 \mathcal{V}_\alpha ( I - \mathcal{P}_{\alpha})^{-1}\mathcal{V}_\alpha^{\sf T}(\mathcal{H} - I\otimes\bar{H})\mathds{1}\bar{H}^{-1}(I - (I- \mu\bar{H})^{n+1})\Vert^2_{\bar{H}}\notag\\
    &\le O(\Vert\mathcal{H} - I\otimes\bar{H}\Vert^2)  \overset{(h)}{=} O(\epsilon^2)
\end{align}
where $(h)$ follows from Assumption \ref{sh}, and $(g)$ follows from (\ref{no1}) and the following equality:
\begin{align}
\mathcal{V}_\alpha^{\sf T}(I\otimes\bar{H})\mathds{1} = (V_R^{\sf T}\mathbbm{1})\otimes \bar{H} = 0
\end{align}

Note that according to (\ref{no1}) and (\ref{wnphd1}), we have
\begin{align}\label{ciad1}
\left\Vert\sum\limits_{i=0}^{n} \mathcal{C}^{i}\mathcal{A}_2 d\right\Vert = O(1)    
\end{align}
Then we have
\begin{align}\label{muchm1}
    &\left\vert\mu\mathds{E} \left(\sum\limits_{i=0}^{n}\mathcal{C}^{i}\mathcal{A}_2 d\right)^{\sf T}(I\otimes\bar{H})(\mathcal{C}^{n+1}{\widetilde{\boldsymbol{\scriptstyle\mathcal{W}}}}_{-1}') + \mathds{E}\Vert\mathcal{C}^{n+1}{\widetilde{\boldsymbol{\scriptstyle\mathcal{W}}}}_{-1}'\Vert_{I\otimes\bar{H}}^2\right\vert \notag\\
    &\le \mu\mathds{E}\left\Vert\sum\limits_{i=0}^{n}\mathcal{C}^{i}\mathcal{A}_2 d\right\Vert\left\Vert I\otimes\bar{H}\right\Vert\left\Vert\mathcal{C}^{n+1}\right\Vert\Vert{\widetilde{\boldsymbol{\scriptstyle\mathcal{W}}}}_{-1}'\Vert + \Vert\mathcal{C}^{n+1}\Vert^2\Vert\bar{H}\Vert\mathds{E}\Vert{\widetilde{\boldsymbol{\scriptstyle\mathcal{W}}}}_{-1}'\Vert^2\notag\\
    &\overset{(a)}{\le} O(\mu)\times o(\sqrt{\frac{\mu}{B}}) + o(\frac{\mu}{B}) \notag\\
   & \overset{(b)}{=} o(\mu^{0.5(3+\eta)}) + o(\mu^{1+\eta})
\end{align}
where $(a)$ follows from Assumption \ref{ass_ori}, (\ref{l_cn}) and (\ref{ciad1}), and $(b)$ follows from (\ref{cbmu}), then we have:
\begin{align}\label{cnwciad}
    \mathds{E}\left\Vert\mathcal{C}^{n+1}{\widetilde{\boldsymbol{\scriptstyle\mathcal{W}}}}_{-1}' + \mu\sum\limits_{i=0}^{n} \mathcal{C}^{i}\mathcal{A}_2 d \right\Vert^2_{I\otimes\bar{H}} {=} \mu^2\left\Vert\sum\limits_{i=0}^{n} \mathcal{C}^{i}\mathcal{A}_2 d \right\Vert^2_{I\otimes\bar{H}} \pm o(\mu^{0.5(3+\eta)}) \pm o(\mu^{1+\eta})
\end{align}
Substituting (\ref{wnphd1}) into (\ref{cnwciad}), we obtain
\begin{align}\label{wnphd12}
    \mathds{E}\left\Vert\mathcal{C}^{n+1}{\widetilde{\boldsymbol{\scriptstyle\mathcal{W}}}}_{-1}' + \mu\sum\limits_{i=0}^{n} \mathcal{C}^{i}\mathcal{A}_2 d \right\Vert^2_{I\otimes\bar{H}} &{=} \mu^2\Vert d^{\sf T}\mathcal{A}_2 \mathcal{V}_\alpha( I - \mathcal{P}_{\alpha})^{-1}(I - \mathcal{P}_{\alpha}^{n+1})\Vert^2_{I\otimes\bar{H}} \pm o(\mu^{0.5(3+\eta)}) \pm o(\mu^{1+\eta}) \pm o(\mu^2)\notag\\
    & =  \mu^2\Vert d^{\sf T}\mathcal{A}_2 \mathcal{V}_\alpha( I - \mathcal{P}_{\alpha})^{-1}(I - \mathcal{P}_{\alpha}^{n+1})\Vert^2_{I\otimes\bar{H}} \pm o(\mu^{1+\eta})\pm o(\mu^2) 
\end{align}
For the centralized method where $\mathcal{V}_\alpha = 0$, we have
\begin{align}\label{mucad_centra}
    \mathds{E}\left\Vert\mathcal{C}^{n+1}{\widetilde{\boldsymbol{\scriptstyle\mathcal{W}}}}_{-1}' + \mu\sum\limits_{i=0}^{n} \mathcal{C}^{i}\mathcal{A}_2 d \right\Vert^2_{I\otimes\bar{H}}  = \mathds{E}\Vert\mathcal{C}^{n+1}{\widetilde{\boldsymbol{\scriptstyle\mathcal{W}}}}_{-1}'\Vert^2_{I\otimes\bar{H}} \le o(\frac{\mu}{B}) = o(\mu^{1+\eta})
\end{align}

We next examine the term related to gradient noise in (\ref{wnph1}). As for the term $\mathds{E}\left\Vert\sum\limits_{i=0}^{n} \mathcal{C}^{i}\mathcal{A}_2\boldsymbol{s}_i^B \right\Vert^2_{I\otimes\bar{H}}$, we consider
\begin{align}\label{rsbnc}
    R_{s,n}^B \overset{\Delta}{=} \mathrm{diag}\{R_{s,1,n}^B(\boldsymbol{w}_{1,n}), R_{s,2,n}^B(\boldsymbol{w}_{2,n}),\ldots, R_{s,K,n}^B(\boldsymbol{w}_{K,n})\}
\end{align} 
from which we obtain
\begin{align}\label{sums}
    \mathds{E}\left\Vert\sum\limits_{i=0}^{n} \mathcal{C}^{i}\mathcal{A}_2\boldsymbol{s}_i^B \right\Vert^2_{I\otimes{\bar{H}}} &= \mathds{E}(\sum\limits_i \mathcal{C}^{i}\mathcal{A}_2\boldsymbol{s}_i^B)^{\sf T}(I\otimes\bar{H})(\sum\limits_j \mathcal{C}^{j}\mathcal{A}_2\boldsymbol{s}_j^B) \notag\\
    &\overset{(a)}{=}\mathds{E}\sum\limits_i (\boldsymbol{s}_i^B)^{\sf T}\mathcal{A}_2(\mathcal{C}^i)^{\sf T}(I\otimes\bar{H})\mathcal{C}^i\mathcal{A}_2\boldsymbol{s}_i^B \notag\\
    & = \mathds{E}\sum\limits_i \mathrm{Tr}((\mathcal{C}^i)^{\sf T}(I\otimes\bar{H})\mathcal{C}^i\mathcal{A}_2\boldsymbol{s}_i^B(\boldsymbol{s}_i^B)^{\sf T}\mathcal{A}_2) \notag\\
    & = \sum\limits_i \mathrm{Tr}((\mathcal{C}^i)^{\sf T}(I\otimes\bar{H})\mathcal{C}^i\mathcal{A}_2R_{s,i}^{B}\mathcal{A}_2)
\end{align}
where $(a)$ follows from the sampling independence over time. Note that the dependence of the gradient covariance matrix $R_{s,i}^{B}$ on ${\widetilde{\boldsymbol{\scriptstyle\mathcal{W}}}}_{i}$ makes the analysis intractable, so we resort to the gradient covariance matrices at $w^{\star}$:
\begin{align}\label{d_rsb}
    R_s^B \overset{\Delta}{=} \mathrm{diag}\{R_{s,1}^B, R_{s,2}^B,\ldots, R_{s,k}^B\}
\end{align}
where
\begin{align}
    R_{s,k}^B \overset{\Delta}{=} \mathds{E}\{\boldsymbol{s}_{k,n}^B(w^\star)(\boldsymbol{s}_{k,n}^B(w^\star))^{\sf T}\}
\end{align}
Similarly, we can define the gradient covariance matrix at $w^\star$ for the case of $B = 1$:
\begin{align}\label{d_rs}
    R_s \overset{\Delta}{=} \mathrm{diag}\{R_{s,1}, R_{s,2},\ldots, R_{s,k}\}
\end{align}
with
\begin{align}
    R_{s,k} \overset{\Delta}{=} \mathds{E}\{\boldsymbol{s}_{k,n}(w^\star)\boldsymbol{s}_{k,n}^{\sf T}(w^\star)\}
\end{align}
According to (\ref{Rskb1}), we have
\begin{align}\label{rsbar}
    R_{s,k}^B {=} \frac{1}{B}R_{s,k}
\end{align}
and recalling (\ref{d_barrs}), we have
 \begin{align}\label{d_barrs_1}
     \bar{R}_s = \mathds{E}\left\{\left(\frac{1}{K}\sum\limits_k \boldsymbol{s}_{k,n}(w^\star)\right)\left(\frac{1}{K}\sum\limits_\ell \boldsymbol{s}_{\ell,n}(w^\star)\right)^{\sf T}\right\} \overset{(a)}{=} \frac{1}{K^2}\sum_{k=1}^{K}  \mathds{E}\{\boldsymbol{s}_{k,n}(w^\star)\boldsymbol{s}_{k,n}^{\sf T}(w^\star)\} = \frac{1}{K^2}\sum_{k=1}^{K} R_{s,k}
 \end{align}
where $(a)$ follows from the independence among agents. 

Substituting (\ref{d_rsb}), (\ref{d_rs}) and (\ref{rsbar}) into (\ref{sums}), we then have
\begin{align}\label{ssshf1}
\mathds{E}\left\Vert\sum\limits_{i=0}^{n} \mathcal{C}^{i}\mathcal{A}_2\boldsymbol{s}_i^B \right\Vert^2_{I\otimes\bar{H}}& = \frac{1}{B}\sum\limits_i \mathrm{Tr}((\mathcal{C}^i)^{\sf T}(I\otimes\bar{H})\mathcal{C}^i\mathcal{A}_2 R_{s}\mathcal{A}_2)+ \frac{1}{B}\sum\limits_i \mathrm{Tr}((\mathcal{C}^i)^{\sf T}(I\otimes \bar{H})\mathcal{C}^i\mathcal{A}_2 \mathds{E}(R_{s,i} - R_s)\mathcal{A}_2)
\end{align}
Before continuing the analysis of (\ref{ssshf1}), we verify that the size of  $R_{s}$ dominates the approximation error of $R_{s,i} - R_s$. For each agent $k$, we recall the definition of $R_{s,k,i}(w)$ in (\ref{drswn}) and use Taylor expansion, and obtain  
\begin{align}
 R_{s,k,n}(\boldsymbol{w}_{k,i-1}) -  R_{s,k} = \nabla R_{s,k}\mathds{E}(\boldsymbol{w}_{k,i-1} - w^\star) \pm O(\mathds{E}\Vert\boldsymbol{w}_{k,i-1} - w^\star\Vert^2) 
\end{align}
with which we have
\begin{align}
    \Vert R_{s,k,i}(\boldsymbol{w}_{k,i-1}) -  R_{s,k}\Vert \le O(\mathds{E}\Vert\tilde{\boldsymbol{w}}_{k,i-1}\Vert) \le O((\mathds{E}\Vert\tilde{\boldsymbol{w}}_{k,i-1}\Vert^2)^{\frac{1}{2}})
\end{align}
Moreover, since the gradient noise across agents are uncorrelated with each other, we have
\begin{align}\label{ersnrs}
    \Vert R_{s,i} -  R_{s}\Vert  = \max\limits_k \Vert R_{s,k,i}(\boldsymbol{w}_{k,i-1}) -  R_{s,k}\Vert  \le  O((\mathds{E}\Vert\widetilde{\boldsymbol{\scriptstyle\mathcal{W}}}_{i-1}\Vert^2)^\frac{1}{2}) {\overset{(a)}{=} O(\mu^{\frac{\gamma'}{2}})}
\end{align}
where $(a)$ follows from (\ref{3uw}). 

Moreover, since for any square matrix $X$, it holds that 
\begin{align}
    \vert\mathrm{Tr}(X)\vert \le c\Vert X\Vert
\end{align}
with a constant $c$, we have
\begin{align}\label{trcctae}
    \vert\mathrm{Tr}(\mathcal{A}_2(R_{s,i} - R_s)\mathcal{A}_2)\vert &= \vert\mathrm{Tr}(\mathcal{A}_2^2(R_{s,i} - R_s))\vert\notag\\
    & \le c\Vert\mathcal{A}_2^2\Vert\Vert R_{s,i} - R_s\Vert \notag\\
    &= O(\mu^{\frac{\gamma'}{2}})
\end{align}
Substituting (\ref{trcctae}) into (\ref{ssshf1}), we obtain
\begin{align}\label{sssf1}
    \mathds{E}\left\Vert\sum\limits_{i=0}^{n} \mathcal{C}^{i}\mathcal{A}_2\boldsymbol{s}_i^B \right\Vert^2_{I\otimes\bar{H}} &=  \frac{1}{B}\sum\limits_{i=0}^{n} \mathrm{Tr}((\mathcal{C}^i)^{\sf T}(I\otimes\bar{H})\mathcal{C}^i\mathcal{A}_2 R_{s}\mathcal{A}_2) \pm O(\frac{\mu^{\frac{\gamma'}{2}}}{B})\sum\limits_{i=0}^{n} \mathrm{Tr}((\mathcal{C}^i)^{\sf T}(I\otimes\bar{H})\mathcal{C}^i) 
\end{align}
where the first term on the right hand side dominates the second term with small $\mu$.  

To proceed, we examine $\sum\limits_{i=0}^{n}\mathrm{Tr}((\mathcal{C}^i)^{\sf T}(I\otimes\bar{H})\mathcal{C}^i\mathcal{A}_2 R_{s}\mathcal{A}_2)$. Here we apply the \emph{block Kronecker product} denoted by $\otimes_b$ and the \emph{block vectorization operation} denoted by $\mathrm{bvec}$ with blocks of size $R^{M\times M}$. Note that the block Kronecker product of any two matrices with  size of $R^{M\times M}$ is equivalent to their Kronecker product, and similarly the block vectorization of any matrix with size of $R^{M\times M}$ is the same as its vectorization. Then we have
\begin{align}\label{sr1}
   \sum\limits_{i=0}^{n} \mathrm{Tr}((\mathcal{C}^i)^{\sf T}(I\otimes\bar{H})\mathcal{C}^i\mathcal{A}_2 R_{s}\mathcal{A}_2) &\overset{(a)}{=} \sum\limits_{i=0}^{n} (\mathrm{bvec}(\mathcal{A}_2 R_{s}\mathcal{A}_2))^{\sf T} \mathrm{bvec}((\mathcal{C}^i)^{\sf T}(I\otimes\bar{H})\mathcal{C}^i)
   )\notag\\
    &\overset{(b)}{=}  (\mathrm{bvec}(\mathcal{A}_2 R_{s}\mathcal{A}_2))^{\sf T}  \sum\limits_{i=0}^{n} (\mathcal{C}^{\sf T}\otimes_b \mathcal{C}^{\sf T})^i  \mathrm{bvec}(I\otimes \bar{H})
\end{align}
where $(a)$ and $(b)$ follows from the following two properties:
\begin{align}
    \label{ptrb}
    \mathrm{bvec}(XYZ) = (Z^{\sf T}\otimes_b X)\mathrm{bvec}(Y)\\
    \label{pbv}
    \mathrm{Tr}(XY) = (\mathrm{bvec}(Y^{\sf T}))^{\sf T}\mathrm{bvec}(X)
\end{align}
where $X$,$Y$ and $Z$ are three arbitrary matrices.

Let $\mathcal{F} = \mathcal{C}^{\sf T}\otimes_b \mathcal{C}^{\sf T}$ and $\bar{\mathcal{F}} =(\bar{\mathcal{C}}\otimes_b \bar{\mathcal{C}})^{\sf T}$ where $\bar{\mathcal{C}}$ is defined in (\ref{Bbar}), it holds that
\begin{align}\label{ctbct}
    \sum\limits_{i=0}^{n} (\mathcal{C}^{\sf T}\otimes_b \mathcal{C}^{\sf T})^i = (I - \mathcal{F}^{n+1})(I - \mathcal{F})^{-1} = (\mathcal{V}\otimes_b\mathcal{V})(I - \bar{\mathcal{F}}^{n+1})(I - \bar{\mathcal{F}})^{-1}(\mathcal{V}^{\sf T}\otimes_b\mathcal{V}^{\sf T})
\end{align}
By recalling (\ref{mcbar}), we have
\begin{align}
    \bar{\mathcal{F}} &=(\bar{\mathcal{C}}\otimes_b \bar{\mathcal{C}})^{\sf T} = \left[\begin{array}{cc}
        I - \mu\bar{H} & -\mu\frac{1}{\sqrt{K}}\mathds{1}^{\sf T}\mathcal{H}\mathcal{A}_2\mathcal{V}_\alpha\\
       -\mu\mathcal{V}_\alpha^{\sf T}\mathcal{H}\frac{1}{\sqrt{K}}\mathds{1} & \mathcal{P}_{\alpha} - \mu\mathcal{V}_\alpha^{\sf T}\mathcal{H}\mathcal{A}_2\mathcal{V}_\alpha
    \end{array}\right]\otimes_b\left[\begin{array}{cc}
        I - \mu\bar{H} & -\mu\frac{1}{\sqrt{K}}\mathds{1}^{\sf T}\mathcal{H}\mathcal{A}_2\mathcal{V}_\alpha\\
       -\mu\mathcal{V}_\alpha^{\sf T}\mathcal{H}\frac{1}{\sqrt{K}}\mathds{1} & \mathcal{P}_{\alpha} - \mu\mathcal{V}_\alpha^{\sf T}\mathcal{H}\mathcal{A}_2\mathcal{V}_\alpha
    \end{array}\right]\notag\\
    & = \left[\begin{array}{cc}
        (I - \mu\bar{H})\otimes_b (I - \mu\bar{H})& O(\mu) \\
         O(\mu)& O(1)
    \end{array}\right]
\end{align}
Note that since $ I - \mu\bar{H} \in R^{M\times M}$, we have
\begin{align}
   (I - \mu\bar{H})\otimes_b (I - \mu\bar{H}) = (I - \mu\bar{H})\otimes (I - \mu\bar{H}) 
\end{align}
from which we have
\begin{align} 
    (I - \bar{\mathcal{F}})^{-1} = \left[\begin{array}{cc}
        \mu \bar{H}\otimes I + \mu I \otimes \bar{H} + O(\mu^2)& O(\mu) \\
         O(\mu)& O(1)
    \end{array}\right]^{-1}
\end{align}
Recall the block matrix inverse operation in (\ref{bmi}), we obtain
\begin{align}\label{fm1}
    (I - \bar{\mathcal{F}})^{-1} =  \left[\begin{array}{cc}
        \frac{1}{\mu} (\bar{H}\otimes I +  I \otimes \bar{H})^{-1}& 0 \\ 
        0& 0 
    \end{array}\right] + O(1)
\end{align}
As for $\bar{\mathcal{F}}^{n+1}$, we have
\begin{align}
\bar{\mathcal{F}}^{n+1} = \left[\begin{array}{cc}
        ( I - \mu\bar{H})^{n+1} \otimes ( I - \mu\bar{H})^{n+1}+ O(\mu^2)& O(\mu) \\
         O(\mu)& O(1)
    \end{array}\right]   
\end{align}
from which we have
\begin{align}\label{mfn}
    I - \bar{\mathcal{F}}^{n+1} =  \left[\begin{array}{cc}
        I- ( I - \mu\bar{H})^{n+1} \otimes ( I - \mu\bar{H})^{n+1} + O(\mu^2)& O(\mu) \\
         O(\mu)& O(1)
    \end{array}\right]
\end{align}
Combining (\ref{mfn}) and (\ref{fm1}), we obtain
\begin{align}\label{cfnf1}
    (I - \bar{\mathcal{F}}^{n+1})(I - \bar{\mathcal{F}})^{-1} = \left[\begin{array}{cc}
        \frac{1}{\mu}\left(I- ( I - \mu\bar{H})^{n+1} \otimes ( I - \mu\bar{H})^{n+1}\right) (\bar{H}\otimes I +  I \otimes \bar{H})^{-1}& 0 \\ 
        0& 0 
    \end{array}\right] + O(1)
\end{align}
Substituting (\ref{cfnf1}) and  (\ref{ctbct}) into (\ref{sr1}), we obtain
\begin{align}\label{wnpsh}
   &\sum\limits_i \mathrm{Tr}((\mathcal{C}^i)^{\sf T}(I\otimes\bar{H})\mathcal{C}^i\mathcal{A}_2 R_{s}\mathcal{A}_2)\notag\\
   &{=} \left(\mathrm{bvec}(\mathcal{A}_2 R_{s}\mathcal{A}_2)\right)^{\sf T}\sum\limits_i (\mathcal{C}^i \otimes_b \mathcal{C}^i)^{\sf T}\mathrm{bvec}(I \otimes \bar{H})\notag\\
   &= (\mathrm{bvec}(\mathcal{A}_2 R_{s}\mathcal{A}_2))^{\sf T} (\mathcal{V}\otimes_b\mathcal{V})\left(\left[\begin{array}{cc}
        \frac{1}{\mu}\left(I- ( I - \mu\bar{H})^{n+1} \otimes ( I - \mu\bar{H})^{n+1}\right) (\bar{H}\otimes I + I \otimes \bar{H})^{-1}& 0 \\ 
        0& 0 
    \end{array}\right] + O(1)\right)\notag\\
    &\quad \times(\mathcal{V}^{\sf T}\otimes_b\mathcal{V}^{\sf T})\mathrm{bvec}(I\otimes \bar{H})\notag\\
    & \overset{(a)}{=} \frac{1}{\mu K^2}(\mathrm{bvec}(\mathcal{A}_2 R_{s}\mathcal{A}_2))^{\sf T} (\mathds{1}\otimes_b\mathds{1})\left(I- ( I - \mu\bar{H})^{n+1} \otimes ( I - \mu\bar{H})^{n+1}\right)(\bar{H}\otimes I + I \otimes \bar{H})^{-1}(\mathds{1}^{\sf T}\otimes_b \mathds{1}^{\sf T})\notag\\
    & \quad \times\mathrm{bvec}(I\otimes \bar{H}) \pm O(1)
\end{align}
where $(a)$ follows from (\ref{vpv1}). We focus on the first term in (\ref{wnpsh}) as it is on the order of $O(\frac{1}{\mu})$, which dominates the second term $O(1)$. According to (\ref{ptrb}), it holds that
\begin{align}\label{kvecih}
   (\mathds{1}^{\sf T}\otimes_b \mathds{1}^{\sf T})\mathrm{bvec}(I\otimes\bar{H})  = \mathrm{bvec}(\mathbbm{1}^{\sf T}\mathbbm{1}\otimes \bar{H}) = K\mathrm{bvec}(\bar{H}) =K\mathrm{vec}(\bar{H})
\end{align}
which enables us to solve $(\bar{H}\otimes I +  I \otimes \bar{H})^{-1}(\mathds{1}^{\sf T}\otimes_b \mathds{1}^{\sf T})\mathrm{bvec}(I\otimes\bar{H})$. Consider the following equation:
\begin{align}
x = (\bar{H}\otimes I +  I \otimes \bar{H})^{-1}(\mathds{1}^{\sf T}\otimes_b \mathds{1}^{\sf T})\mathrm{bvec}(I\otimes\bar{H}) = K(\bar{H}\otimes I +  I \otimes \bar{H})^{-1}\mathrm{vec}(\bar{H})
\end{align}
where $x$ is a vector of size $R^{M\times 1}$, which is also the solution to the following linear system:
\begin{align}\label{lsh}
    (\bar{H}\otimes I +  I \otimes \bar{H})x = K\mathrm{vec}(\bar{H})
\end{align}
Let $X = \mathrm{unvec}(x)$ whose vector representation is $x$, and unvectorize the right-hand side of (\ref{lsh}) with (\ref{ptrb}), we obtain the following continuous-time Lyapunov equation:
\begin{align}
    \bar{H}X + X\bar{H} = K \bar{H}
\end{align}
for which we can verify the solution is given by
\begin{align}\label{so_xh}
    X = \frac{K}{2}I_M
\end{align}
Substituting (\ref{so_xh}) into (\ref{wnpsh}), we have
\begin{align}\label{cashf22}
    &\sum\limits_i \mathrm{Tr}((\mathcal{C}^i)^{\sf T}(I\otimes\bar{H})\mathcal{C}^i\mathcal{A}_2 R_{s}\mathcal{A}_2)\notag\\
    & = \frac{1}{\mu K^2}(\mathrm{bvec}(\mathcal{A}_2 R_{s}\mathcal{A}_2))^{\sf T} (\mathds{1}\otimes_b\mathds{1})\left(I- ( I - \mu\bar{H})^{n+1} \otimes ( I - \mu\bar{H})^{n+1}\right)(\bar{H}\otimes I + I \otimes \bar{H})^{-1}(\mathds{1}^{\sf T}\otimes_b \mathds{1}^{\sf T})\mathrm{bvec}(I\otimes \bar{H}) \pm O(1)\notag\\
    & \overset{(a)}{=} \frac{1}{2\mu K}(\mathrm{bvec}(\mathcal{A}_2 R_{s}\mathcal{A}_2))^{\sf T} \mathrm{bvec}\left(\mathds{1}\mathds{1}^{\sf T} - \mathds{1}( I - \mu\bar{H})^{n+1}( I - \mu\bar{H})^{n+1}\mathds{1}^{\sf T}\right) \pm O(1)\notag\\
    & \overset{(b)}{=} \frac{1}{2\mu K}\mathrm{Tr}\left((I - (I - \mu\bar{H})^{2(n+1)})\mathds{1}^{\sf T}\mathcal{A}_2R_{s}\mathcal{A}_2\mathds{1}\right)\pm O(1)\notag\\
& \overset{(c)}{=} \frac{1}{2\mu K}\mathrm{Tr}\left((I - (I - \mu \bar{H})^{2(n+1)})\mathds{1}^{\sf T}R_{s}\mathds{1}\right)\pm O(1)\notag\\
& \overset{(d)}{=} \frac{K}{2\mu }\mathrm{Tr}\left((I - (I - \mu \bar{H})^{2(n+1)})\bar{R}_s \right)\pm O(1)
\end{align}
where $(a)$ follows from (\ref{ptrb}), $(b)$ follows from (\ref{pbv}), $(c)$ follows from (\ref{p_a1a2}), and $(d)$ follows from (\ref{d_barrs_1}). 

As for the size of $\sum\limits_i \mathrm{Tr}((\mathcal{C}^i)^{\sf T}(I\otimes\bar{H})\mathcal{C}^i)$, by using a similar technique to $\sum\limits_i \mathrm{Tr}((\mathcal{C}^i)^{\sf T}(I\otimes\bar{H})\mathcal{C}^i\mathcal{A}_2 R_{s}\mathcal{A}_2)$, we have
\begin{align}\label{ccarah}
 & \sum\limits_k \mathrm{Tr}((\mathcal{C}^i)^{\sf T}(I\otimes\bar{H})\mathcal{C}^i)\overset{(a)}{=}(\mathrm{bvec}(I))^{\sf T}  \sum\limits_i (\mathcal{C}^{\sf T}\otimes_b \mathcal{C}^{\sf T})^i  \mathrm{bvec}(I\otimes\bar{H})   \notag\\
  & \overset{(b)}{=} \frac{1}{\mu K^2}(\mathrm{bvec}(I))^{\sf T}(\mathds{1}\otimes_b\mathds{1})\left(I- ( I - \mu\bar{H})^{n+1} \otimes ( I - \mu\bar{H})^{n+1}\right)(\bar{H}\otimes I + I \otimes \bar{H})^{-1}(\mathds{1}^{\sf T}\otimes_b \mathds{1}^{\sf T})\mathrm{bvec}(I\otimes \bar{H}) \pm O(1) \notag\\
    &= O(\frac{1}{\mu})
\end{align}
where $(a)$ follows from (\ref{ptrb}) and (\ref{pbv}), $(b)$ follows from (\ref{wnpsh}).

Substituting (\ref{cashf22}) and (\ref{ccarah}) into (\ref{sssf1}), we obtain
\begin{align}\label{cashf3}
  \mathds{E}\left\Vert\sum\limits_{i=0}^{n} \mathcal{C}^{i}\mathcal{A}_2\boldsymbol{s}_i^B \right\Vert^2_{I\otimes\bar{H}}& =  \frac{K}{2\mu B}\mathrm{Tr}\left((I - (I - \mu \bar{H})^{2(n+1)})\bar{R}_s \right)\pm O(\frac{\mu^{\frac{\gamma'}{2}}}{\mu B})   
\end{align}
where in the right-hand side the first term $O(\frac{1}{\mu B})$ dominates the second term $O(\frac{\mu^{\frac{\gamma'}{2}}}{\mu B})$.

We further substitute  (\ref{wnphd12}) and (\ref{cashf3}) into (\ref{wnph1}), and obtain
\begin{align}\label{er_g}
\mathds{E}\Vert{\widetilde{\boldsymbol{\scriptstyle\mathcal{W}}}}_{n}'\Vert^2_{I\otimes\bar{H}} = & \frac{\mu K}{2B}\mathrm{Tr}\left(\left(I - (I - \mu \bar{H})^{2(n+1)}\right)\bar{R}_s \right)+ \mu^2\Vert d^{\sf T}\mathcal{A}_2 \mathcal{V}_\alpha( I - \mathcal{P}_{\alpha})^{-1}(I - \mathcal{P}_{\alpha}^{n+1})\Vert^2_{I\otimes\bar{H}} \notag\\
& \pm O(\frac{\mu^{\frac{\gamma'}{2}+1}}{B})   \pm o(\mu^{1+\eta}) + o(\mu^2)
\end{align}
while for the centralize method, we substitute (\ref{mucad_centra}) and (\ref{cashf3}) into (\ref{wnph1}), and obtain
\begin{align}\label{er_g_centra}
\mathds{E}\Vert{\widetilde{\boldsymbol{\scriptstyle\mathcal{W}}}}_{n}'\Vert^2_{I\otimes\bar{H}} = & \frac{\mu K}{2B}\mathrm{Tr}\left(\left(I - (I - \mu \bar{H})^{2(n+1)}\right)\bar{R}_s \right) \pm O(\frac{\mu^{\frac{\gamma'}{2}+1}}{B}) + o(\mu^{1+\eta})
\end{align}
from which the escaping efficiency of the three distributed methods can be derived.

Basically, for the \emph{centralized} method, we substitute  (\ref{er_g_centra}) into (\ref{erave_p}), and obtain
\begin{align}\label{er_centra}
   \mathrm{ER}_{n,cen} &= \frac{\mu }{4B}\mathrm{Tr}\left(\left(I - (I - \mu \bar{H})^{2(n+1)}\right)\bar{R}_s\right)\pm O(\mu^{1.5\gamma'}) \pm O(\frac{\mu^{\frac{\gamma'}{2}+1}}{B}) +   o(\mu^{1+\eta})  \notag\\
   & \overset{(a)}{=} \frac{\mu }{4B}\mathrm{Tr}\left(\left(I - (I - \mu \bar{H})^{2(n+1)}\right)\bar{R}_s \right)\pm O\left(\mu^{1.5(1+\eta)}\right)+  o(\mu^{1+\eta}) \notag\\
   & = \frac{\mu }{4B}\mathrm{Tr}\left(\left(I - (I - \mu \bar{H})^{2(n+1)}\right)\bar{R}_s \right) +  o(\mu^{1+\eta})
\end{align}
where $(a)$ follows from $\gamma' = 1+\eta$ and the following equality:
\begin{align}
   O(\frac{\mu^{\frac{\gamma'}{2}+1}}{B}) =  O(\mu^{\frac{1+\eta}{2}+1+\eta}) =  O\left(\mu^{1.5(1+\eta)}\right)
\end{align}
For \emph{consensus}, we apply $A_2 = I$, and substitute  (\ref{er_g}) into (\ref{erave_p}), which gives
\begin{align}\label{er_consen}
   \mathrm{ER}_{n,con} = & \frac{\mu }{4B}\mathrm{Tr}\left(\left(I - (I - \mu \bar{H})^{2(n+1)}\right)\bar{R}_s \right)+ \frac{\mu^2}{2K}\Vert d^{\sf T}\mathcal{V}_\alpha( I - \mathcal{P}_{\alpha})^{-1}(I - \mathcal{P}_{\alpha}^{n+1})\Vert^2_{I\otimes\bar{H}} \pm o(\mu^2)   \pm o(\mu^{1+\eta})
\end{align}
For the \emph{diffusion} strategy, we apply $A_2 = A$. In this case, we have
\begin{align}
    \mathcal{A}_2\mathcal{V}_\alpha = \mathcal{V}_{\alpha}\mathcal{P}_{\alpha}
\end{align}
with which we obtain
\begin{align}\label{er_diff}
   \mathrm{ER}_{n,dif} = & \frac{\mu}{4B}\mathrm{Tr}\left(\left(I - (I - \mu \bar{H})^{2(n+1)}\right)\bar{R}_s \right)+ \frac{\mu^2}{2K}\Vert d^{\sf T}\mathcal{V}_\alpha \mathcal{P}_\alpha(I - \mathcal{P}_{\alpha})^{-1}(I - \mathcal{P}_{\alpha}^{n+1})\Vert^2_{I\otimes\bar{H}} \pm o(\mu^2) \pm o(\mu^{1+\eta})
\end{align}
{Recalling (\ref{er_w_c_w_n}), the common $O(\frac{\mu}{B})$ term in (\ref{er_centra}), (\ref{er_consen}), and (\ref{er_diff}), associated with  gradient noise, corresponds to the size of the term in (\ref{er_w_c_w_n}) that is related to the network centroid. In addition, the additional $O(\mu^2)$ term in the escaping efficiency of decentralized methods arises due to the consensus distance.}

Finally, we specify the number of iterations required for the algorithms to escape the local basin. For the centralized method, recall (\ref{er_centra}), and let $\sigma_i$ with $i=1,\ldots,M$ denote the i-th largest eigenvalue of $\bar{H}$, we have
\begin{align}\label{iter_1}
    \mathrm{ER}_{n, cen}  &= \frac{\mu }{4B}\mathrm{Tr}\left(\left(I - (I - \mu \bar{H})^{2(n+1)}\right)\bar{R}_s \right) \notag\\
    &\overset{(a)}{=} \frac{\mu }{4B}\mathrm{Tr}\left(\left(I - e^{- 2\mu(n+1) \bar{H}}\right)\bar{R}_s \right) \pm  o\left(\frac{\mu}{B}\right)
\end{align}
where $(a)$ follows from the matrix exponential, i.e.,
\begin{align}
    e^{-\mu\bar{H}} = \sum\limits_{j=0}^{\infty} \frac{1}{j!}\big(-\mu\bar{H}\big)^j = I - \mu\bar{H} + o(\mu)
\end{align}
and
\begin{align}
    (e^{-\mu\bar{H}} + O(\mu^2))^{2(n+1)} = e^{-2\mu(n+1)\bar{H}} + o(1)
\end{align}
Note that
\begin{align}
    \mathrm{Tr}\left(I - e^{- 2\mu(n+1) \bar{H}}\right) \lambda_{\min}(\bar{R}_s) \le \mathrm{Tr}\left(\left(I - e^{- 2\mu(n+1) \bar{H}}\right)\bar{R}_s\right) \le \mathrm{Tr}\left(I - e^{- 2\mu(n+1) \bar{H}}\right)\lambda_{\max}(\bar{R}_s)
\end{align}
where $\lambda_{\min}(\bar{R}_s)$ and $\lambda_{\max}(\bar{R}_s)$ denote the smallest and largest eigenvalues of $\bar{R}_s$, respectively. Thus, we have:
\begin{align}\label{iter_2}
\mathrm{Tr}\left(\left(I - e^{- 2\mu(n+1) \bar{H}}\right)\bar{R}_s\right) = c_1 \mathrm{Tr}\left(I - e^{- 2\mu(n+1) \bar{H}}\right) =c_1 \sum\limits_{i=1}^{M} (1 - e^{-2\mu(n+1)\sigma_i})
\end{align}
where 
\begin{align}
    \lambda_{\min}(\bar{R}_s) \le c_1 \le \lambda_{\max}(\bar{R}_s)
\end{align}
Also, for any positive-definite matrix $X$, we have 
\begin{align}\label{iter_3}
    \mathrm{Tr}(X) = c_2\lambda_{\max}(X)
\end{align}
with $1 \le c_2 \le M$, where $\lambda_{\max}(X)$ is the largest eigenvalue of $X$. Let $c\overset{\Delta}{=}c_1c_2$, and substitute \eqref{iter_2} and \eqref{iter_3} into \eqref{iter_1}, we obtain
\begin{align}
       \mathrm{ER}_n {=} \frac{\mu c}{4B}\sigma_1(1 - e^{-2\mu(n+1)\sigma_1})\pm o\left(\frac{\mu}{B}\right) 
\end{align}

Let $\mathrm{ER}_{n, cen} = h$, then we need to solve the following equation associated with $n$:
\begin{align}
    \frac{\mu c}{4B}\sigma_1(1 - e^{-2\mu(n+1)\sigma_1}) = h \pm o\left(\frac{\mu}{B}\right)
\end{align}
for which we can verify the solution is given by
\begin{align}
    n = \frac{1}{2\mu\sigma_1}\mathrm{log}\frac{1}{1 - \frac{4B(h \pm o\left(\frac{\mu}{B}\right))}{\mu c \sigma_1}} -1 = O(\frac{1}{\mu})
\end{align}
This means that on average the centralized method leaves the local basin in $O(\frac{1}{\mu})$ number of iterations given that it can successfully escape from the local minimum. This is also the reason why we consider $n\le O(\frac{1}{\mu})$ when analyzing the escaping efficiency.   As for the two decentralized methods, unfortunately it is not tractable to obtain a closed-form solution for $n$ with the $O(\mu^2)$ terms. However, since decentralized methods escape faster than the centralized approach, we know the number of iterations required for decentralized methods to leave the local basin will not exceed the centralized one. Therefore, we also consider $n\le O(\frac{1}{\mu})$ when analyzing the escaping efficiency of decentralized methods.

\section{Proof for Corollary \ref{coro2}: Comparison between consensus and diffusion}\label{pcoro2}
In this section, we compare the escaping efficiency of consensus and diffusion. Recall (\ref{er_consen}) and (\ref{er_diff}), from which we observe that the only difference between $\mathrm{ER}_{n,con}$ and $\mathrm{ER}_{n,dif}$ is that there is an extra $\mathcal{P}_\alpha$ in the $O(\mu^2)$ term of diffusion. Thus, we only need to compare the $O(\mu^2)$ terms. Recall that $P_\alpha$ is a diagonal matrix with elements from the second largest-magnitude eigenvalue $\lambda_2$ to the smallest-magnitude eigenvalue $\lambda_K$ of $A$ appearing on the diagonal:
\begin{align}\label{ullp}
P_\alpha = \left[\begin{array}{ccc}
    \lambda_2&&\\
    &\ddots&\\
    &&\lambda_K
\end{array}\right]
\end{align}
where the absolute value of each $\lambda_k$ for $k=\{2\cdots K\}$ is strictly smaller than 1. Also,
\begin{align}
  \mathcal{P}_\alpha = P_\alpha \otimes I_M  
\end{align}

 Let 
\begin{align}
    x = \mathcal{V}_\alpha^{\sf T}d = \mathrm{col}_{k}\{x_k\} \in \mathbbm{R}^{(K-1)M}
\end{align}
which is a block vector consisting of $K-1$ sub-vectors denoted by $x_k$ for $k =2,\ldots, K$, and the dimension of $x_k$ is $M$.

Then, for consensus we have
\begin{align}
\Vert d^{\sf T}\mathcal{V}_\alpha (I - \mathcal{P}_{\alpha})^{-1}(I - \mathcal{P}_{\alpha}^{n+1})\Vert^2_{I\otimes\bar{H}} = \Vert x \Vert^{2}_{((I - P_\alpha)^{-1}(I - P_\alpha^{n+1}))^2\otimes \bar{H} }
\end{align}
Note that
\begin{align}
  {((I - P_\alpha)^{-1}(I - P_\alpha^{n+1}))^2\otimes \bar{H} } =   \left[\begin{array}{ccc}
    \frac{(1 - \lambda_2^{n+1})^2}{(1 - \lambda_2)^2}\bar{H}&&\\
    &\ddots&\\
    &&\frac{(1 - \lambda_K^{n+1})^2}{(1 - \lambda_K)^2}\bar{H}
\end{array}\right]
\end{align}
from which we have 
\begin{align}
\Vert x \Vert^{2}_{((I - P_\alpha)^{-1}(I - P_\alpha^{n+1}))^2\otimes \bar{H} } &= \left[\begin{array}{ccc}
    x_2^{\sf T} &\cdots& x_K^{\sf T}
\end{array}\right] \left[\begin{array}{ccc}
    \frac{(1 - \lambda_2^{n+1})^2}{(1 - \lambda_2)^2}\bar{H}&&\\
    &\ddots&\\
    &&\frac{(1 - \lambda_K^{n+1})^2}{(1 - \lambda_K)^2}\bar{H}
\end{array}\right]    \left[\begin{array}{c}
     x_2\\
     \vdots\\
     x_K
\end{array}\right] = \sum\limits_{k=2}^{K} \frac{(1-\lambda_k^{n+1})^2}{(1-\lambda_k)^2}\Vert x_k\Vert^2_{\bar{H}}
\end{align}

Similarly, for diffusion we have
\begin{align}
    \Vert d^{\sf T}\mathcal{V}_\alpha\mathcal{P}_{\alpha} (I - \mathcal{P}_{\alpha})^{-1}(I - \mathcal{P}_{\alpha}^{n+1})\Vert^2_{I\otimes\bar{H}} &= \left[\begin{array}{ccc}
    x_2^{\sf T} &\cdots& x_K^{\sf T}
\end{array}\right] \left[\begin{array}{ccc}
    \frac{\lambda_2^2(1 - \lambda_2^{n+1})^2}{(1 - \lambda_2)^2}\bar{H}&&\\
    &\ddots&\\
    &&\frac{\lambda_K^2(1 - \lambda_K^{n+1})^2}{(1 - \lambda_K)^2}\bar{H}
\end{array}\right]    \left[\begin{array}{c}
     x_2\\
     \vdots\\
     x_K
\end{array}\right] \notag\\
& = \sum\limits_{k=2}^{K} \frac{\lambda_k^2(1-\lambda_k^{n+1})^2}{(1-\lambda_k)^2}\Vert x_k\Vert^2_{\bar{H}}
\end{align}
Since $\lambda_k$ for $k=2,\ldots,K$ are strictly inside the unit circle, i.e.,
\begin{align}
    \lambda_k^2 < 1
\end{align}
we have
\begin{align}\label{dif_consen_s}
    \sum\limits_{k=2}^{K} \frac{\lambda_k^2(1-\lambda_k^{n+1})^2}{(1-\lambda_k)^2}\Vert x_k\Vert^2_{\bar{H}} \le \sum\limits_{k=2}^{K} \frac{(1-\lambda_k^{n+1})^2}{(1-\lambda_k)^2}\Vert x_k\Vert^2_{\bar{H}}
\end{align}
from which we have
\begin{align}
    \mathrm{ER}_{n, dif} \le \mathrm{ER}_{n,con}
\end{align}
\section{Proof for Corollary \ref{th_er_op}}\label{per_l}
In this section, we examine the closed-form excess-risk performance of the centralized, consensus and diffusion methods in the long term when $n\to \infty$ which corresponds to their optimization performance. To achieve the goal, we first need to verify how accurately the approximate model defined in (\ref{uni_dis_a}) can approximate the original model in (\ref{uni_dis}). To do this, similar to Lemmas \ref{mse} and \ref{mse_centra}, we need to analyze the upper bounds of $\mathds{E}\Vert{\widetilde{\boldsymbol{\scriptstyle\mathcal{W}}}}_{n}\Vert^2$, $\mathds{E}\Vert{\widetilde{\boldsymbol{\scriptstyle\mathcal{W}}}}_{n}\Vert^2$ and $\vert \mathds{E}\Vert\widetilde{\boldsymbol{\scriptstyle\mathcal{W}}}_n'\Vert^2 - \mathds{E}\Vert\widetilde{\boldsymbol{\scriptstyle\mathcal{W}}}_n \Vert^2\vert$, for which the following two lemmas can be established:
\begin{lemma}\label{mse_l}
(\textbf{Deviation bounds of decentralized methods in the long term}). For a fixed small step size $\mu$ and local batch size $B$ such that 
\begin{align}\label{cbmu_l}
    \frac{1}{B} \le c\mu^{\eta}
\end{align}
where $c \ll \infty$ and $\eta \ge 0$, and under assumptions \ref{as1}, \ref{ass_ori} and \ref{ass_smooth}, and assume $J(w)$ is locally strongly-convex around $w^{\star}$, after sufficient iterations, it can be verified for consensus and diffusion that the second and fourth-order moments of 
 ${\widetilde{\boldsymbol{\scriptstyle\mathcal{W}}}}_{n}$, the second-order moment of  ${\widetilde{\boldsymbol{\scriptstyle\mathcal{W}}}}_{n}'$, and the approximation error of the short term model are upper bounded. Basically, it holds that 
\begin{align}
\mathds{E}\Vert{\widetilde{\boldsymbol{\scriptstyle\mathcal{W}}}}_{n}\Vert^2 &\le O(\mu^{\gamma})\\
\mathds{E}\Vert{\widetilde{\boldsymbol{\scriptstyle\mathcal{W}}}}_{n}'\Vert^2 &\le O(\mu^{\gamma})\\
\mathds{E}\Vert{\widetilde{\boldsymbol{\scriptstyle\mathcal{W}}}}_{n}\Vert^4 &\le O(\mu^{2\gamma})
\end{align}
and
\begin{align}
\vert \mathds{E}\Vert\widetilde{\boldsymbol{\scriptstyle\mathcal{W}}}_n'\Vert^2 - \mathds{E}\Vert\widetilde{\boldsymbol{\scriptstyle\mathcal{W}}}_n \Vert^2 \vert \le O(\mu^{1.5\gamma})
\end{align}
where $\gamma = \min\{1+\eta,2\}$.
\end{lemma}
\begin{lemma}\label{mse_centra_l}
(\textbf{Deviation bounds of the centralized method in the long term}). Under the same conditions of Lemma \ref{mse_l}, after sufficient iterations, the second and fourth-order moments of 
 ${\widetilde{\boldsymbol{\scriptstyle\mathcal{W}}}}_{n}$, the second-order moment of  ${\widetilde{\boldsymbol{\scriptstyle\mathcal{W}}}}_{n}'$, and the approximation error of the short term model, related to the centralized method are guaranteed to be upper bounded. Basically, it holds that 
\begin{align}
\mathds{E}\Vert{\widetilde{\boldsymbol{\scriptstyle\mathcal{W}}}}_{n}\Vert^2 &\le O(\mu^{1+\eta})\\
\mathds{E}\Vert{\widetilde{\boldsymbol{\scriptstyle\mathcal{W}}}}_{n}'\Vert^2 &\le O(\mu^{1+\eta})\\
\mathds{E}\Vert{\widetilde{\boldsymbol{\scriptstyle\mathcal{W}}}}_{n}\Vert^4 &\le O(\mu^{2(1+\eta)})
\end{align}
and
\begin{align}
\vert \mathds{E}\Vert\widetilde{\boldsymbol{\scriptstyle\mathcal{W}}}_n'\Vert^2 - \mathds{E}\Vert\widetilde{\boldsymbol{\scriptstyle\mathcal{W}}}_n \Vert^2 \vert \le O(\mu^{1.5(1+\eta)})
\end{align}
\end{lemma}
The Proofs for Lemmas \ref{mse_l} and \ref{mse_centra_l} can be found in Appendix \ref{mse_l_3}.

Recall (\ref{erave_p}), Lemmas \ref{mse_l} and \ref{mse_centra_l} guarantee that the excess-risk value $\mathrm{ER}_n$ satisfies the following equality:
\begin{align}\label{erave_p_l}
    \mathrm{ER}_n = \frac{1}{2K}\mathds{E}\Vert{\widetilde{\boldsymbol{\scriptstyle\mathcal{W}}}}_{n}'\Vert^2_{I\otimes\bar{H}} \pm O(\mu^{1.5\xi})
\end{align}
where $\xi = \gamma$ for decentralized methods, while $\xi = 1+\eta$ for the centralized method. 

Similar to Appendix \ref{per}
, we examine the size of $\mathds{E}\Vert{\widetilde{\boldsymbol{\scriptstyle\mathcal{W}}}}_{n}'\Vert^2_{I\otimes\bar{H}}$, where (\ref{wnph1}) is still valid. First, we know from Theorem 9.3 in \cite{sayed2014adaptation} that $\mathcal{C}$ is a stable matrix, thus when $n\to\infty$, we have:
\begin{align}
\lim_{n\to\infty}\mathds{E}\Vert\mathcal{C}^{n+1}{\widetilde{\boldsymbol{\scriptstyle\mathcal{W}}}}_{-1}'\Vert^2 = 0
\end{align}
By following the same derivation process of (\ref{sumci})--(\ref{wnphd12}), (\ref{wnphd12}) is sill valid.  Specifically, since $P_\alpha$ is a diagonal matrix where the absolute values of all elements are smaller than 1, in the steady state when $n\to\infty$, we have
\begin{align}\label{wnphd12_l_2}
    \lim_{n\to \infty}\mathds{E}\left\Vert\mathcal{C}^{n+1}{\widetilde{\boldsymbol{\scriptstyle\mathcal{W}}}}_{-1}' + \mu\sum\limits_{i=0}^{n} \mathcal{C}^{i}\mathcal{A}_2 d \right\Vert^2_{I\otimes\bar{H}}
    = \mu^2\Vert d^{\sf T}\mathcal{A}_2 \mathcal{V}_\alpha( I - \mathcal{P}_{\alpha})^{-1}\Vert^2_{I\otimes\bar{H}} \pm o(\mu^2)
\end{align}
In the centralized method where $\mathcal{V}_\alpha= 0$, we have
\begin{align}\label{mucad_centra_2}
    \lim_{n\to \infty}\mathds{E}\Vert\mathcal{C}^{n+1}{\widetilde{\boldsymbol{\scriptstyle\mathcal{W}}}}_{-1}' + \mu\sum\limits_{i=0}^{n} \mathcal{C}^{i}\mathcal{A}_2 d \Vert^2_{I\otimes\bar{H}} = 0
\end{align}

Then similar to (\ref{rsbnc})--(\ref{cashf3}), for the case of $n\to\infty$, we have the following equality for the gradient noise term:
\begin{align}\label{sssf1_l_f}
  \mathds{E}\left\Vert\sum\limits_{i=0}^{n} \mathcal{C}^{i}\mathcal{A}_2\boldsymbol{s}_i^B \right\Vert^2_{I\otimes\bar{H}}& =  \frac{K}{2\mu B}\mathrm{Tr}\left(\bar{R}_s \right)\pm O(\frac{\mu^{\frac{\xi}{2}}}{\mu B})   
\end{align}

We finally derive closed-form expressions for the excess risk in the long term when $n\to\infty$. Basically, for the \emph{centralized method}, we substitute (\ref{wnph1}), (\ref{mucad_centra_2}), (\ref{sssf1_l_f}) and $\xi = 1+\eta$ into (\ref{erave_p_l}), and obtain
\begin{align}
    \mathrm{ER}_{\infty,cen} &= \frac{\mu }{4B}\mathrm{Tr}\left(\bar{R}_s \right)\pm O\left(\mu^{1.5(1+\eta)}\right)
\end{align}

As for \emph{decentralized methods}, we separately show the results of consensus and diffusion. That is, for consensus, we substitute  (\ref{wnph1}), (\ref{wnphd12_l_2}), (\ref{sssf1_l_f}), $\xi = \gamma$ and $A_2 = I$ into (\ref{erave_p_l}), and obtain
\begin{align}\label{ep_consen_p}
    \mathrm{ER}_{\infty,con} = & \frac{\mu }{4B}\mathrm{Tr}\left(\bar{R}_s \right)+ \frac{\mu^2}{2K}\Vert d^{\sf T}\mathcal{V}_\alpha( I - \mathcal{P}_{\alpha})^{-1})\Vert^2_{I\otimes\bar{H}}\pm O(\mu^{1.5\gamma})   \pm  o(\mu^2)
\end{align}
Similarly, for diffusion, we substitute (\ref{wnph1}), (\ref{wnphd12_l_2}), (\ref{sssf1_l_f}), $\xi = \gamma$ and $A_2 = A$ into (\ref{erave_p_l}), and obtain
\begin{align}\label{ep_diff_p}
    \mathrm{ER}_{\infty,dif} = & \frac{\mu }{4B}\mathrm{Tr}\left(\bar{R}_s  \right)+ \frac{\mu^2}{2K}\Vert d^{\sf T}\mathcal{V}_\alpha\mathcal{P}_{\alpha}( I - \mathcal{P}_{\alpha})^{-1})\Vert^2_{I\otimes\bar{H}}\pm O(\mu^{1.5\gamma})   \pm  o(\mu^2)
\end{align}
In the large-batch training regime when $O(\mu^2)$  dominates $O(\mu^{1.5\gamma})$, it is obvious that
\begin{align}
    \mathrm{ER}_{\infty,cen} \le \mathrm{ER}_{\infty,con}, \; \; \mathrm{ER}_{\infty,cen} \le \mathrm{ER}_{\infty,dif} 
\end{align}
As for the long-term excess risk in (\ref{ep_consen_p}) and (\ref{ep_diff_p}), we can use the same proof logic from (\ref{ullp})--(\ref{dif_consen_s}), from which we obtain
\begin{align}
\Vert d^{\sf T}\mathcal{V}_\alpha \mathcal{P}_{\alpha}(I - \mathcal{P}_{\alpha})^{-1}\Vert^2_{I\otimes\bar{H}} = \sum\limits_{k=2}^{K} \frac{\lambda_k^2}{(1-\lambda_k)^2}\Vert x_k\Vert^2_{\bar{H}} \le \sum\limits_{k=2}^{K} \frac{1}{(1-\lambda_k)^2}\Vert x_k\Vert^2_{\bar{H}} = \Vert d^{\sf T}\mathcal{V}_\alpha(I - \mathcal{P}_{\alpha})^{-1}\Vert^2_{I\otimes\bar{H}}
\end{align}
so that
\begin{align}
    \mathrm{ER}_{\infty, dif} \le \mathrm{ER}_{\infty,con}
\end{align}

\section{Proof for Lemmas \ref{mse_l} and \ref{mse_centra_l}}\label{mse_l_3}
The proofs for Lemmas \ref{mse_l} and \ref{mse_centra_l} are similar to Appendices \ref{mse2}--\ref{aest}, and also similar to the proofs in \cite{sayed2014adaptation}.
\subsection{Proof for $\mathds{E}\Vert{\widetilde{\boldsymbol{\scriptstyle\mathcal{W}}}}_{n}\Vert^2$.}
Assume $J(w)$ is locally  $\nu$-strongly-convex around $w^{\star}$. We first verify the bounds for $\mathds{E}\Vert{\widetilde{\boldsymbol{\scriptstyle\mathcal{W}}}}_{n}\Vert^2$: By following the same logic of (\ref{vvt})--(\ref{barw2}), we substitute the strongly-convex condition into (\ref{barw2}), and choose $0< t = 1 -\mu \nu < 1$, then we obtain
\begin{align}\label{tildew2_l}
     \mathds{E}\left[\begin{array}{c}
         \Vert\bar{\boldsymbol{w}}_{n}\Vert^2\\
          \Vert\check{\boldsymbol{w}}_{n}\Vert^2
    \end{array}\right]\le \left[\begin{array}{cc}
       1 - \mu\nu + O(\mu^2)& O(\mu) \\
         O(\mu^2)&  \rho(P_\alpha) + O(\mu^2)
    \end{array}\right]\left[\begin{array}{c}
         \mathds{E}\Vert\bar{\boldsymbol{w}}_{n-1}\Vert^2\\
         \mathds{E} \Vert\check{\boldsymbol{w}}_{n-1}\Vert^2
    \end{array}\right] + \left[\begin{array}{c}
         O(\frac{\mu^2}{B})\\
         O(\mu^2)
    \end{array}\right]
\end{align}
Let 
\begin{align}
    \Gamma_3 = \left[\begin{array}{cc}
       1 - O(\mu)& O(\mu) \\
         O(\mu^2)& \rho(P_\alpha) + O(\mu^2)
    \end{array}\right]
\end{align}
It can be easily verified that the spectral radius of $\Gamma_3$ denoted by $\rho(\Gamma_3)$ is smaller than 1 with sufficiently small $\mu$:
\begin{align}\label{sta_ga}
    \rho(\Gamma_3) \le \Vert\Gamma_3\Vert_1 = \max\{1 - O(\mu) + O(\mu^2), \rho(P_\alpha)  + O(\mu)\} < 1
\end{align}
Thus $\Gamma_3$ is a stable matrix such that 
\begin{align}
    \lim_{n\to \infty} \Gamma_3^n = 0
\end{align}
By iterating (\ref{tildew2_l}), we have
\begin{align}\label{ftw_l}
    \lim_{n\to\infty} \mathds{E}\left[\begin{array}{c}
         \Vert\bar{\boldsymbol{w}}_{n}\Vert^2\\
          \Vert\check{\boldsymbol{w}}_{n}\Vert^2
    \end{array}\right] &\le \Gamma_3^{n+1}\mathds{E}\left[\begin{array}{c}
         \Vert\bar{\boldsymbol{w}}_{-1}\Vert^2\\
          \Vert\check{\boldsymbol{w}}_{-1}\Vert^2
    \end{array}\right]+ \sum\limits_{i = 0}^{n} \Gamma_3^{i}\left[\begin{array}{c}
         O(\frac{\mu^2}{B})\\
         O(\mu^2)
    \end{array}\right] = (I - \Gamma_3)^{-1}\left[\begin{array}{c}
         O(\frac{\mu^2}{B})\\
         O(\mu^2)
    \end{array}\right] \notag\\
    &= \left[\begin{array}{cc}
       O(\frac{1}{\mu})& O(1) \\
         O(\mu)&  O(1)
    \end{array}\right] \left[\begin{array}{c}
         O(\frac{\mu^2}{B})\\
         O(\mu^2)
    \end{array}\right] = \left[\begin{array}{c}
         O(\frac{\mu}{B}) + O(\mu^2)\\
         O(\mu^2)
    \end{array}\right]
\end{align}
from which we have 
\begin{align}\label{wn2_l}
\lim_{n\to\infty}\mathds{E}\Vert{\widetilde{\boldsymbol{\scriptstyle\mathcal{W}}}}_{n}\Vert^2 \le \Vert\mathcal{V}\Vert^2\mathds{E}\Vert{\mathcal{V}^{\sf T}\widetilde{\boldsymbol{\scriptstyle\mathcal{W}}}}_{n}\Vert^2 = \Vert\mathcal{V}\Vert^2\mathds{E}(\Vert\bar{\boldsymbol{w}}_{n}\Vert^2+\Vert\check{\boldsymbol{w}}_{n}\Vert^2)\le O(\frac{\mu}{B}) + O(\mu^2) = O(\mu^{\gamma})
\end{align}
where $\gamma = \min\{1+\eta,2\}$.

Similarly, for the \emph{centralized method}, we only need to analyze $\mathds{E}\Vert\bar{\boldsymbol{w}}_{n}\Vert^2$:
\begin{align}\label{fbar_centra_l}
     \lim_{n\to\infty} \mathds{E}\Vert \bar{\boldsymbol{w}}_{n}\Vert^2 \le (1 - \mu\nu + O(\mu^2))^{n+1}\mathds{E}\Vert\bar{\boldsymbol{w}}_{-1}\Vert^2 + O(\frac{1}{\mu})\times O(\frac{\mu^2}{B}) = O(\frac{\mu}{B}) = O(\mu^{1+\eta})
\end{align}
where 
\begin{align}
    \lim_{n\to\infty} (1 - \mu\nu + O(\mu^2))^{n+1} = 0
\end{align}
for sufficiently small $\mu$.
\subsection{Proof for $\mathds{E}\Vert{\widetilde{\boldsymbol{\scriptstyle\mathcal{W}}}}_{n}\Vert^4$.}
 The derivation process is similar to Appendix \ref{mse4}. The only difference is in (\ref{barw4}) for the strongly-convex case, if we choose $t = 1 - \mu\nu$, it holds that
\begin{align}\label{f_barw_l}
    \mathds{E}\Vert\bar{\boldsymbol{w}}_n\Vert^4 \le&(1-\mu\nu)\mathds{E}\Vert\bar{\boldsymbol{w}}_{n-1}\Vert^4 + O(\mu)\mathds{E}\Vert\check{\boldsymbol{w}}_{n-1}\Vert^4 + O(\frac{\mu^4}{B^2})\mathds{E}\Vert\bar{\boldsymbol{w}}_{n-1}\Vert^4 + O(\frac{\mu^4}{B^2})\mathds{E}\Vert\check{\boldsymbol{w}}_{n-1}\Vert^4 + O(\frac{\mu^4}{B^2})\notag\\
    & +O(\frac{\mu^2}{B})\mathds{E}\Vert\bar{\boldsymbol{w}}_{n-1}\Vert^2(\Vert\check{\boldsymbol{w}}_{n-1}\Vert^2 +\Vert\bar{\boldsymbol{w}}_{n-1}\Vert^2 +O(1)) + O(\frac{\mu^3}{B})\mathds{E}\Vert\check{\boldsymbol{w}}_{n-1}\Vert^2(\Vert\check{\boldsymbol{w}}_{n-1}\Vert^2 +\Vert\bar{\boldsymbol{w}}_{n-1}\Vert^2 +O(1))\notag\\
    &\le \left(1 - \mu\nu + O(\frac{\mu^2}{B})\right)\mathds{E}\Vert\bar{\boldsymbol{w}}_{n-1}\Vert^4 + O(\mu)\mathds{E}\Vert\check{\boldsymbol{w}}_{n-1}\Vert^4 + O(\frac{\mu^3}{B^2}) + O(\frac{\mu^4}{B})
\end{align}
from which we have
\begin{align}\label{f_w4_l}
\mathds{E}\left[\begin{array}{c}
         \Vert\bar{\boldsymbol{w}}_{n}\Vert^4\\
          \Vert\check{\boldsymbol{w}}_{n}\Vert^4
    \end{array}\right]\le \left[\begin{array}{cc}
       1 - O(\mu) & O(\mu) \\
         O(\frac{\mu^2}{B})&  \rho(P_\alpha)+O(\frac{\mu^2}{B})
    \end{array}\right]\left[\begin{array}{c}
         \mathds{E}\Vert\bar{\boldsymbol{w}}_{n-1}\Vert^4\\
         \mathds{E} \Vert\check{\boldsymbol{w}}_{n-1}\Vert^4
    \end{array}\right] + \left[\begin{array}{c}
         O(\frac{\mu^3}{B^2}) + O(\frac{\mu^4}{B})\\
         O(\mu^4)
    \end{array}\right]
\end{align}
Let 
\begin{align}
    \Gamma_4 = \left[\begin{array}{cc}
       1 - O(\mu) & O(\mu) \\
         O(\frac{\mu^2}{B})&  \rho(P_\alpha)+O(\frac{\mu^2}{B})
    \end{array}\right]
\end{align}
Similar to (\ref{sta_ga}), it can be verified that $\Gamma_4$ is a stable matrix. Then by iterating (\ref{f_w4_l}), after enough iterations when $n\to\infty$ we have
\begin{align}\label{f_w42_l}
\mathds{E}\left[\begin{array}{c}
         \Vert\bar{\boldsymbol{w}}_{n}\Vert^4\\
          \Vert\check{\boldsymbol{w}}_{n}\Vert^4
    \end{array}\right]& \le  (I-\Gamma_4)^{-1}\left[\begin{array}{c}
         O(\frac{\mu^3}{B^2}) + O(\frac{\mu^4}{B})\\
         O(\mu^4)
    \end{array}\right]  = \left[\begin{array}{cc}
       O(\frac{1}{\mu})& O(1) \\
         O(\frac{\mu}{B})&  O(1)
    \end{array}\right]\left[\begin{array}{c}
         O(\frac{\mu^3}{B^2}) + O(\frac{\mu^4}{B})\\
         O(\mu^4)
    \end{array}\right] \notag\\
    & =  \left[\begin{array}{c}
         O(\frac{\mu^2}{B^2})+O(\frac{\mu^3}{B}) +O( \mu^4)\\
         O(\mu^4)
    \end{array}\right]
\end{align}
and, therefore,\begin{align}\label{ffw4_2}
\mathds{E}\Vert{\widetilde{\boldsymbol{\scriptstyle\mathcal{W}}}}_{n}\Vert^4 &\le \Vert\mathcal{V}\Vert^4\mathds{E}\Vert{\mathcal{V}^{\sf T}\widetilde{\boldsymbol{\scriptstyle\mathcal{W}}}}_{n}\Vert^4 = \Vert\mathcal{V}\Vert^4 \mathds{E}\left(\Vert\bar{\boldsymbol{w}}_{n}\Vert^2 + \Vert\check{\boldsymbol{w}}_{n}\Vert^2\right)^2 \overset{(a)}{\le} 2\Vert\mathcal{V}\Vert^4 \mathds{E}\left(\Vert\bar{\boldsymbol{w}}_{n}\Vert^4 + \Vert\check{\boldsymbol{w}}_{n}\Vert^4\right) \notag\\
   &\le O(\frac{\mu^2}{B^2})+O(\frac{\mu^3}{B}) +O( \mu^4) = O(\mu^{2+2\eta})+O(\mu^{3+\eta})+O(\mu^4)
\end{align}
where $(a)$ follows from Jensen's inequality. If $\eta \le 1$, then
\begin{align}\label{eta1}
    \mathds{E}\Vert{\widetilde{\boldsymbol{\scriptstyle\mathcal{W}}}}_{n}\Vert^4 \le O(\mu^{2+2\eta})
\end{align}
while if $\eta>1$, then
\begin{align}\label{eta2}
    \mathds{E}\Vert{\widetilde{\boldsymbol{\scriptstyle\mathcal{W}}}}_{n}\Vert^4 \le O(\mu^4)
\end{align}
Combining (\ref{eta1}) and (\ref{eta2}), we have
\begin{align}\label{ffw4_n}
    \lim_{n\to\infty}\mathds{E}\Vert{\widetilde{\boldsymbol{\scriptstyle\mathcal{W}}}}_{n}\Vert^4 \le O(\mu^{2\gamma})
\end{align}
Similarly, for the centralized method, we substitute (\ref{fbar_centra_l}) into (\ref{f_barw_l}) and iterate it, and obtain
\begin{align}
\label{f_bar_s4_centra_2}
    \mathds{E}\Vert\bar{\boldsymbol{w}}_n\Vert^2 {\le} \left(1 - \mu\nu + O(\frac{\mu^2}{B})\right)^{n+1}\mathds{E}\Vert\bar{\boldsymbol{w}}_{-1}\Vert^4  + \frac{1 - \left(1 - \mu\nu + O(\frac{\mu^2}{B})\right)^{n+1} }{1 - (1 - \mu\nu + O(\frac{\mu^2}{B}))}\times O(\frac{\mu^3}{B^2})
\end{align}
from which for $n\to \infty$, the \emph{centralized method} satisfies
\begin{align}
\label{wn4_centra_2}
\lim_{n\to\infty}\mathds{E}\Vert{\widetilde{\boldsymbol{\scriptstyle\mathcal{W}}}}_{n}\Vert^4  \le O(\frac{\mu^2}{B^2}) = O(\mu^{2(1+\eta)})
\end{align}
\subsection{Proof for  $\vert \mathds{E}\Vert\widetilde{\boldsymbol{\scriptstyle\mathcal{W}}}_n'\Vert^2 - \mathds{E}\Vert\widetilde{\boldsymbol{\scriptstyle\mathcal{W}}}_n \Vert^2 \vert$ and $\mathds{E}\Vert{\widetilde{\boldsymbol{\scriptstyle\mathcal{W}}}}_{n}'\Vert^2$.}
We finally analyze the bounds of $\vert \mathds{E}\Vert\widetilde{\boldsymbol{\scriptstyle\mathcal{W}}}_n'\Vert^2 - \mathds{E}\Vert\widetilde{\boldsymbol{\scriptstyle\mathcal{W}}}_n \Vert^2 \vert$ and $\mathds{E}\Vert{\widetilde{\boldsymbol{\scriptstyle\mathcal{W}}}}_{n}'\Vert^2$. The proofs are similar to Appendix \ref{aest}. The only difference is that we now consider the long-term properties of $\mathds{E}\Vert\widetilde{\boldsymbol{\scriptstyle\mathcal{W}}}_n'\Vert^2 - \mathds{E}\Vert\widetilde{\boldsymbol{\scriptstyle\mathcal{W}}}_n \Vert^2 \vert$ when $n\to \infty$. Similar to (\ref{gammap}) and (\ref{szn_2}), after enough iterations when $n$ is sufficiently large, and with (\ref{gammapin}) and (\ref{ffw4_n}), we have
\begin{align}\label{szn_2_l}
    \lim\limits_{n\to \infty}\mathds{E}\left[\begin{array}{c}
         \Vert\bar{\boldsymbol{z}}_{n}\Vert^2\\
          \Vert\check{\boldsymbol{z}}_{n}\Vert^2
    \end{array}\right] \le (I - \Gamma')^{-1}\left[\begin{array}{c}
        O(\mu^{2\gamma+1}) \\
          O(\mu^{2\gamma+2})
    \end{array}\right]\le  \left[\begin{array}{cc}
       O(\frac{1}{\mu}) & O(1) \\
         O(\mu)&  O(1)
    \end{array}\right] \left[\begin{array}{c}
        O(\mu^{2\gamma+1})\\
          O(\mu^{2\gamma+2})\end{array}\right] =  O(\mu^{2\gamma}) 
\end{align}
from which we  have
\begin{align}\label{ew4_s_l}
\lim\limits_{n\to \infty}\mathds{E}\Vert\widetilde{\boldsymbol{\scriptstyle\mathcal{W}}}_n' - \widetilde{\boldsymbol{\scriptstyle\mathcal{W}}}_n \Vert^2 = \mathds{E}  \Vert\mathcal{V}\mathcal{V}^{\sf T}{{\boldsymbol{{z}}}}_{n}\Vert^2 \le \Vert\mathcal{V}^2\Vert\mathds{E}\Vert\mathcal{V}^{\sf T}{{\boldsymbol{{z}}}}_{n}\Vert^2=\Vert\mathcal{V}^2\Vert\mathds{E}(\Vert\bar{\boldsymbol{z}}_{n}\Vert^2+\Vert\check{\boldsymbol{z}}_{n}\Vert^2)\le O(\mu^{2\gamma}) 
\end{align}
Then similar to (\ref{dian})--(\ref{ewwp_4}), we obtain
\begin{align}
\lim_{n\to\infty}\mathds{E}\Vert\widetilde{\boldsymbol{\scriptstyle\mathcal{W}}}_n' \Vert^2 \le O(\mu^{\gamma}), \quad\quad
    \lim_{n\to\infty}\vert \mathds{E}\Vert\widetilde{\boldsymbol{\scriptstyle\mathcal{W}}}_n' \Vert^2 - \mathds{E}\Vert\widetilde{\boldsymbol{\scriptstyle\mathcal{W}}}_n\Vert^2 \vert \le O(\mu^{1.5\gamma})
\end{align}

As for the \emph{centralized method}, we substitute (\ref{wn4_centra_2}) into (\ref{s_barz_centra}), with which when $n\to\infty$, we have
\begin{align}
\lim\limits_{n\to \infty}\mathds{E}\Vert\widetilde{\boldsymbol{\scriptstyle\mathcal{W}}}_n' \Vert^2 \le O(\mu^{1+\eta}), \quad\quad
    \lim\limits_{n\to \infty}\vert \mathds{E}\Vert\widetilde{\boldsymbol{\scriptstyle\mathcal{W}}}_n' \Vert^2 - \mathds{E}\Vert\widetilde{\boldsymbol{\scriptstyle\mathcal{W}}}_n\Vert^2 \vert \le O(\mu^{1.5(1+\eta))})
\end{align}

\section{Additional simulation results}\label{asr_ap}
{In this section, we detail the experimental setup for our study. The image classification tasks are performed on the CIFAR-10 and CIFAR-100 datasets. For data preprocessing, we adopt the standard data augmentation scheme, and normalize the data using channel-specific means and standard deviations, as outlined in \cite{HeZRS16, LinSPJ20}. In decentralized training, the full dataset is randomly split into 16 subsets, each assigned to one local model, which only observes its respective subset during training. Centralized training follows the conventional single-agent learning approach. All decentralized experiments are simulated on a single Tesla V100 32GB GPU. For centralized experiments, due to CUDA memory limitations, the models are distributed across four Tesla V100 GPUs.} 

{We evaluate the three distributed methods across various neural network architectures, graph structures, and batch sizes to ensure consistent and reliable results. The chosen neural network architectures include ResNet-18, WideResNet-28-10, and DenseNet-121. For the graph structures in the multi-agent system, simulations are conducted on two distinct topologies: (1) A random graph generated using the Metropolis rule \cite{sayed2014adaptation} (Figure \ref{graph16}(a)). (2) A ring graph structure (Figure \ref{graph16}(b)). Both topologies consist of 16 nodes, with each node corresponding to a local model. ResNet-18 is evaluated on both random and ring graphs, while WideResNet-28-10 and DenseNet-121 are simulated only on the random graph. Regarding local batch sizes, we consider three configurations: 128, 256, and 512. In centralized training, the global batch size is scaled by a factor of $K$ (the number of nodes) relative to the local batch size. For example, when the local batch is 128 in decentralized methods, the corresponding global batch size for centralized training is $128\times16 = 2048$. Additionally, ResNet-18 is evaluated on all of the three local batch size configurations, and WideResNet-28-10 and DenseNet-121 are simulated with  the local batch 256.}

{For all experiments, the vanilla SGD optimizer is used to align with our theoretical findings, with the momentum parameter set to 0. We employ a piecewise learning rate schedule, starting with an initial learning rate of 0.2, which is reduced by a factor of 10 at $50\%$ and $75\%$ of the total epochs. The weight decay parameter is set to 0 for CIFAR-10 experiments and $1e-4$ for CIFAR-100 experiments. The number of training epochs varies across different settings, and they can be found from the x-axis of the plots depicting the evolution of training risk.}

{We now show the results associated with WideResNet-28-10 and DenseNet-121 on CIFAR10 and CIFAR100. We simulate these two neural networks with batch size $B = 256$:}

\begin{figure*}[ht]
\begin{center}
\subfigure[WideResNet-28-10, CIFAR10]{\includegraphics[width=.3\linewidth]{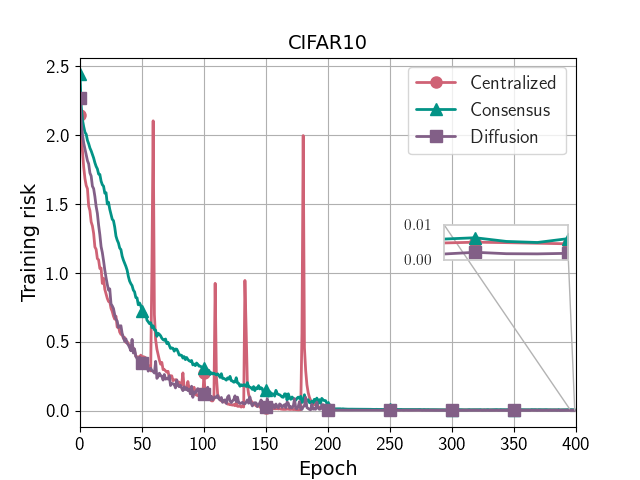}}
\subfigure[WideResNet-28-10, CIFAR100]{\includegraphics[width=.3\linewidth]{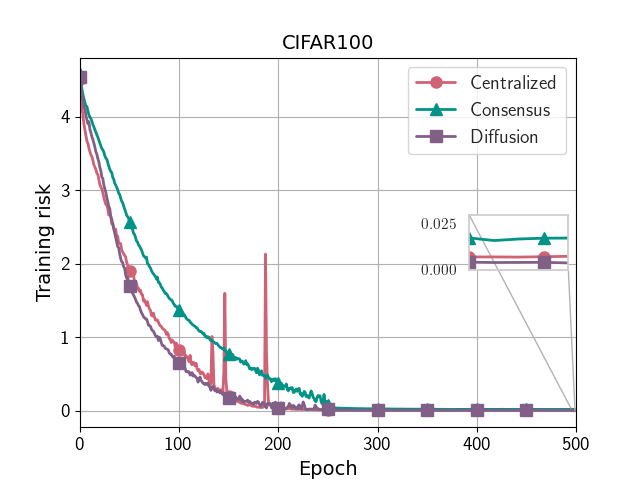}}\\
\subfigure[DenseNet-121, CIFAR10]{\includegraphics[width=.3\linewidth]{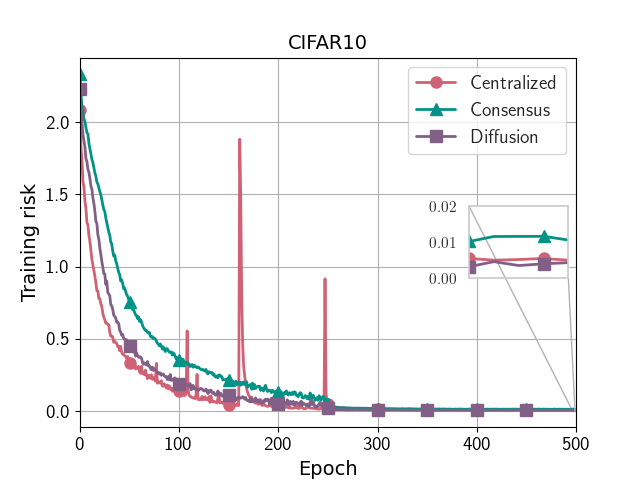}}
\subfigure[DenseNet-121, CIFAR100]{\includegraphics[width=.3\linewidth]{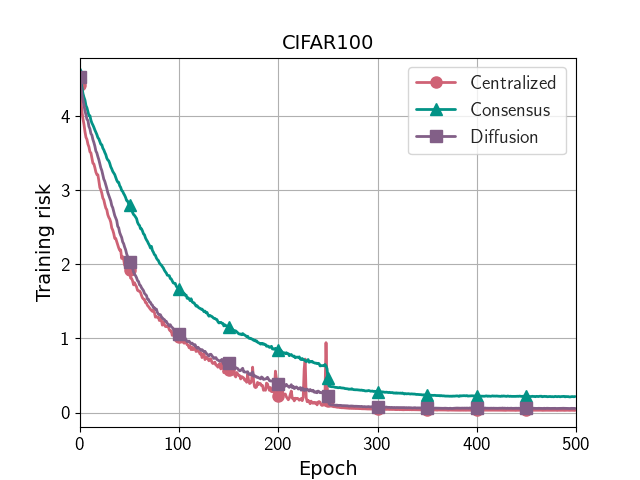}}
\caption{The evolution of training risk with WideResNet-28-10 and DenseNet-121 when training from scratch.}
\label{f4}
\end{center}
\end{figure*}

\begin{figure*}[ht]
\begin{center}
\subfigure[WideResNet-28-10, CIFAR10]{\includegraphics[width=.3\linewidth]{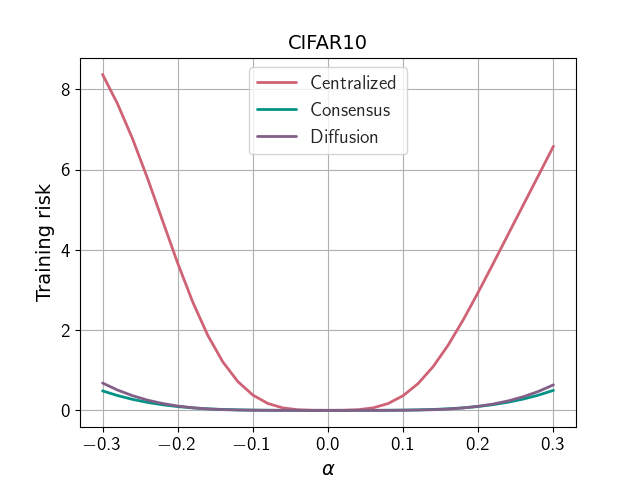}}
\subfigure[WideResNet-28-10, CIFAR100]{\includegraphics[width=.3\linewidth]{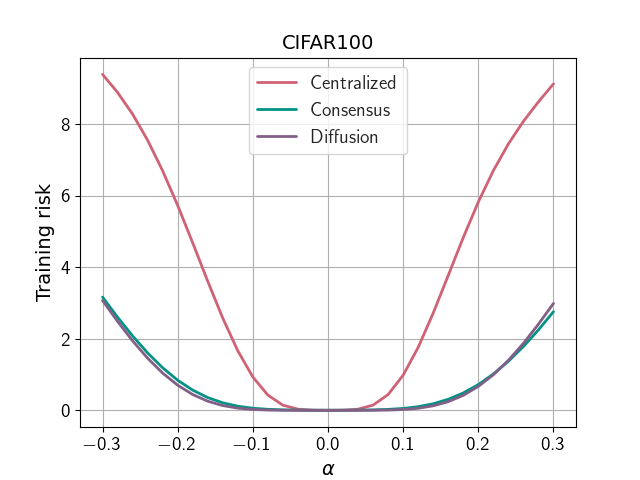}}\\
\subfigure[DenseNet-121, CIFAR10]{\includegraphics[width=.3\linewidth]{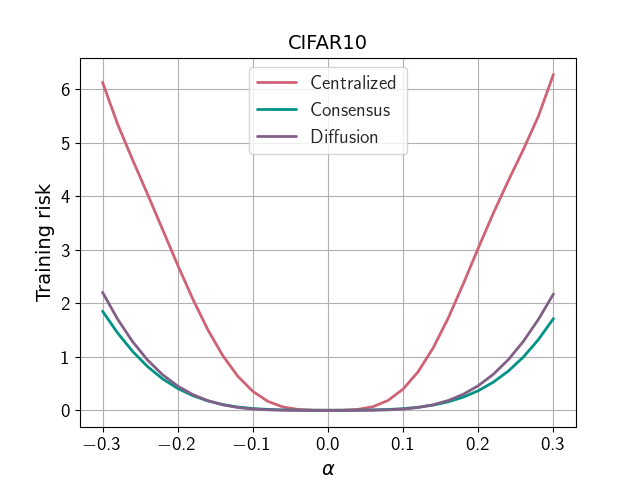}}
\subfigure[DenseNet-121, CIFAR100]{\includegraphics[width=.3\linewidth]{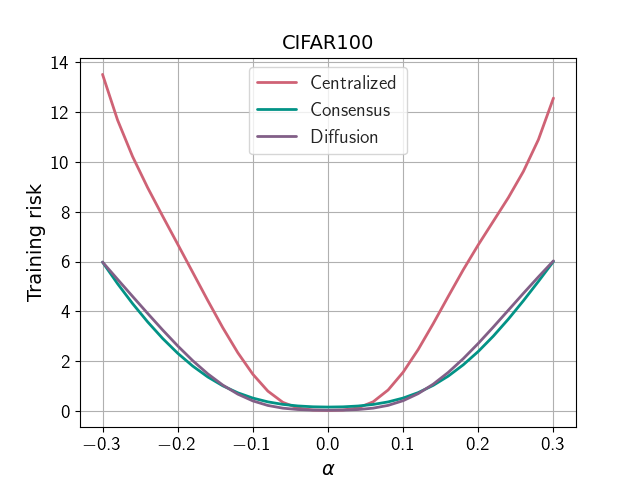}}
\caption{Flatness visualization of models trained from scratch with WideResNet-28-10 and DenseNet-121.}
\label{f5}
\end{center}
\end{figure*}

{We first show the results corresponding to the scenario of training from scratch. We illustrate the optimization performance and flatness separately in Figure \ref{f4} and \ref{f5}. We further show the test accuracy of the three methods in Table \ref{tableap}, from which we observe again that diffusion consistently outperforms the other two methods. Similar to ResNet-18, this phenomenon can be interpreted by the trade-off between flatness and optimization. For the comparison between decentralized and centralized methods, the flatness plays an important role since it can be observed from Figure \ref{f4} and \ref{f5} that decentralized training methods enable flatter models than the centralized approach without significantly losing optimization performance. However, as for the comparison between diffusion and consensus, the optimization performance values more as their flatness is very close. Specifically, we observe from Figure \ref{f4} that diffusion has smaller training risk than consensus. All in all, in the scenario of training from scratch, diffusion achieves a favorable balance between optimization and generalization so that provides the best accuracy among the three methods.}

\begin{table*}[ht]
\caption{Test accuracy of distributed algorithms trained from scratch on CIFAR10 and CIFAR100 with WideResNet-28-10 and Densenet-121.}
\label{tableap}
\footnotesize
\centering
\begin{spacing}{1.6}
\begin{tabular}{ccccc}
\toprule
Dataset&  Architecture & Centralized & Consensus & Diffusion\\
\midrule
{\multirow{2}{*}{CIFAR10}} & {WideResNet-28-10}& $89.76\pm0.34\%$ &$92.61\pm0.11\%$&$\textbf{92.65}\pm0.04\%$\\
                           & DenseNet-121    & $89.23\pm0.12\%$ &$91.16\pm0.08\%$&$\textbf{91.47}\pm0.18\%$\\
\cline{1-5}
{\multirow{2}{*}{CIFAR100}} & WideResNet-28-10& $67.61\pm0.20\%$ &$73.13\pm0.37\%$&$\textbf{74.18}\pm0.03\%$\\
                           & DenseNet-121    & $64.28\pm0.12\%$ &$68.05\pm0.33\%$&$\textbf{68.29}\pm0.31\%$\\
\bottomrule
\end{tabular}
\end{spacing}
\end{table*}

\begin{figure*}[ht]
\begin{center}
\subfigure[WideResNet-28-10, CIFAR10]{\includegraphics[width=.3\linewidth]{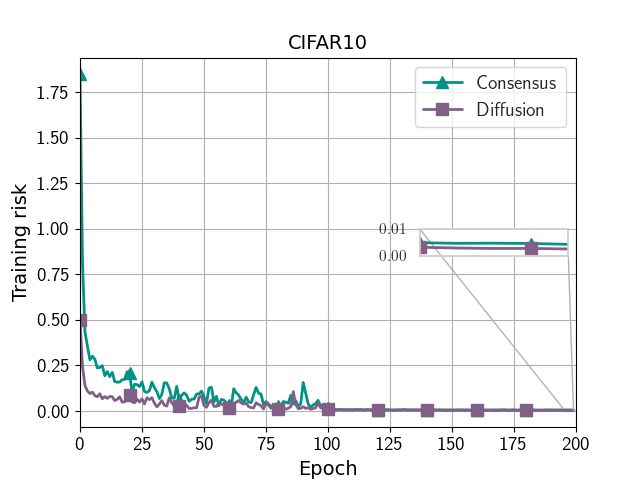}}
\subfigure[WideResNet-28-10, CIFAR100]{\includegraphics[width=.3\linewidth]{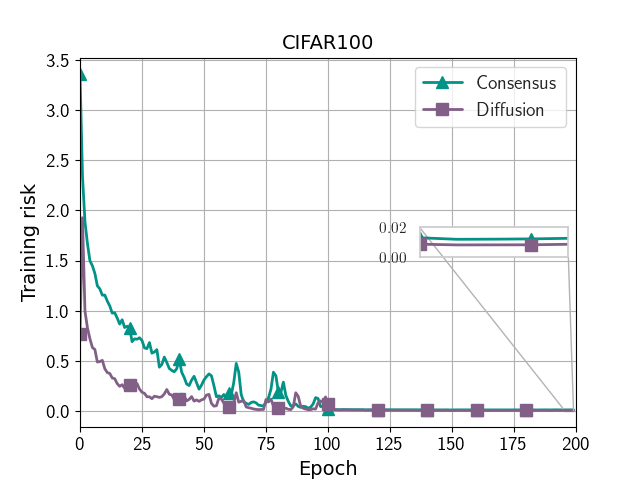}}\\
\subfigure[DenseNet-121, CIFAR10]{\includegraphics[width=.3\linewidth]{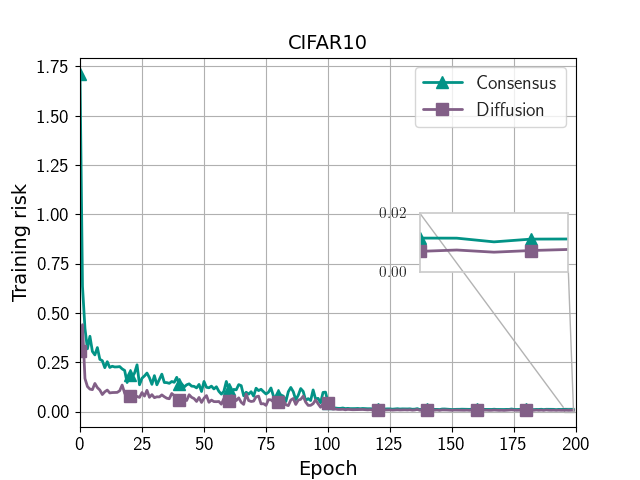}}
\subfigure[DenseNet-121, CIFAR100]{\includegraphics[width=.3\linewidth]{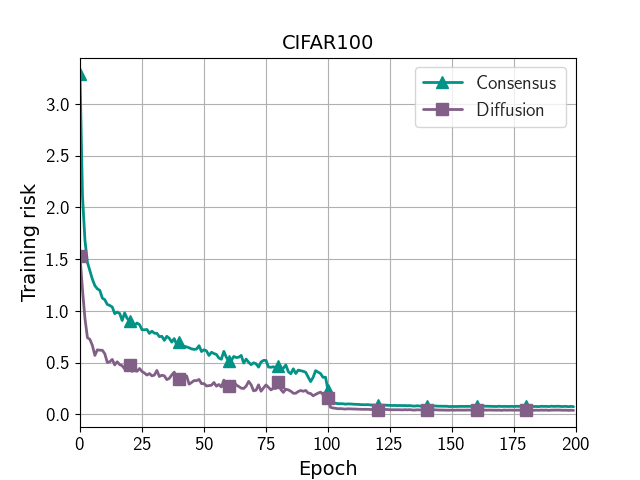}}
\caption{{The evolution of training risk with WideResNet-28-10 and DenseNet-121 using pretrained initialization.}}
\label{f4_pretrained}
\end{center}
\end{figure*}

\begin{figure*}[ht]
\begin{center}
\subfigure[WideResNet-28-10, CIFAR10]{\includegraphics[width=.3\linewidth]{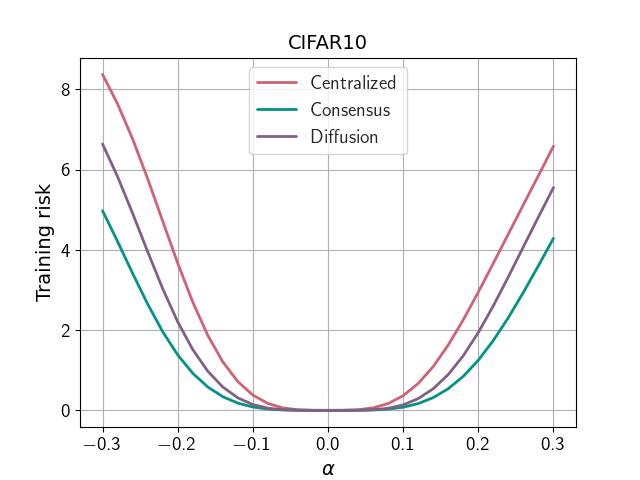}}
\subfigure[WideResNet-28-10, CIFAR100]{\includegraphics[width=.3\linewidth]{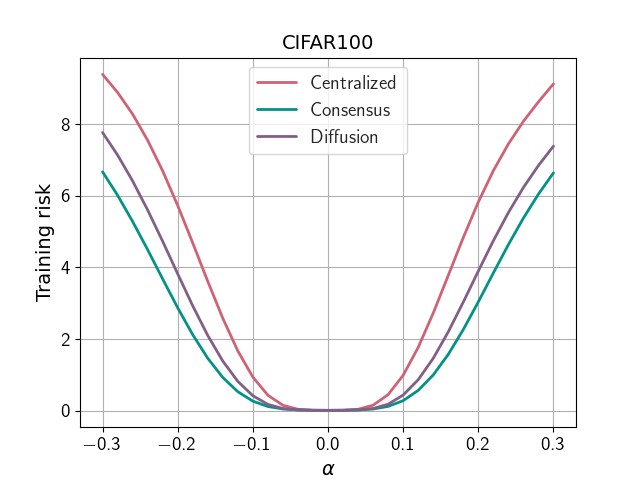}}\\
\subfigure[DenseNet-121, CIFAR10]{\includegraphics[width=.3\linewidth]{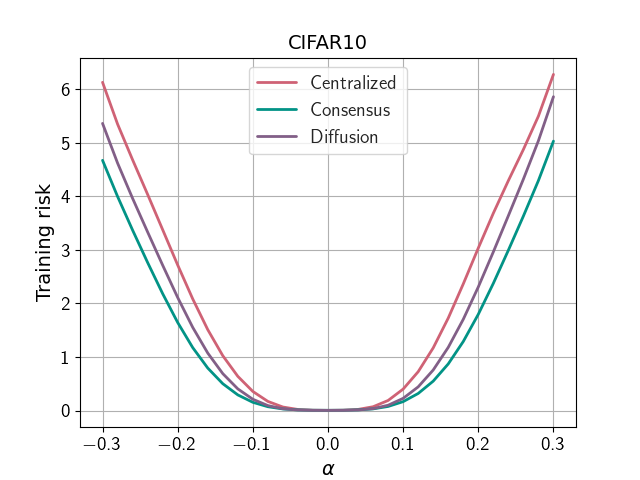}}
\subfigure[DenseNet-121, CIFAR100]{\includegraphics[width=.3\linewidth]{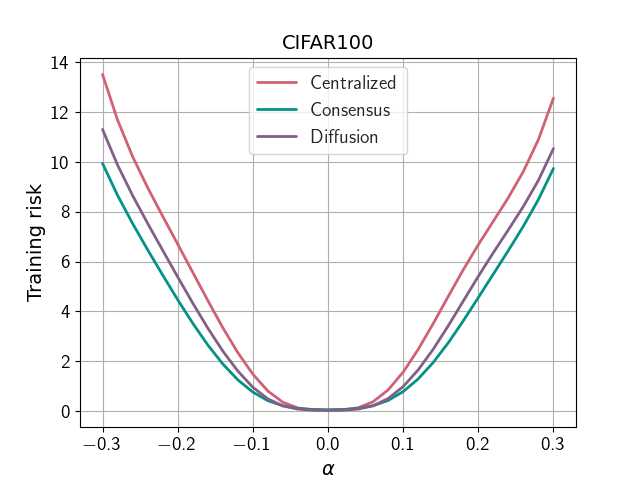}}
\caption{{Flatness visualization of models using pretrained initialization with WideResNet-28-10 and DenseNet-121.}}
\label{f5_pretrained}
\end{center}
\end{figure*}

{We next present the results of the scenario where decentralized algorithms are initialized at pretrained models. The optimization performance and flatness are depicted in Figure \ref{f4_pretrained} and \ref{f5_pretrained}. Additionally, the test accuracy is shown in Table \ref{tableap_pretrain}. Similar to the results of ResNet-18, decentralized methods converge to flatter minima than the centralized method without losing significant optimization performance, enabling them achieving higher test accuracy. Furthermore, when we delve more into the comparison between consensus and diffusion, consensus exhibits better test accuracy by converging to flatter minima while maintaining comparable optimization performance. In summary, in the scenario of pretrained initialization, consensus strikes the most favorable balance between optimization and generalization among the three methods.}

\begin{table*}[ht]
\caption{{Test accuracy of distributed algorithms using pretrained initialization on CIFAR10 and CIFAR100 with WideResNet-28-10 and Densenet-121.}}
\label{tableap_pretrain}
\footnotesize
\centering
\begin{spacing}{1.6}
\begin{tabular}{ccccc}
\toprule
Dataset&  Architecture & Centralized & Consensus & Diffusion\\
\midrule
{\multirow{2}{*}{CIFAR10}} & {WideResNet-28-10}& $89.76\pm0.34\%$ &$\textbf{91.27}\pm0.26\%$&$90.96\pm0.41\%$\\
                           & DenseNet-121    & $89.23\pm0.12\%$ &$\textbf{90.42}\pm0.26\%$&$90.09\pm0.17\%$\\
\cline{1-5}
{\multirow{2}{*}{CIFAR100}} & WideResNet-28-10& $67.61\pm0.20\%$ &$\textbf{71.30}\pm0.22\%$&$69.82\pm0.12\%$\\
                           & DenseNet-121    & $64.28\pm0.12\%$ &$\textbf{67.45}\pm0.20\%$&$66.50\pm0.16\%$\\
\bottomrule
\end{tabular}
\end{spacing}
\end{table*}

\end{document}